\documentclass{article}

\usepackage{microtype}
\usepackage{graphicx}
\usepackage{subfigure}
\usepackage{booktabs} 

\usepackage{bbm}

\usepackage{hyperref}

\usepackage{amsmath,amssymb,graphicx,color,booktabs,bm,relsize,enumitem,multirow,amsthm,subfigure,epsfig, caption,mathtools,xcolor}

\usepackage{adjustbox}
\newcommand{\reviewer}[3]{
	\expandafter\newcommand\csname #1\endcsname[1]{
		\textcolor{#3}{[#2: ##1]}
	}
}

\usepackage[accepted]{icml2020}

\def\mytitle{Context-aware Dynamics Model \\for Generalization in Model-Based Reinforcement Learning} 

\icmltitlerunning{Context-aware Dynamics Model for Generalization in Model-Based Reinforcement Learning}

\begin{document}

\twocolumn[
\icmltitle{\mytitle}



\icmlsetsymbol{equal}{*}

\begin{icmlauthorlist}
\icmlauthor{Kimin Lee}{ucb,equal}
\icmlauthor{Younggyo Seo}{kaist,equal}
\icmlauthor{Seunghyun Lee}{kaist}
\icmlauthor{Honglak Lee}{umich,gb}
\icmlauthor{Jinwoo Shin}{kaist}
\end{icmlauthorlist}

\icmlaffiliation{kaist}{KAIST}
\icmlaffiliation{gb}{Google Brain}
\icmlaffiliation{ucb}{UC Berkeley}
\icmlaffiliation{umich}{University of Michigan Ann Arbor}

\icmlcorrespondingauthor{Kimin Lee}{kiminlee@berkeley.edu}

\icmlkeywords{Machine Learning, ICML}
\vskip 0.3in
]



\printAffiliationsAndNotice{\icmlEqualContribution} 

\begin{abstract}
Model-based reinforcement learning (RL) enjoys several benefits, such as data-efficiency and planning, by learning a model of the environment’s dynamics. However, learning a global model that can generalize across different dynamics is a challenging task. To tackle this problem, we decompose the task of learning a global dynamics model into two stages: (a) learning a context latent vector that captures the local dynamics, then (b) predicting the next state conditioned on it. In order to encode dynamics-specific information into the context latent vector, we introduce a novel loss function that encourages the context latent vector to be useful for predicting both forward and backward dynamics. The proposed method achieves superior generalization ability across various simulated robotics and control tasks, compared to existing RL schemes.
\end{abstract}

\section{Introduction} \label{sec:intro}

Model-based reinforcement learning (RL) with high-capacity function approximators, 
such as deep neural networks (DNNs), has been used to solve a variety of sequential decision-making problems, including board games (e.g., Go and Chess \cite{schrittwieser2019mastering}), video games (e.g., Atari games \cite{kaiser2019model,schrittwieser2019mastering}), and complex robotic control tasks \cite{zhang2018solar,nagabandi2018learning,hafner2019dream}.
By learning a model of the environment's dynamics and utilizing the dynamics model for planning, model-based RL achieves superior data-efficiency to model-free RL methods in general \cite{deisenroth2011pilco,levine2014learning}, and sometimes even beats model-free RL methods trained on sufficient amount of data
\cite{schrittwieser2019mastering,hafner2018learning,hafner2019dream,kaiser2019model}. 

\begin{figure} [t] \centering
\subfigure[CartPole with varying pole lengths]
{
\includegraphics[width=0.47\textwidth]{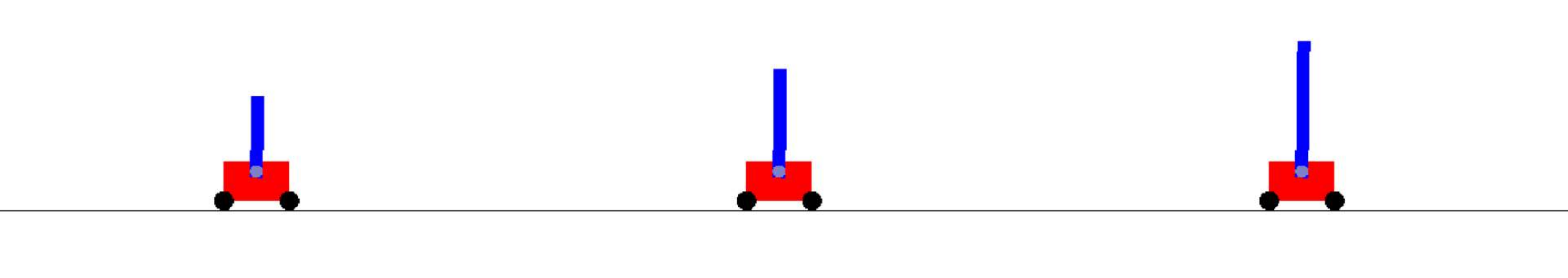} \label{fig:cartpole_example}} 
\\
\subfigure[Pendulum with varying pendulum lengths]{
\includegraphics[width=0.47\textwidth]{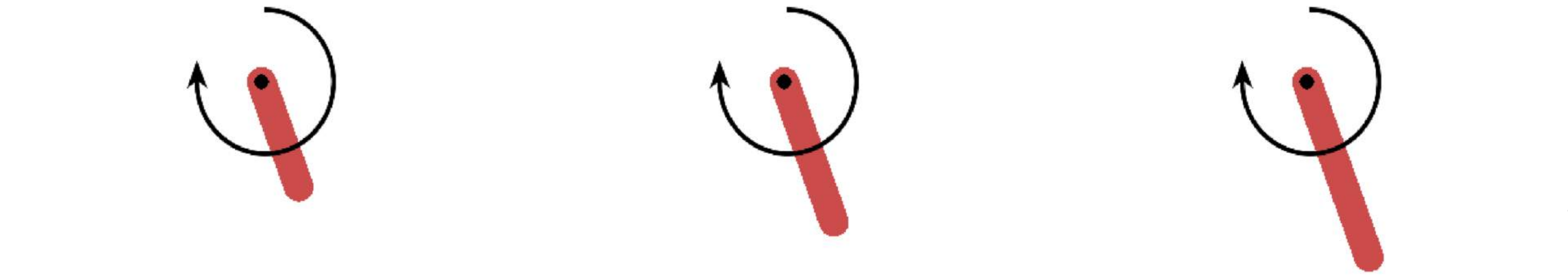} \label{fig:pendulum_example}}
\\
\subfigure[HalfCheetah with varying body masses]
{
\includegraphics[width=0.47\textwidth]{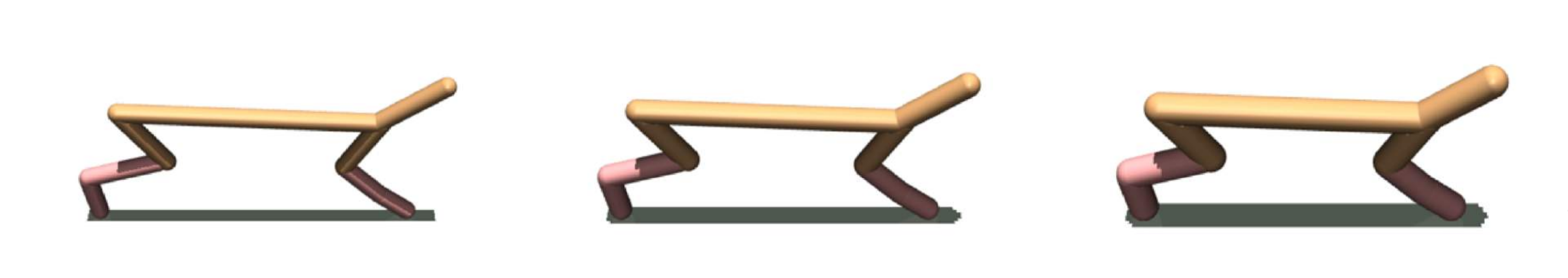}
\label{fig:halfCheetah_example}}
\\
\subfigure[Ant with varying body masses]
{
\includegraphics[width=0.47\textwidth]{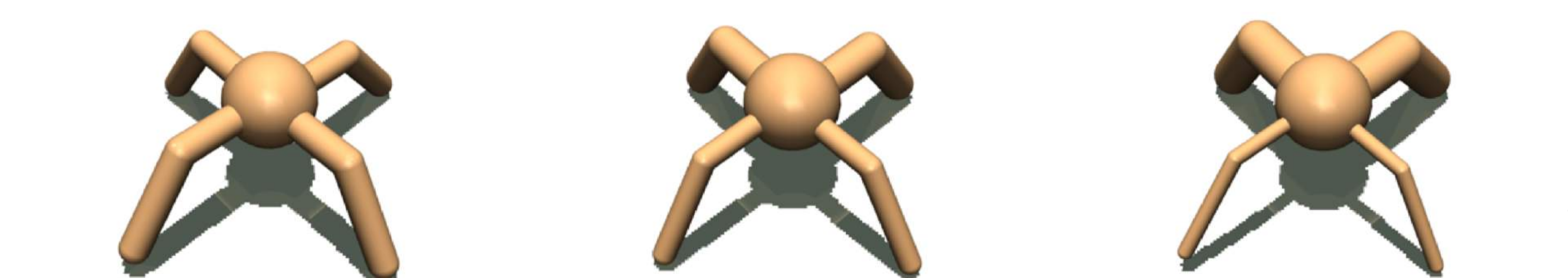}
\label{fig:ant_example}}
\caption{Examples from (a) CartPole, (b) Pendulum, (c) HalfCheetah, and (d) Ant. We change the transition dynamics of each environment by modifying its environment parameters.}
\label{fig:examples_environments}
\vspace{0.15in}
\end{figure}

However, it has been evidenced that model-based RL methods often struggle to generalize to an unseen environment with different transition dynamics \citep{nagabandi2018learning,nagabandi2018deep}.
For example, 
when using a vanilla model-based RL method for the simple CartPole task (see Figure~\ref{fig:examples_environments}),
even a minimal change in the pole mass leads to an inaccurate next-step prediction. {This, in turn, leads to poor planning, and eventually performance degradation, as demonstrated in our experiments (see Table~\ref{tbl:mbrl_test}).}
Such failure to take account of transition dynamics shift makes them unreliable for real-life deployment, which requires robustness to a myriad of changing environmental factors.

To improve the generalization capabilities of model-based RL methods, several strategies have been proposed,
including meta-learning \cite{nagabandi2018learning,nagabandi2018deep} and graph networks \citep{sanchez2018graph} 
(see Section~\ref{sec:related_work} for further details). 
A notable such example is the model-based meta-RL method suggested by \citet{nagabandi2018learning}, 
where a meta-learned prior model adapts to a recent trajectory segment, 
either by updating hidden state of a recurrent model \cite{duan2016rl}, or by updating model parameters via a small number of gradient steps \cite{finn2017model}.
However, it is questionable whether a simple gradient or hidden state updates would capture the rich contextual information that the environment offers. 
Instead, we argue that separating context encoding (i.e., capturing the contextual information)
and transition inference (i.e., predicting the next state conditioned on the captured information) can be more effective for learning the environment dynamics.

{\bf Contribution}. 
In this paper, we develop a context-aware dynamics model (CaDM) capable of generalizing across a distribution of environments with varying transition dynamics.
First, to capture the contextual information, 
we introduce a context encoder that produces a latent vector from a recent experience.
Then, by conditioning our forward dynamics model on this latent vector, 
we effectively perform an online adaptation to the unseen environment. 
We emphasize that the proposed CaDM can incorporate any dynamics model, e.g., fully-connected network or recurrent neural network, 
simply by conditioning the dynamics model on the encoder output.
The novel ingredient of CaDM is its loss function 
that forces the latent vector to be useful for predicting not only the next states (forward dynamics), but also the previous states (backward dynamics).
Moreover, we consider an additional regularization that encourages temporal consistency of the context latent vector: 
we force the context latent vector obtained in the current timestep to be useful for predictions made in the nearby future timesteps.
We discover that the proposed context latent vector aptly captures the rich contextual information of the environment, which enables fast adaptation to the changing dynamics (see Figure~\ref{fig:tsne_ours_context}). 
Finally, we explore the possibility of improving the generalization abilities of a model-free RL method \cite{schulman2017proximal} by giving the learned context latent vector as an additional input to the policy. 

We demonstrate the effectiveness of our method on a variety of simulated control tasks from OpenAI gym \cite{brockman2016openai} and MuJoCo physics engine \cite{todorov2012mujoco}. For evaluation, we measure the performance of model-based RL methods in a range of unseen (yet related) environments with different transition dynamics (see Figure~\ref{fig:examples_environments}).
In our experiments, 
CaDM significantly reduces the performance gap between training and test environments 
when compared to baselines, including ensemble methods \cite{chua2018deep} 
and model-based meta-RL methods \cite{nagabandi2018learning}.
Furthermore, we show that CaDM can also improve model-free RL methods:
it improves the generalization performance of proximal policy optimization (PPO; \citealt{schulman2017proximal}), compared to {other} context learning methods \citep{rakelly2019efficient,zhou2019environment} {for model-free RL.} 
We believe that our approach could be influential to other relevant topics, such as sim-to-real transfer \citep{peng2018sim}.

\section{Related Work} \label{sec:related_work}

{\bf Model-based reinforcement learning}.
In model-based RL, we learn a dynamics model that approximates the true environment dynamics.
Such a dynamics model can then be used for control by planning \cite{atkeson1997comparison,lenz2015deepmpc,finn2017deep}, 
or for improving the data-efficiency of model-free RL methods \cite{sutton1990integrated, gu2016continuous, janner2019trust}.
Recently,
by incorporating high-capacity function approximators like DNNs,
model-based RL has shown remarkable progress even in complex domains \cite{oh2015action,chua2018deep,hafner2018learning,hafner2019dream,schrittwieser2019mastering}.
For example,
\citet{chua2018deep} showed that an ensemble of probabilistic DNNs could reduce modeling errors and achieve the asymptotic performance on-par with model-free RL algorithms.
It also has been observed that model-based RL methods can be applied for solving high-dimensional visual domain tasks \cite{hafner2018learning,hafner2019dream,schrittwieser2019mastering}. 
However, developing a generalizable model-based RL method still remains a longstanding challenge.

\begin{figure*} [t] \centering
\subfigure[Forward prediction]
{
\includegraphics[width=0.31\textwidth]{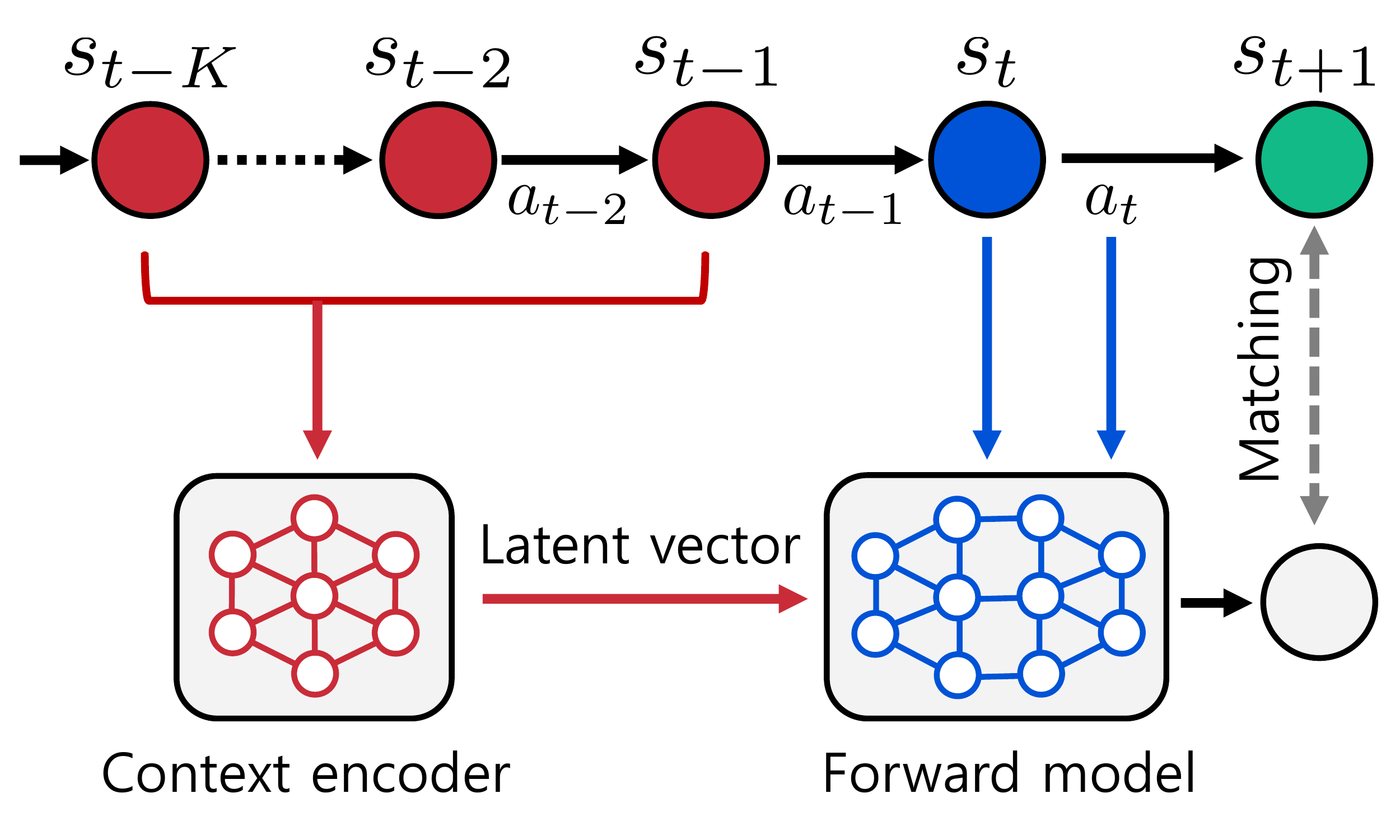} \label{fig:forward}} 
\,
\subfigure[Backward prediction]
{
\includegraphics[width=0.31\textwidth]{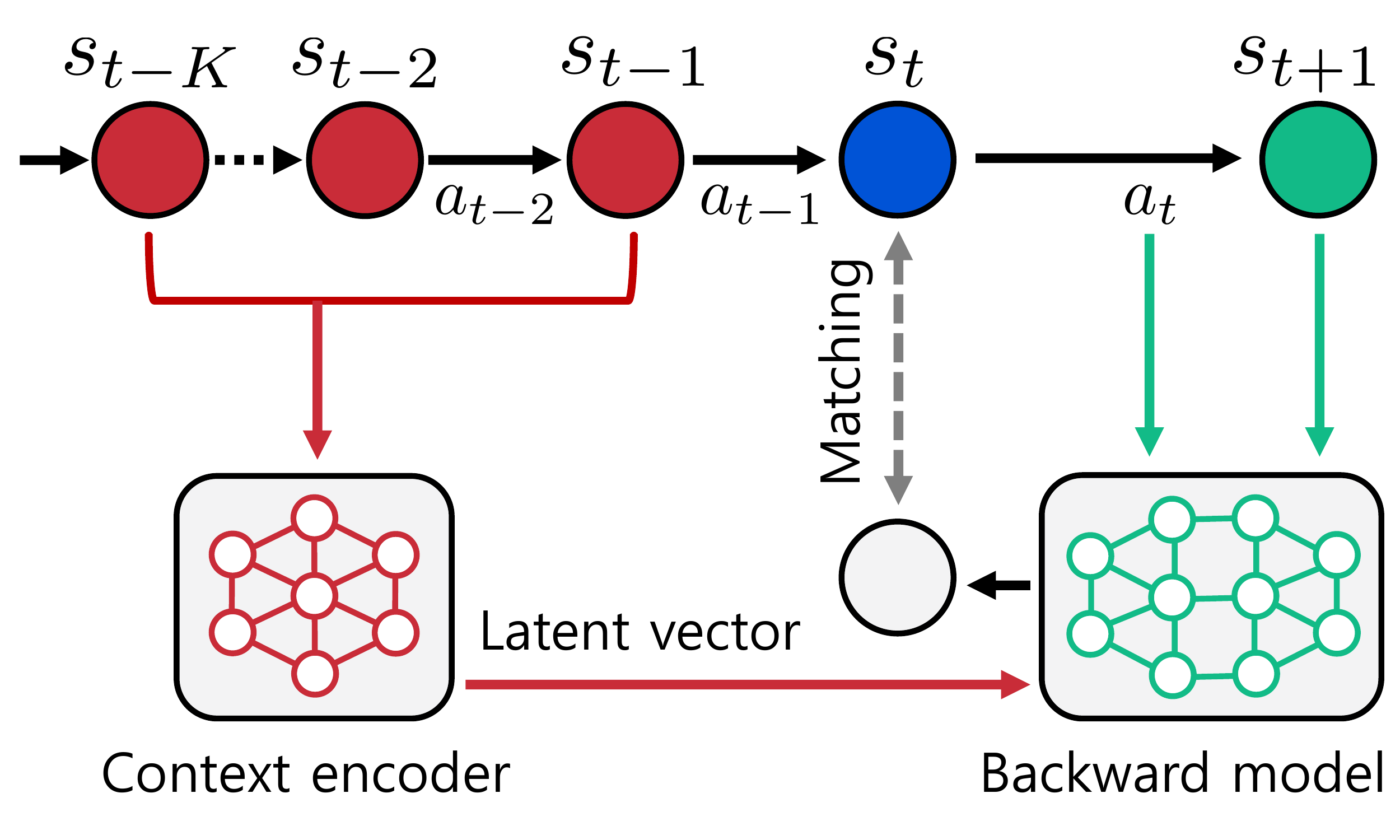} \label{fig:backward}}
\,
\subfigure[Future-step prediction]
{
\includegraphics[width=0.31\textwidth]{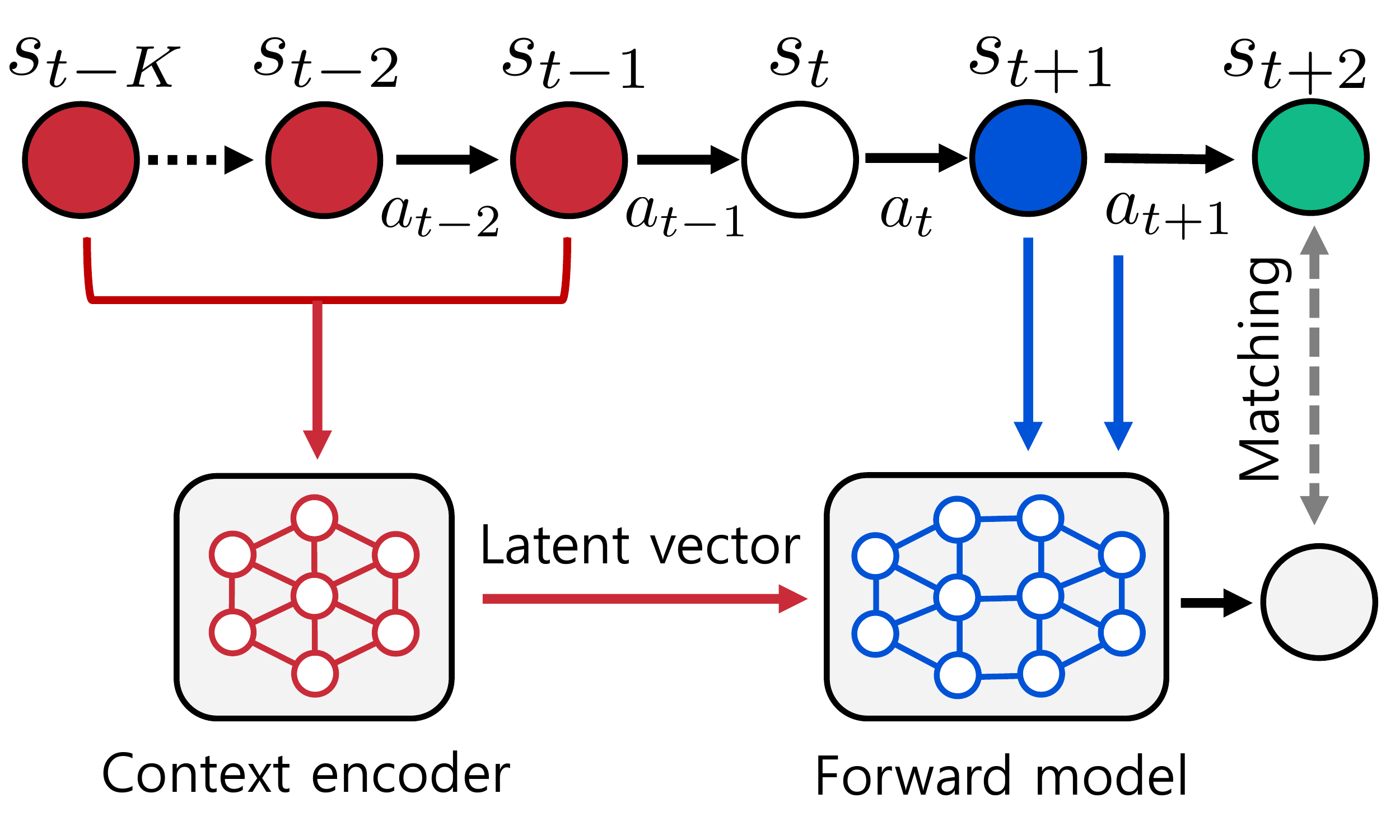} \label{fig:multi_step}}
\caption{Illustrations of our framework. 
We decompose the task of learning a global dynamics model into context encoding and transition inference. 
(a) Our dynamics model predicts the next state conditioned on the latent vector. 
(b) We introduce a backward dynamics model that predicts a previous state by utilizing a context latent vector. 
(c) We force the context latent vector to be temporally consistent by utilizing it for predictions in the future timesteps.}
\label{fig:overview}
\end{figure*}

{\bf Dynamics generalization in deep RL}.
Generalization of model-based RL has recently gained considerable attention in the research community. 
\citet{sanchez2018graph} proposed to model a dynamics model using a graph network,
and 
\citet{nagabandi2018learning,nagabandi2018deep} studied model-based meta-RL methods where the meta-learner learns to update the model according to dynamics changes.
Notably, \citet{nagabandi2018learning} proposed to adapt the dynamics model to recent trajectory segments, either by updating hidden representations of a recurrent model \cite{duan2016rl} or by updating model parameters via a small number of gradient steps \cite{finn2017model}.
However, it is questionable whether such methods capture the environment context properly, for they are overburdened with two tasks: adaptation and {next-state inference}. We instead propose to disentangle the two tasks by introducing a context encoder that specializes in the former, and a conditioned dynamics model that specializes in the latter. 

Several model-free RL methods, 
including adversarial policy training \cite{morimoto2001robust, pinto2017robust, rajeswaran2016epopt},
structured policy with graph neural network \cite{wang2018nervenet},
online system identification \cite{yu2017preparing, zhou2019environment},
and meta-learning \cite{finn2017model,rakelly2019efficient}, 
have been proposed to improve the generalization ability of RL agents across dynamics changes.
In particular, \citet{rakelly2019efficient} proposed a meta-RL method that adapts to a new environment by inferring latent context variables from a small number of trajectories. 
However, our method differs from this approach, in that we train our context encoder by utilizing it for dynamics prediction, as opposed to maximizing the expected returns.

\section{Problem Statement} \label{sec:problem}
We consider the standard RL framework where an agent interacts with its environment in discrete time.
Formally, we formulate our problem as a Markov Decision Process (MDP; \citealt{sutton2018reinforcement}) defined as a tuple $\left( \mathcal{S}, \mathcal{A}, p, r, \gamma, \rho_0\right)$,
with $\mathcal{S}$ denoting the state space, 
$\mathcal{A}$ the action space, 
$p\left(s^\prime|s,a\right)$ the transition dynamic,
$r\left(s,a\right)$ the reward function,
$\rho_0$ the initial state distribution,
and $\gamma \in [0,1)$ the discount factor.
The goal of RL is to find a policy that produces an action to take from each given state so as to maximize the expected return defined as the total accumulated reward. 
We tackle this problem in the context of model-based RL by learning a forward dynamics model $f$, which approximates the transition dynamics $p\left(s^\prime|s,a\right)$.

In order to address the problem of generalization, we further consider the distribution of MDPs, where the transition dynamics $p_c\left(s^\prime| s,a \right)$ varies according to a context $c$. For instance, a robot agent's transition dynamics may change when some of its parts malfunction due to unexpected damages. Our goal is to learn a generalizable forward dynamics model that is robust to such dynamics changes, i.e., approximating a distribution of transition dynamics.
Specifically, given a set of training environments with contexts sampled from $p_{\tt train}(c)$, we aim to learn a forward dynamics model that can produce accurate predictions for test environments with unseen contexts sampled from $p_{\tt test}(c)$.

\section{Context-aware Dynamics Model} \label{sec:main_method}

In this section, we propose a context-aware dynamics model (CaDM) that can generalize to unseen environments with varying transition dynamics.
Our scheme separates the task of reasoning about the environment dynamics into (a) encoding the dynamics-specific information into a latent vector (context encoding), and (b) predicting the next state conditioned on the latent vector (transition inference). To extract contextual information effectively,
we propose a novel loss function that encourages the context latent vector to be useful for various auxiliary prediction tasks. We also discuss how the learned context latent vector can help improve the generalization abilities of model-free RL methods.

\subsection{Context Encoder and Dynamics Model} \label{sec:framework}

To capture the true unknown context $c$ of the environment, 
we introduce a context encoder $g$ parameterized by $\phi$, 
which produces a latent vector $z_t = g\left(\tau^{\tt P}_{t,K}; \phi\right)$ given $K$ past transitions $\tau^{\tt P}_{t,K} = \{ (s_{t-K},a_{t-K}) \cdots (s_{t-1},a_{t-1})\}$. 
The intuition is that the true context of the underlying MDP can be captured from recent experiences. 
Note that similar ideas have been explored by \citet{nagabandi2018learning,rakelly2019efficient,zhou2019environment}.

We introduce two dynamics models: (a) forward dynamics model $f\left(s_{t+1}|s_t,a_t,z_t;\theta \right)$ that predicts the {\em next state} given the current state, current action and current context latent vector (see Figure~\ref{fig:forward}), and (b) backward dynamics model $b\left(s_{t}|s_{t+1},a_t,z_t;\psi\right)$ that predicts the {\em current state} given the next state, current action and current context latent vector (see Figure~\ref{fig:backward}).
Under the assumption that MDPs with similar contexts will behave similarly \cite{modi2018markov}, our method can generalize to “nearby” unseen dynamics by capturing context information from training environments.
Forward and backward dynamics models are parameterized by $\theta$ and $\psi$, respectively, and are architecture-agnostic, i.e., any existing model architecture can be used.
Indeed, we experimentally confirm that the performance of stochastic models \cite{chua2018deep} significantly improves when combined with our method (see Section~\ref{sec:exp_model_based}).


We optimize the context encoder and the dynamics models by minimizing the following loss function:
\begin{align} \label{eq:main_obj}
&\mathcal{L}^{\tt pred} =  \mathbb{E}_{(\tau^{\tt F}_{t,M},\tau^{\tt P}_{t,K})\sim \mathcal{B}}\Big[ \mathcal{L}^{\tt pred}_{\tt forward} + \beta \mathcal{L}^{\tt pred}_{\tt backward}\Big], \\
&\mathcal{L}^{\tt pred}_{\tt forward} = - \frac{1}{M} \sum_{i=t}^{t+M-1} \log f\left(s_{i+1}|s_i,a_i,g\left(\tau^{\tt P}_{t,K}; \phi\right);\theta \right), \notag\\
&\mathcal{L}^{\tt pred}_{\tt backward} = - \frac{1}{M} \sum_{i=t}^{t+M-1} \log b\left(s_{i}|s_{i+1},a_i,g\left(\tau^{\tt P}_{t,K}; \phi\right);\psi \right), \notag 
\end{align}
where 
$\tau^{\tt F}_{t,M} = \{\left(s_{t}, a_{t}\right), \cdots,\left(s_{t+M}, a_{t+M}\right) \}$ is the future trajectory segment,
$\mathcal{B}=\{(\tau^{\tt F}_{t,M}, \tau^{\tt P}_{t,K})\}$ the training dataset,
$\beta>0$ the penalty parameter,
and $M>0$ the number of future samples.
Here, we remark that both of our dynamics models share the same context latent vector $g\left(\tau^{\tt P}_{t,K}; \phi\right)$ as an additional input. Our intuition is that dynamics-specific context (e.g., environment parameters) must be useful for predicting both forward and backward transitions. 
Here, our motivation is that predicting backward transitions can also capture contextual information while mitigating the risk of overly focusing on predicting only the "seen" forward dynamics.
Additionally,
in order to handle long-horizon tasks,
we encourage the context latent vector to be useful for predictions multiple timesteps into the future (see Figure~\ref{fig:multi_step}).
Overall, the various prediction tasks introduced (i.e., future-step forward and backward predictions) help the context latent vector to be temporally consistent.

Using a model-based RL method to identify the environment dynamics was also studied by \citet{zhou2019environment}. 
However, the dynamics model in their method is not for producing accurate predictions, and
their main focus is improving the generalizability of model-free RL methods.
On the other hand, we focus on learning an accurate and generalizable dynamics model.

\subsection{Additional Training Details} \label{sec:additional_tech}

{\bf History of transitions}. In our experiments, we use the state difference $\Delta s_{t} = s_{t+1}-s_{t}$ instead of the raw state $s_{t}$ as an input to the context encoder:
\begin{align*}
  \tau^{\tt P}_{t,K} = \{\left(\Delta s_{t-K}, a_{t-K}\right), \cdots,\left(\Delta s_{t-1}, a_{t-1}\right) \}.  
\end{align*}
We found that this simple technique provides further performance improvement.

\begin{algorithm}[t]
\caption{Training context-aware dynamics model} \label{alg:training}
\begin{algorithmic}[1]
\STATE {\bf Inputs}: the number of past observations $K$ and future observations $M$, learning rate $\alpha$, and batch size $B$.

\vspace{0.05in}
\hrule
\vspace{0.05in}
\STATE Initialize parameters of forward dynamics model $\theta$, backward dynamics model $\psi$, context encoder $\phi$. 
\STATE Initialize dataset $\mathcal{B} \leftarrow \emptyset$.
\FOR{each iteration}
\STATE {{\textsc{// Collect training samples}}}
\STATE Sample $c\sim p_{\tt seen}\left(c\right)$. 
\FOR{$t=1$ {\bfseries to} TaskHorizon}  
\STATE Get context latent vector $z_t= g\left(\tau^{\tt P}_{t,K}; \phi\right)$.
\STATE Collect samples $\{(s_t,a_t,s_{t+1},r_t,\tau^{\tt P}_{t,K})\}$ from the environment with transition dynamics $p_c$ using the planning algorithm described in Section~\ref{sec:additional_tech}.
\STATE Update $\mathcal{B} \leftarrow \mathcal{B}\cup \{(s_t,a_t,s_{t+1},r_t,\tau^{\tt P}_{t,K})\}$.
\ENDFOR
\STATE {{\textsc{// Update dynamics models and encoder}}}
\FOR{$i=1$ {\bfseries to} $B$}
\STATE Sample $\tau^{\tt P}_{i,K}, \tau^{\tt F}_{i,M} \sim \mathcal{B}$.
\STATE Get context latent vector $z_i= g\left(\tau^{\tt P}_{i,K}; \phi\right)$.
\STATE $\mathcal{L}_{i}^{\tt pred} \leftarrow \mathcal{L}^{\tt pred} \left(\tau^{\tt F}_{i,M}, z_i,\theta,\psi \right)$ in \eqref{eq:main_obj}.
\ENDFOR
\STATE Update
$\theta \leftarrow \theta - \alpha \nabla_{\theta}
\frac{1}{B} \sum\limits_{i=1}^{B} \mathcal{L}^{\tt pred}_{i}$
\STATE Update 
$\psi \leftarrow \psi - \alpha \nabla_{\psi}
\frac{1}{B} \sum\limits_{i=1}^{B} \mathcal{L}^{\tt pred}_{i}$
\STATE Update
$\phi \leftarrow \phi - \alpha \nabla_{\phi}
\frac{1}{B} \sum\limits_{i=1}^{B} \mathcal{L}^{\tt pred}_{i}$.
\ENDFOR
\end{algorithmic}
\end{algorithm}

{\bf Planning Algorithm}.
We use a model predictive control (MPC; \citealt{garcia1989model}) to select actions based on the forward dynamics prediction. 
Specifically, we use the cross entropy method (CEM; \citealt{botev2013cross}), where $N$ candidate action sequences are iteratively sampled from a candidate distribution, which is adjusted based on best performing action samples.
Then, we use the mean of adjusted candidate distribution as action and re-plan at every timestep.


\subsection{Combination with Model-free RL}  \label{sec:model-free}

As a natural extension, we show that the learned context latent vector can be utilized for improving the generalization abilities of model-free RL methods.
Here, we briefly mention that various model-free RL methods are also known to suffer from poor generalization \cite{yu2017preparing,packer2018assessing,zhou2019environment}.
One of the major research directions for dynamics generalization in model-free RL is developing a context-conditional policy. Specifically, \citet{yu2017preparing,rakelly2019efficient,zhou2019environment} showed that a policy can be more robust to dynamics changes when it takes context information as an additional input.
Motivated by this, we consider a context-conditioned policy $\pi \left(a_t|s_t,g\left(\tau^{\tt P}_{t,K}; \phi\right)\right)$ conditioned on our learned context latent vector. 
Compared to existing context-conditional policies \cite{rakelly2019efficient,zhou2019environment}, 
our model-free RL method demonstrates superior generalization abilities (see Table~\ref{tbl:mfrl_test}).

\section{Experiments} \label{sec:exp}
In this section, we evaluate the performance of our CaDM method to answer the following questions:
\begin{itemize}[topsep=1pt]
    \item [$\bullet$] Is CaDM more robust to dynamics changes compared to other model-based RL methods (see Table~\ref{tbl:mbrl_test})?
    \item [$\bullet$] Can CaDM be combined with model-free RL methods to improve their generalization abilities (see Table~\ref{tbl:mfrl_test})?
    \item [$\bullet$] Does the proposed prediction loss \eqref{eq:main_obj} improve the test performance (see Figure~\ref{fig:mbrl_ablation_test})?
    \item [$\bullet$] Can CaDM make accurate predictions (see Figure~\ref{fig:prediction_error} and Figure~\ref{fig:modelling error})?
    \item [$\bullet$] Does our context encoder extract meaningful contextual information (see Figure~\ref{fig:tsne_ours_context})?
\end{itemize}

\subsection{Setups}

{\bf Environments}. 
We demonstrate the effectiveness of our proposed method on simulated robots (i.e., HalfCheetah, Ant, CrippledHalfCheetah, and SlimHumanoid) using the MuJoCo physics engine \cite{todorov2012mujoco} and classic control tasks (i.e., CartPole and Pendulum) from OpenAI Gym \cite{brockman2016openai}.
The goal of CartPole is to prevent the pole from falling over by pushing the cart left and right, while that of Pendulum is to swing up the pendulum and keep it in the upright position.
As for HalfCheetah, Ant, and SlimHumanoid, the goal is to move forward as fast as possible, while keeping the control cost minimal.
As in \citet{packer2018assessing, zhou2019environment}, we modify the environment parameters (e.g., mass, length, damping) that characterize the transition dynamics (see Figure~\ref{fig:examples_environments}).
In the case of CrippledHalfCheetah, one of the actuators is randomly crippled to change the transition dynamics.

For both training and testing, we sample environment parameters at the beginning of each episode. During training, we randomly sample environment parameters from a predefined training range. At test time, we measure each model's performance in unseen environments characterized by parameters outside the training range. In order to utilize model-predictive control (MPC), we assume that the reward function is known, as in \cite{chua2018deep, nagabandi2018learning}. 

\begin{figure} [t] \centering
\includegraphics[width=0.5\textwidth]{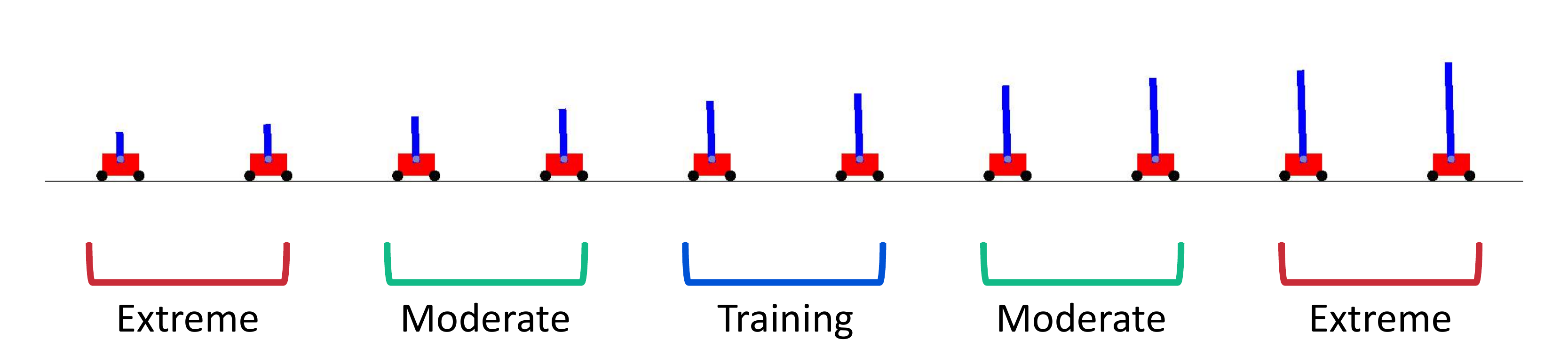}
\vspace{-0.1in}
\caption{Illustration of training and test parameter ranges.}
\vspace{0.1in}
\label{fig:test_range}
\end{figure}

Note that generalization performance is measured in two different regimes: moderate and extreme, where the former draw environment parameters from a closer range to the training range, compared to the latter (see Figure~\ref{fig:test_range}). For all our experiments, we select the model with the highest average return during training and report the test performance.
We report mean and standard deviation across five runs.
Due to space limitations, we provide more experimental details in the supplementary material.

{\bf Baselines and our method}. We consider the following model-based RL methods as baselines:
\begin{itemize}[topsep=1pt]
    \item [$\bullet$] Vanilla dynamics model (Vanilla DM): Dynamics model trained to minimize the standard one-step forward prediction loss. The model is fixed during test time (i.e., no adaptation).
    \item [$\bullet$] Stacked dynamics model (Stacked DM): Vanilla dynamics model which takes the past $K \in \{5,10,15\}$ transitions as an additional input. A comparison with this model evaluates the benefit of introducing a context latent vector.
    \item [$\bullet$] Gradient-Based Adaptive Learner (GrBAL; \citealt{nagabandi2018learning}): Model-based meta-RL method which trains a dynamics model by optimizing an adaptation meta-objective. At test time, the meta-learned prior dynamics model adapts to a recent trajectory segment by taking gradient steps.\footnotemark 
    \item [$\bullet$] Recurrence-Based Adaptive Learner (ReBAL; \citealt{nagabandi2018learning}): Model-based meta-RL method similar to GrBAL. However, instead of taking gradient steps, ReBAL uses a recurrent model that learns its own update rule, i.e. updating its hidden state.\footnotemark[\value{footnote}]
    \item [$\bullet$] Probabilistic ensemble dynamics model (PE-TS; \citealt{chua2018deep}): An ensemble of probabilistic dynamics models designed to incorporate both environment stochasticity and subjective uncertainty into the model.
\end{itemize}

We combine CaDM with two baseline model-based RL methods, Vanilla DM and PE-TS.: the results of ReBAL + CaDM can also be found in the supplementary material, where it underperforms Vanilla + CaDM or PE-TS + CaDM overall.
We remark that the proposed method can be applied to any model-based RL methods because we separate a context encoder from forward dynamics models, as shown in Figure~\ref{fig:overview}. Due to space limitation, we provide details about model architecture and hyperparameters in the supplementary material.
Our code is available at \url{https://github.com/younggyoseo/CaDM}.

\footnotetext{We used a reference implementation publicly available at \url{https://github.com/iclavera/learning_to_adapt}.}

\begin{figure*} [t!] \centering
\vspace{-0.08in}
\subfigure[HalfCheetah]
{
\includegraphics[width=0.23\textwidth]{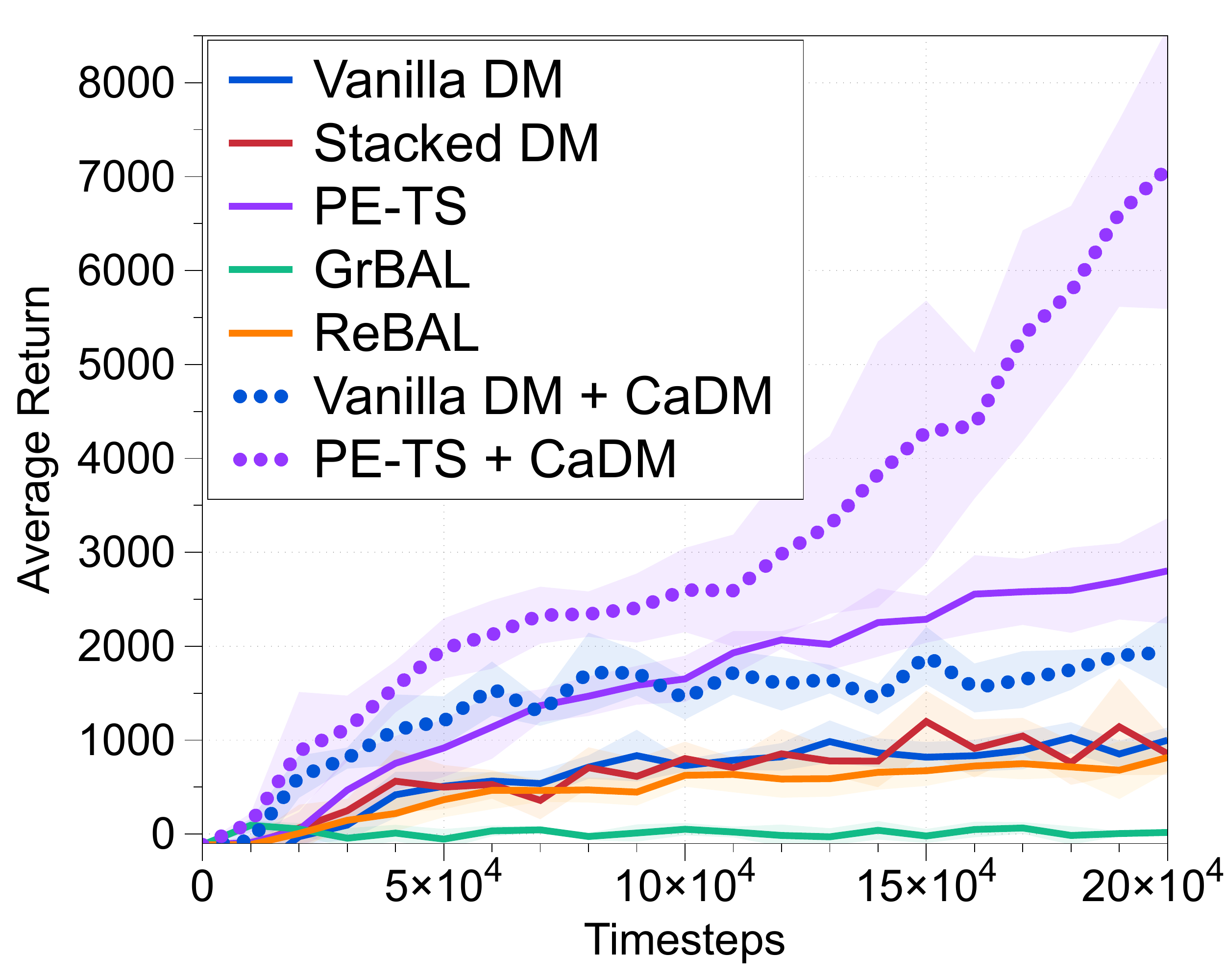} \label{fig:mbrl_halfcheetah_moderate}} 
\hfill
\subfigure[Ant]
{
\includegraphics[width=0.23\textwidth]{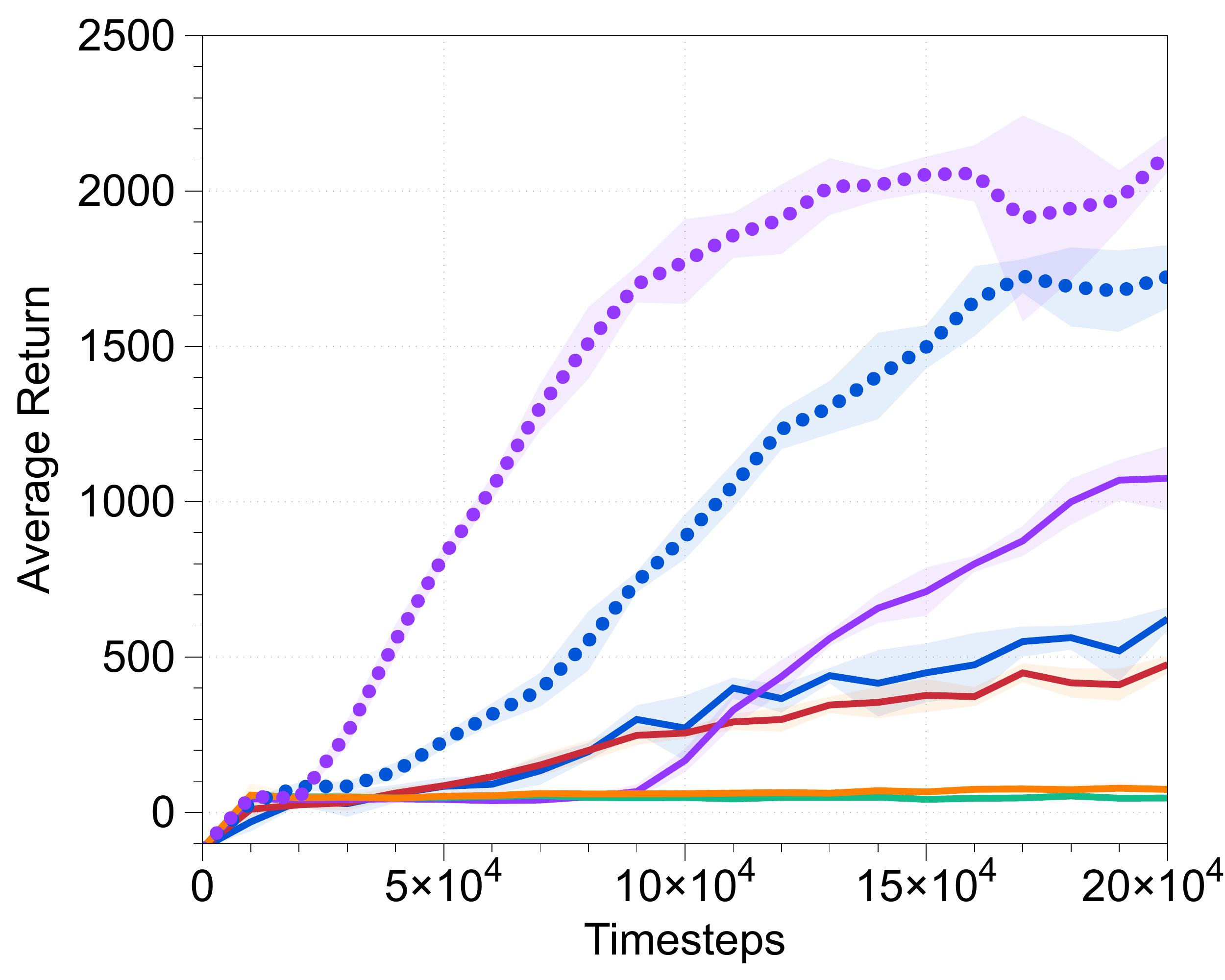} \label{fig:mbrl_ant_moderate}}
\hfill
\subfigure[CrippledHalfCheetah]
{
\includegraphics[width=0.23\textwidth]{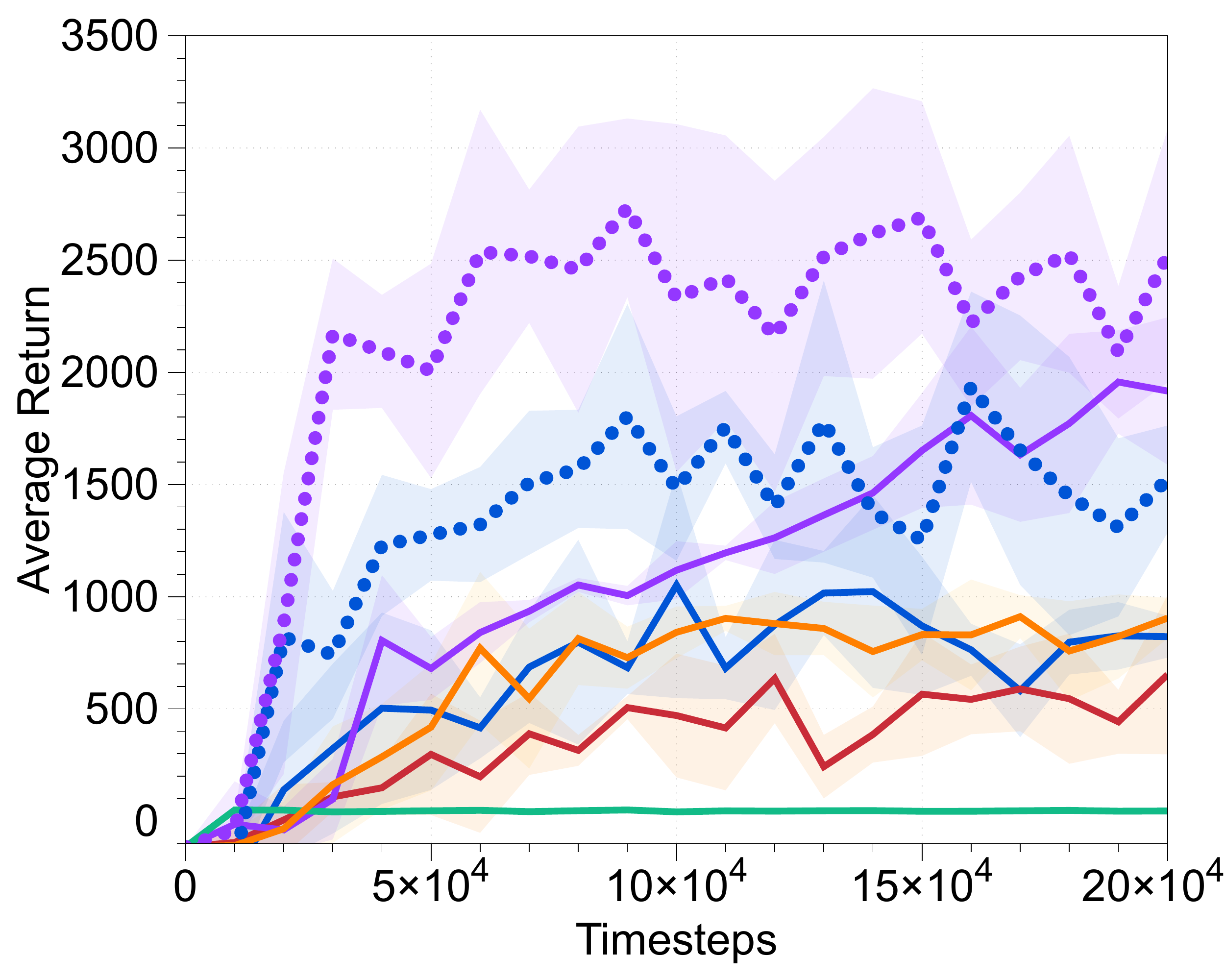}
\label{fig:mbrl_crippled_halfcheetah_moderate}}
\hfill
\subfigure[SlimHumanoid]
{
\includegraphics[width=0.23\textwidth]{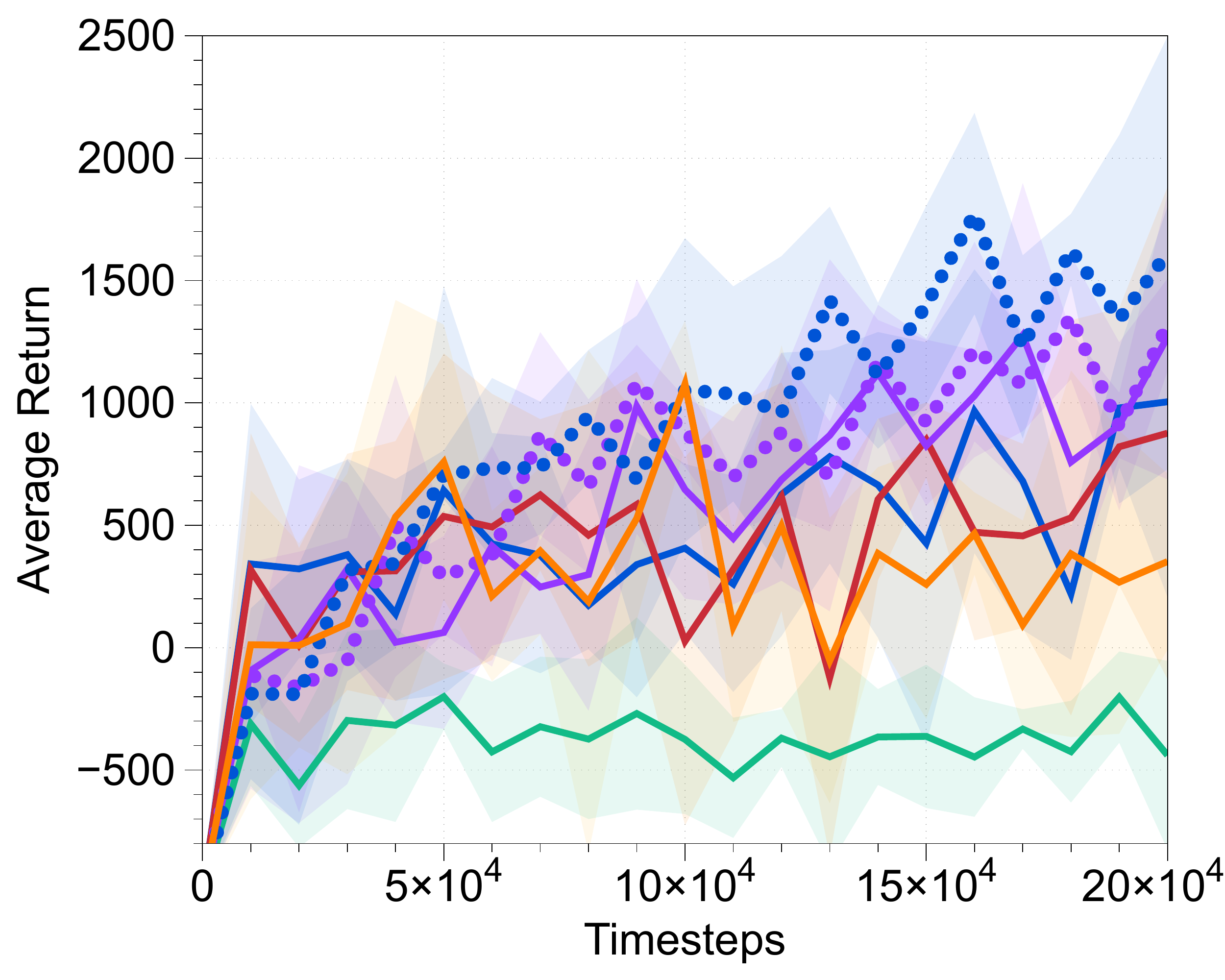}
\label{fig:mbrl_slim_humanoid_moderate}}
\hfill
\vspace{-0.1in}
\caption{The average returns of trained dynamics models on unseen (moderate) environments. The results show the mean and standard deviation of returns averaged over five runs. The full figures of all environments are in the supplementary material.}
\vspace{-0.07in}
\label{fig:mbrl_moderate}
\end{figure*}


\begin{table*}[t!]
\centering
\begin{adjustbox}{max width=\textwidth}
\begin{tabular}{
c
c@{\hspace{0.2cm}}
c
c
c
c@{\hspace{0.2cm}}
c
c
c}
\toprule
  &
  &
  \multicolumn{3}{c}{CartPole} &
  &
  \multicolumn{3}{c}{Pendulum} \\ 
  \cline{3-5} \cline{7-9}
\multicolumn{1}{c}{} &
  &
  Training &
  Test (moderate) &
  Test (extreme) &
  &
  Training &
  Test (moderate) &
  Test (extreme)  \\ \midrule
  Vanilla DM &
  &
  176.6$\pm$ \scriptsize{10.2} &
  124.7$\pm$ \scriptsize{8.1} &
  105.9$\pm$ \scriptsize{3.7} &
  &
  -419.2$\pm$ \scriptsize{95.1} &
  -928.4$\pm$ \scriptsize{56.8} &
  -1170.6$\pm$ \scriptsize{51.3} \\
  Stacked DM &
  &
  176.9$\pm$ \scriptsize{6.3} &
  131.9$\pm$ \scriptsize{6.0} &
  106.6$\pm$ \scriptsize{11.0} &
  &
  -1158.7$\pm$ \scriptsize{124.7} &
  -1308.0$\pm$ \scriptsize{49.6} &
  -1288.3$\pm$ \scriptsize{65.3} \\
  GrBAL &
  &
  118.1$\pm$ \scriptsize{40.2} &
  87.5$\pm$ \scriptsize{35.1} &
  81.6$\pm$ \scriptsize{21.3} &
  &
  -726.8$\pm$ \scriptsize{322.6} &
  -1027.7$\pm$ \scriptsize{68.1} &
  -1048.1$\pm$ \scriptsize{111.8} \\
  ReBAL &
  &
  105.7$\pm$ \scriptsize{45.8} &
  48.4$\pm$ \scriptsize{34.5} &
  54.3$\pm$ \scriptsize{37.0} &
  &
  -460.4$\pm$ \scriptsize{74.5} &
  -860.8$\pm$ \scriptsize{140.3} &
  -1026.4$\pm$ \scriptsize{51.5} \\
  PE-TS &
  &
  194.3$\pm$ \scriptsize{4.1} &
  171.3$\pm$ \scriptsize{18.5} &
  142.9$\pm$ \scriptsize{23.6} &
  &
  -577.2$\pm$ \scriptsize{198.1} &
  -985.7$\pm$ \scriptsize{64.2} &
  -1221.1$\pm$ \scriptsize{31.0} \\ \midrule
  Vanilla + CaDM &
  &
  185.7$\pm$ \scriptsize{7.7} &
  154.3$\pm$ \scriptsize{13.2} &
  118.6$\pm$ \scriptsize{6.6} &
  &
  \textbf{-415.2}$\pm$ \scriptsize{60.4} &
  \textbf{-593.7}$\pm$ \scriptsize{48.6} &
  \textbf{-967.9}$\pm$ \scriptsize{83.5} \\
  PE-TS + CaDM &
  &
  \textbf{196.1}$\pm$ \scriptsize{4.8} &
  \textbf{187.3}$\pm$ \scriptsize{11.2} &
  \textbf{149.4}$\pm$ \scriptsize{26.2} &
  &
  -537.0$\pm$ \scriptsize{114.6} &
  -705.5$\pm$ \scriptsize{41.7} &
  -1098.4$\pm$ \scriptsize{47.1} \\
\midrule \midrule
  &
  &
  \multicolumn{3}{c}{Half-cheetah} &
  &
  \multicolumn{3}{c}{Ant} \\ 
  \cline{3-5} \cline{7-9}
\multicolumn{1}{c}{} &
  &
  Training &
  Test (moderate) &
  Test (extreme) &
  &
  Training &
  Test (moderate) &
  Test (extreme)  \\ \midrule
  Vanilla DM &
  &
  1560.7$\pm$ \scriptsize{453.1} &
  1026.7$\pm$ \scriptsize{164.7} &
  686.7$\pm$ \scriptsize{189.4} &
  &
  646.4$\pm$ \scriptsize{89.0} &
  520.0$\pm$ \scriptsize{97.6} &
  385.8$\pm$ \scriptsize{85.2} \\
  Stacked DM &
  &
  1301.4$\pm$ \scriptsize{310.5} &
  761.1$\pm$ \scriptsize{236.6} &
  661.5$\pm$ \scriptsize{220.5} &
  &
  492.3$\pm$ \scriptsize{68.7} &
  417.1$\pm$ \scriptsize{46.8} &
  338.9$\pm$ \scriptsize{51.5} \\
  GrBAL &
  &
  117.0$\pm$ \scriptsize{88.7} &
  -43.7$\pm$ \scriptsize{106.9} &
  -94.5$\pm$ \scriptsize{141.3} &
  &
  55.0$\pm$ \scriptsize{10.0} &
  46.5$\pm$ \scriptsize{6.5} &
  42.9$\pm$ \scriptsize{3.8} \\
  ReBAL &
  &
  1086.7$\pm$ \scriptsize{90.0} &
  657.5$\pm$ \scriptsize{184.9} &
  396.6$\pm$ \scriptsize{188.5} &
  &
  100.1$\pm$ \scriptsize{12.3} &
  73.1$\pm$ \scriptsize{15.5} &
  53.0$\pm$ \scriptsize{17.2} \\
  PE-TS &
  &
  4347.1$\pm$ \scriptsize{300.9} &
  2019.6$\pm$ \scriptsize{274.8} &
  1422.3$\pm$ \scriptsize{162.8} &
  &
  1183.3$\pm$ \scriptsize{51.1} &
  1075.1$\pm$ \scriptsize{103.6} &
  856.6$\pm$ \scriptsize{66.5} \\ \midrule
\textbf{}
  Vanilla + CaDM &
  &
  3536.5$\pm$ \scriptsize{641.7} &
  1556.1$\pm$ \scriptsize{260.6} &
  1264.5$\pm$ \scriptsize{228.7} &
  &
  1851.0$\pm$ \scriptsize{113.7} &
  1315.7$\pm$ \scriptsize{45.5} &
  821.4$\pm$ \scriptsize{113.5} \\
  PE-TS + CaDM &
  &
  \textbf{8264.0}$\pm$ \scriptsize{1374.0} &
  \textbf{7087.2}$\pm$ \scriptsize{1495.6} &
  \textbf{4661.8}$\pm$ \scriptsize{783.9} &
  &
  \textbf{2848.4}$\pm$ \scriptsize{61.9} &
  \textbf{2121.0}$\pm$ \scriptsize{60.4} &
  \textbf{1200.7}$\pm$ \scriptsize{21.8} \\ 
\midrule \midrule
  &
  &
  \multicolumn{3}{c}{CrippledHalfCheetah} &
  &
  \multicolumn{3}{c}{SlimHumanoid} \\ 
  \cline{3-5} \cline{7-9}
\multicolumn{1}{c}{} &
  &
  Training &
  Test (moderate) &
  Test (extreme) &
  &
  Training &
  Test (moderate) &
  Test (extreme)  \\ \midrule
  Vanilla DM &
  &
  1005.1$\pm$ \scriptsize{429.0} &
  870.0$\pm$ \scriptsize{308.0} &
  577.3$\pm$ \scriptsize{76.5} &
  &
  1119.8$\pm$ \scriptsize{317.6} &
  1004.4$\pm$ \scriptsize{798.2} &
  1155.5$\pm$ \scriptsize{556.9} \\
  Stacked DM &
  &
  630.6$\pm$ \scriptsize{211.3} &
  545.1$\pm$ \scriptsize{289.8} &
  417.9$\pm$ \scriptsize{145.8} &
  &
  1057.4$\pm$ \scriptsize{547.5} &
  876.2$\pm$ \scriptsize{1005.2} &
  651.8$\pm$ \scriptsize{449.9} \\
  GrBAL &
  &
  151.9$\pm$ \scriptsize{122.7} &
  -9.2$\pm$ \scriptsize{17.1} &
  16.6$\pm$ \scriptsize{23.0} &
  &
  -62.6$\pm$ \scriptsize{233.1} &
  -562.8$\pm$ \scriptsize{253.5} &
  -398.6$\pm$ \scriptsize{177.2} \\
  ReBAL &
  &
  701.7$\pm$ \scriptsize{119.7} &
  904.5$\pm$ \scriptsize{90.7} &
  833.0$\pm$ \scriptsize{118.0} &
  &
  1205.8$\pm$ \scriptsize{546.8} &
  85.8$\pm$ \scriptsize{388.9} &
  108.7$\pm$ \scriptsize{357.6} \\
  PE-TS &
  &
  1846.8$\pm$ \scriptsize{380.7} &
  1916.5$\pm$ \scriptsize{328.2} &
  1227.6$\pm$ \scriptsize{35.2} &
  &
  1339.6$\pm$ \scriptsize{524.0} &
  758.6$\pm$ \scriptsize{528.8} &
  810.4$\pm$ \scriptsize{363.4} \\ \midrule
\textbf{}
  Vanilla + CaDM &
  &
  2435.1$\pm$ \scriptsize{880.4} &
  1375.3$\pm$ \scriptsize{290.6} &
  966.9$\pm$ \scriptsize{89.4} &
  &
  \textbf{1758.2}$\pm$ \scriptsize{459.1} &
  \textbf{1228.9}$\pm$ \scriptsize{374.0} &
  \textbf{1487.9}$\pm$ \scriptsize{339.0} \\
  PE-TS + CaDM &
  &
  \textbf{3294.9}$\pm$ \scriptsize{733.9} &
  \textbf{2618.7}$\pm$ \scriptsize{647.1} &
  \textbf{1294.2}$\pm$ \scriptsize{214.9} &
  &
  1371.9$\pm$ \scriptsize{400.0} &
  903.7$\pm$ \scriptsize{343.9} &
  814.5$\pm$ \scriptsize{274.8} \\
  \bottomrule
\end{tabular}
\end{adjustbox}
\vspace{-0.1in}
\caption{The performance (average returns) of trained dynamics models on various control tasks. The transition dynamics of environments are changing in both training and test environments.
The results show the mean and standard deviation of returns averaged over five runs.}
\label{tbl:mbrl_test}
\vspace{-0.15in}
\end{table*}

\begin{figure*} [t] \centering
\vspace{-0.08in}
\subfigure[HalfCheetah]
{
\includegraphics[width=0.23\textwidth]{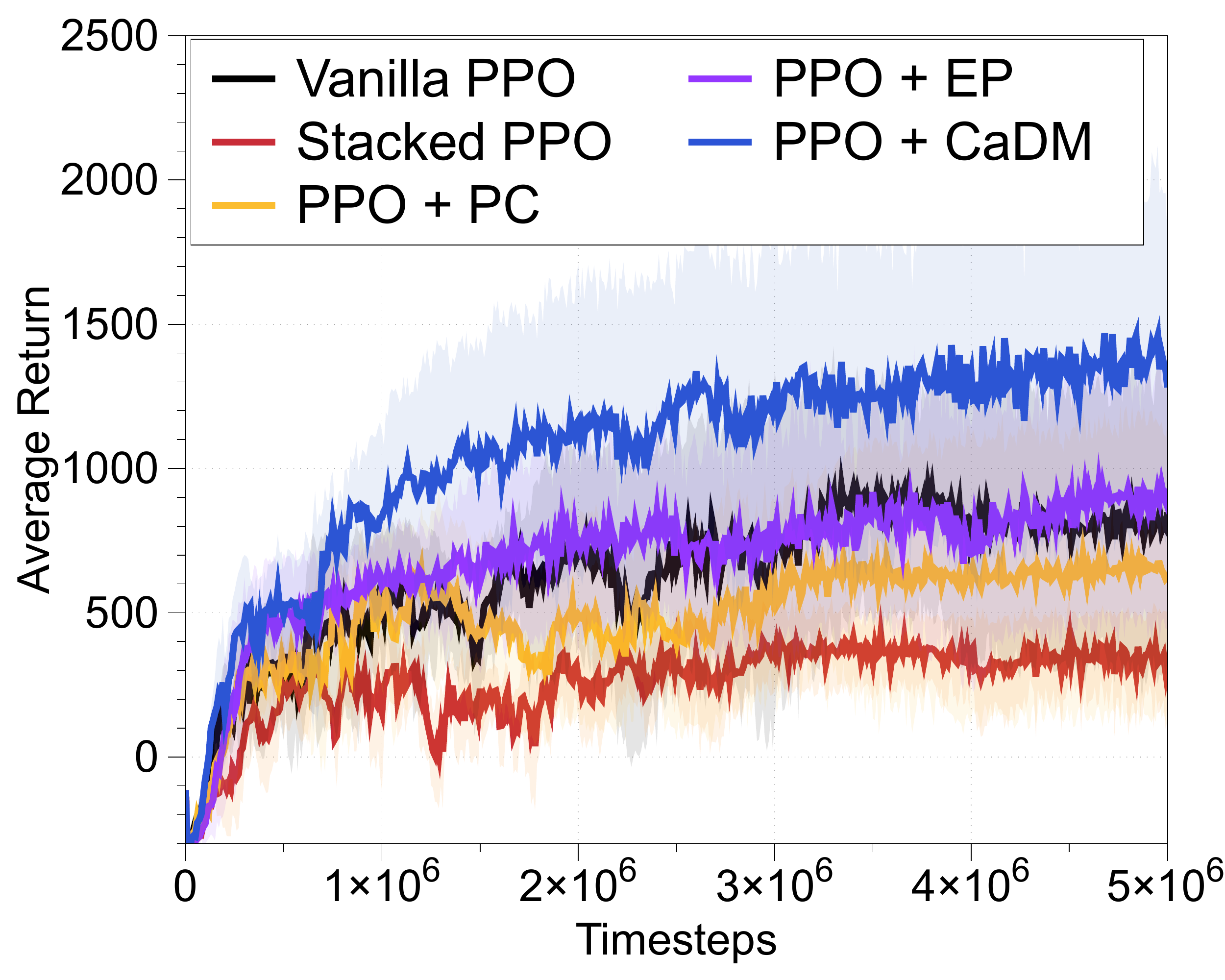} \label{fig:mfrl_halfcheetah_moderate}} 
\hfill
\subfigure[Ant]
{
\includegraphics[width=0.23\textwidth]{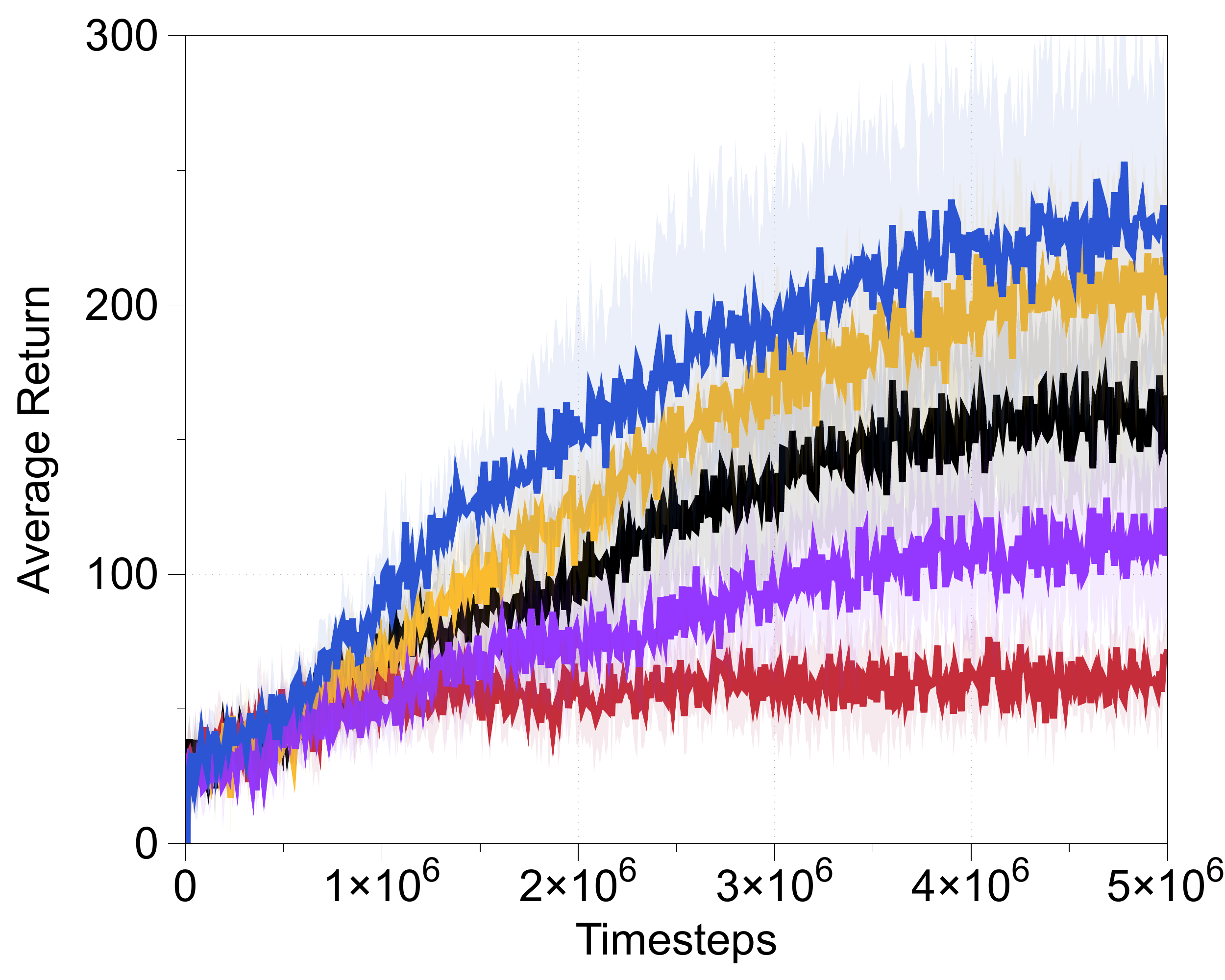} \label{fig:mfrl_ant_moderate}}
\hfill
\subfigure[CrippledHalfCheetah]
{
\includegraphics[width=0.23\textwidth]{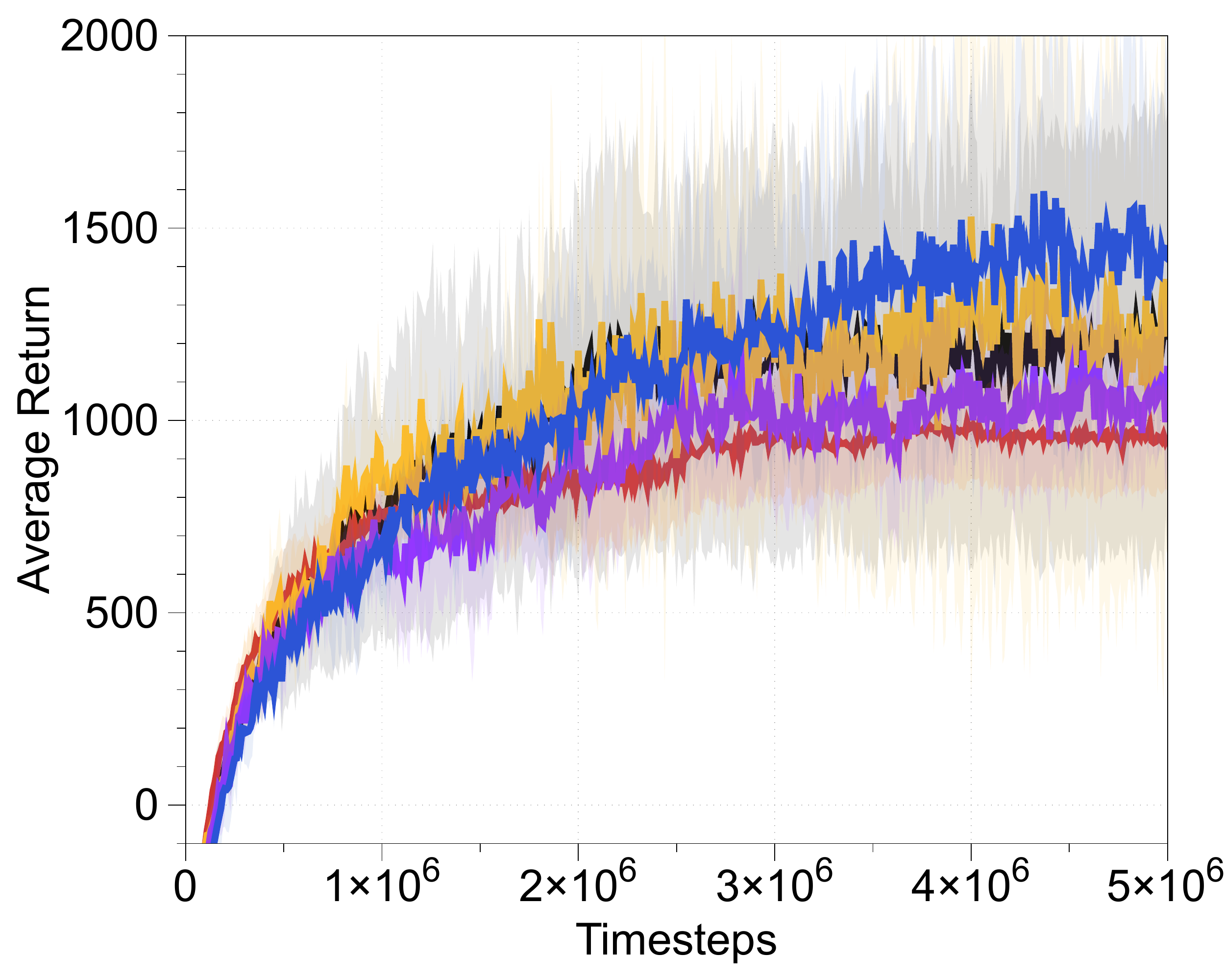}
\label{fig:mfrl_crippled_halfcheetah_moderate}}
\hfill
\subfigure[SlimHumanoid]
{
\includegraphics[width=0.23\textwidth]{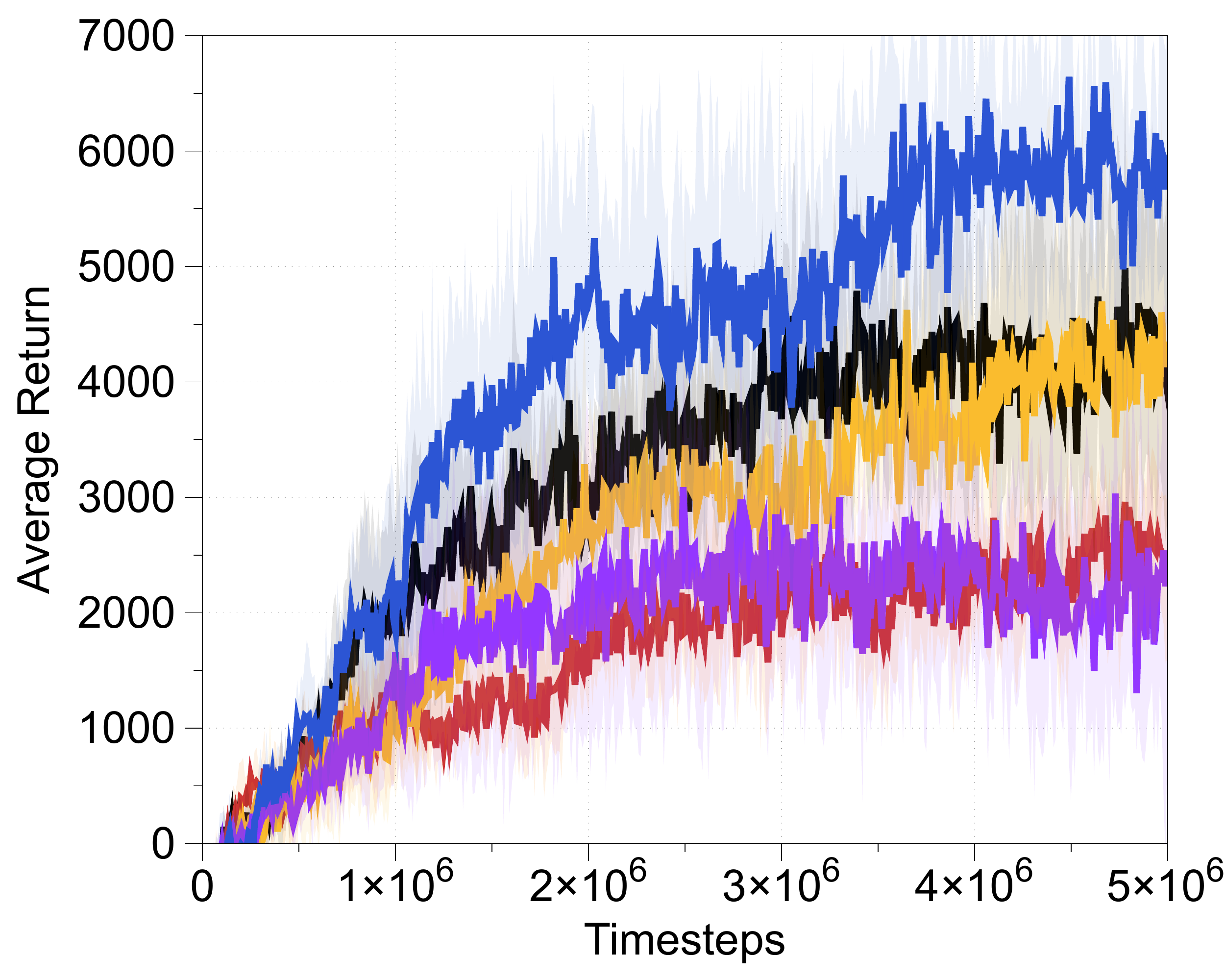}
\label{fig:mfrl_slim_humanoid_moderate}}
\hfill
\vspace{-0.1in}
\caption{The average returns of trained agents on unseen (moderate) environments. The results show the mean and standard deviation of returns averaged over five runs. The full figures of all environments are in the supplementary material.}
\vspace{-0.07in}
\label{fig:mfrl_moderate}
\end{figure*}

\begin{table*}[t!]
\centering
\begin{adjustbox}{max width=\textwidth}
\begin{tabular}{
c
c@{\hspace{0.2cm}}
c
c
c
c@{\hspace{0.2cm}}
c
c
c}
\toprule
  &
  &
  \multicolumn{3}{c}{CartPole} &
  &
  \multicolumn{3}{c}{Pendulum} \\ 
  \cline{3-5} \cline{7-9}
\multicolumn{1}{c}{} &
  &
  Training &
  Test (moderate) &
  Test (extreme) &
  &
  Training &
  Test (moderate) &
  Test (extreme)  \\ \midrule
  Vanilla PPO &
  &
  \textbf{200.0}$\pm$ \scriptsize{0.0} &
  \textbf{199.1}$\pm$ \scriptsize{0.9} &
  187.8$\pm$ \scriptsize{4.7} &
  &
  -945.1$\pm$ \scriptsize{126.0} &
  -1113.2$\pm$ \scriptsize{69.1} &
  -1356.8$\pm$ \scriptsize{48.0} \\
  Stacked PPO &
  &
  198.6$\pm$ \scriptsize{1.7} &
  197.8$\pm$ \scriptsize{1.3} &
  189.2$\pm$ \scriptsize{6.1} &
  &
  -316.9$\pm$ \scriptsize{197.3} &
  -475.7$\pm$ \scriptsize{228.1} &
  -488.2$\pm$ \scriptsize{178.2} \\
  PPO + PC &
  &
  \textbf{200.0}$\pm$ \scriptsize{0.0} &
  198.0$\pm$ \scriptsize{1.4} &
  187.5$\pm$ \scriptsize{10.9} &
  &
  -451.1$\pm$ \scriptsize{248.6} &
  -645.3$\pm$ \scriptsize{320.7} &
  -1136.6$\pm$ \scriptsize{251.0} \\
  PPO + EP &
  &
  \textbf{200.0}$\pm$ \scriptsize{0.0} &
  196.3$\pm$ \scriptsize{4.0} &
  184.5$\pm$ \scriptsize{9.7} &
  &
  -255.7$\pm$ \scriptsize{9.5} &
  -374.3$\pm$ \scriptsize{24.6} &
  -\textbf{256.7}$\pm$ \scriptsize{26.4} \\ \midrule
\textbf{}
  PPO + CaDM &
  &
  \textbf{200.0}$\pm$ \scriptsize{0.0} &
  197.9$\pm$ \scriptsize{3.0} &
  \textbf{193.0}$\pm$ \scriptsize{3.5} &
  &
  \textbf{-199.3}$\pm$ \scriptsize{22.2} &
  \textbf{-279.8}$\pm$ \scriptsize{42.1} &
  -426.4$\pm$ \scriptsize{227.0} \\
  \midrule\midrule
  &
  &
  \multicolumn{3}{c}{HalfCheetah} &
  &
  \multicolumn{3}{c}{Ant} \\ 
  \cline{3-5} \cline{7-9}
\multicolumn{1}{c}{} &
  &
  Training &
  Test (moderate) &
  Test (extreme) &
  &
  Training &
  Test (moderate) &
  Test (extreme)  \\ \midrule
  Vanilla PPO &
  &
  2043.4$\pm$ \scriptsize{802.9} &
  807.7$\pm$ \scriptsize{553.6} &
  574.0$\pm$ \scriptsize{645.6} &
  &
  211.9$\pm$ \scriptsize{44.5} &
  149.4$\pm$ \scriptsize{27.0} &
  117.3$\pm$ \scriptsize{23.1} \\
  Stacked PPO &
  &
  1125.4$\pm$ \scriptsize{85.5} &
  361.1$\pm$ \scriptsize{141.7} &
  5.7$\pm$ \scriptsize{208.1} &
  &
  90.6$\pm$ \scriptsize{16.3} &
  53.2$\pm$ \scriptsize{10.6} &
  46.0$\pm$ \scriptsize{10.9} \\
  PPO + PC &
  &
  1584.9$\pm$ \scriptsize{404.3} &
  642.1$\pm$ \scriptsize{488.3} &
  462.1$\pm$ \scriptsize{534.5} &
  &
  249.9$\pm$ \scriptsize{85.0} &
  207.0$\pm$ \scriptsize{33.8} &
  163.5$\pm$ \scriptsize{30.4} \\
  PPO + EP &
  &
  1620.9$\pm$ \scriptsize{491.5} &
  895.3$\pm$ \scriptsize{445.1} &
  674.2$\pm$ \scriptsize{686.8} &
  &
  138.8$\pm$ \scriptsize{34.9} &
  107.8$\pm$ \scriptsize{19.9} &
  93.5$\pm$ \scriptsize{32.4} \\ \midrule
\textbf{}
  PPO + CaDM &
  &
  \textbf{2652.0}$\pm$ \scriptsize{1133.6} &
  \textbf{1224.2}$\pm$ \scriptsize{630.0} &
  \textbf{1021.1}$\pm$ \scriptsize{676.6} &
  &
  \textbf{268.6}$\pm$ \scriptsize{77.0} &
  \textbf{228.8}$\pm$ \scriptsize{48.4} &
  \textbf{199.2}$\pm$ \scriptsize{52.1}\\ 
  \midrule\midrule
  &
  &
  \multicolumn{3}{c}{CrippledHalfCheetah} &
  &
  \multicolumn{3}{c}{SlimHumanoid} \\ 
  \cline{3-5} \cline{7-9}
\multicolumn{1}{c}{} &
  &
  Training &
  Test (moderate) &
  Test (extreme) &
  &
  Training &
  Test (moderate) &
  Test (extreme)  \\ \midrule
  Vanilla PPO &
  &
  2059.6$\pm$ \scriptsize{658.3} &
  1223.6$\pm$ \scriptsize{559.9} &
  781.7$\pm$ \scriptsize{270.3} &
  &
  7685.5$\pm$ \scriptsize{2599.4} &
  3761.3$\pm$ \scriptsize{1582.4} &
  2751.6$\pm$ \scriptsize{869.4} \\
  Stacked PPO &
  &
  1238.1$\pm$ \scriptsize{102.5} &
  967.1$\pm$ \scriptsize{146.6} &
  904.4$\pm$ \scriptsize{146.5} &
  &
  4831.0$\pm$ \scriptsize{688.1} &
  2443.0$\pm$ \scriptsize{535.6} &
  1577.8$\pm$ \scriptsize{573.5} \\
  PPO + PC &
  &
  \textbf{2920.7}$\pm$ \scriptsize{771.7} &
  1162.2$\pm$ \scriptsize{456.5} &
  546.3$\pm$ \scriptsize{215.9} &
  &
  7130.1$\pm$ \scriptsize{3378.0} &
  3928.5$\pm$ \scriptsize{1848.7} &
  2362.6$\pm$ \scriptsize{781.9} \\
  PPO + EP &
  &
  1494.2$\pm$ \scriptsize{311.7} &
  1017.0$\pm$ \scriptsize{201.1} &
  719.0$\pm$ \scriptsize{438.5} &
  &
  4824.7$\pm$ \scriptsize{1508.7} &
  2224.7$\pm$ \scriptsize{882.9} &
  1293.4$\pm$ \scriptsize{729.0} \\ \midrule
\textbf{}
  PPO + CaDM &
  &
  2356.6$\pm$ \scriptsize{624.3} &
  \textbf{1454.0}$\pm$ \scriptsize{462.6} &
  \textbf{1025.0}$\pm$ \scriptsize{296.2} &
  &
  \textbf{10455.0}$\pm$ \scriptsize{1004.9} &
  \textbf{4975.7}$\pm$ \scriptsize{1305.7} &
  \textbf{3015.1}$\pm$ \scriptsize{1508.3} \\ \bottomrule
\end{tabular}
\end{adjustbox}
\vspace{-0.1in}
\caption{The performance (average return) of trained agents on various control tasks. The transition dynamics of environments are changing in both training and test environments.
The results show the mean and standard deviation of returns averaged over five runs.} 
\label{tbl:mfrl_test}
\vspace{-0.15in}
\end{table*}


\subsection{Comparison with the Model-based RL Methods} \label{sec:exp_model_based}

Table~\ref{tbl:mbrl_test} shows the performance of various model-based RL methods on both training and test environments (see the supplementary material for training curve plots).
Our method significantly improves both training and test performances of various model-based methods in all the environments.
Especially, the performance gain due to CaDM becomes much more significant in more complex environments (e.g., long-horizon and high-dimensional domains like halfCheetah, Ant, and Humanoid.
For example, when combined with PE-TS, CaDM improves the average return from 2019.6 to 7087.2 for the HalfCheetah environment in the moderate regime.
This demonstrates the applicability of our method to any model-based RL methods.
One important result is that stacking input transitions sometimes degrades the performances in both training and test environments, which implies that our approach to separate context encoding and transition inference is indeed more effective for approximating true context $c$ than a na\"ive stacking method.
We also found that the proposed method is more stable compared to model-based meta-RL methods (i.e., GrBAL and ReBAL), which update model parameters
or hidden states.

\begin{figure*} [t] \centering
\vspace{-0.08in}
\subfigure[Effects of prediction loss]
{
\includegraphics[width=0.3\textwidth]{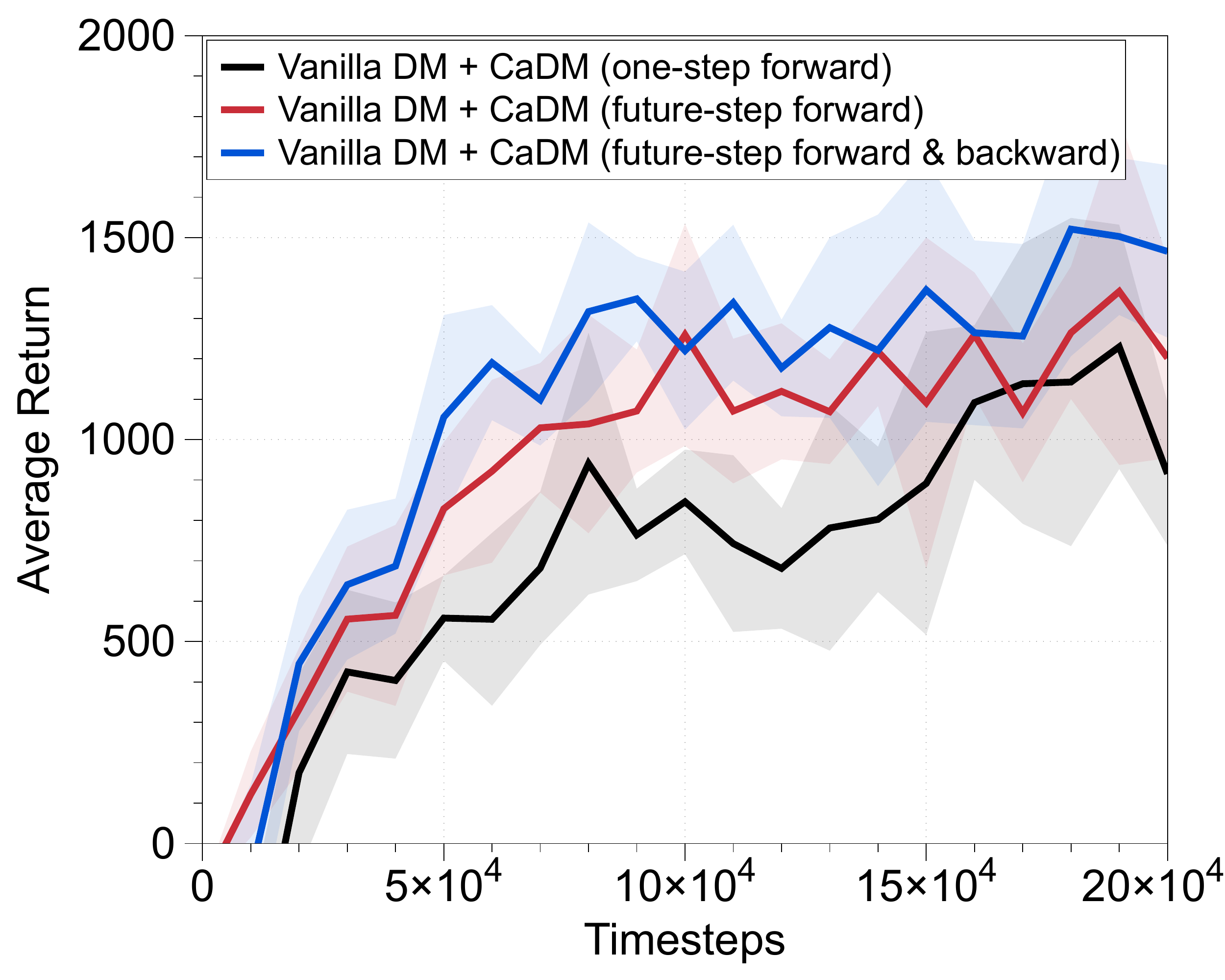} \label{fig:mbrl_ablation_test}} 
\,
\subfigure[Prediction error]
{
\includegraphics[width=0.3\textwidth]{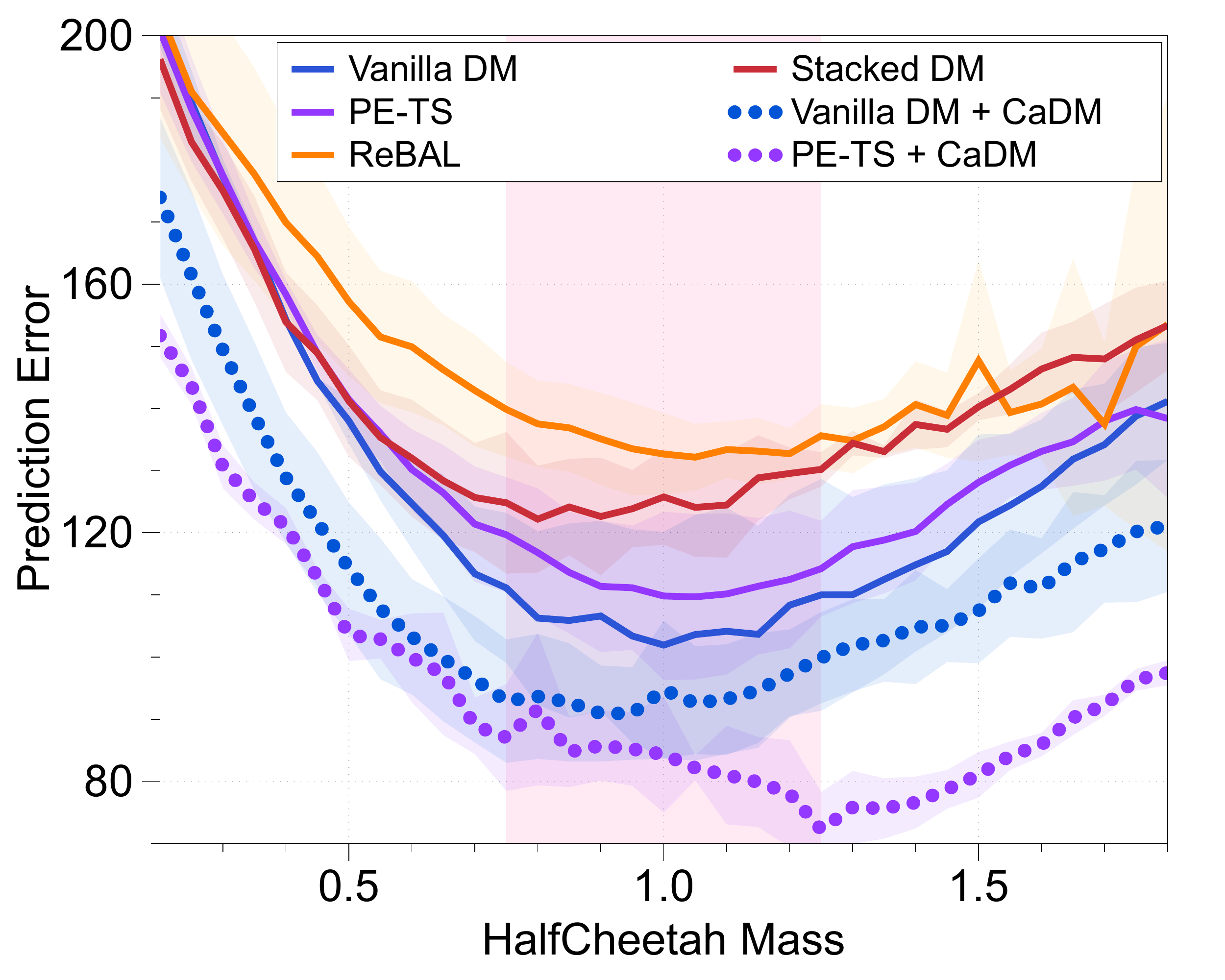} \label{fig:prediction_error}}
\,
\subfigure[Context latent vector]
{
\includegraphics[width=0.293\textwidth]{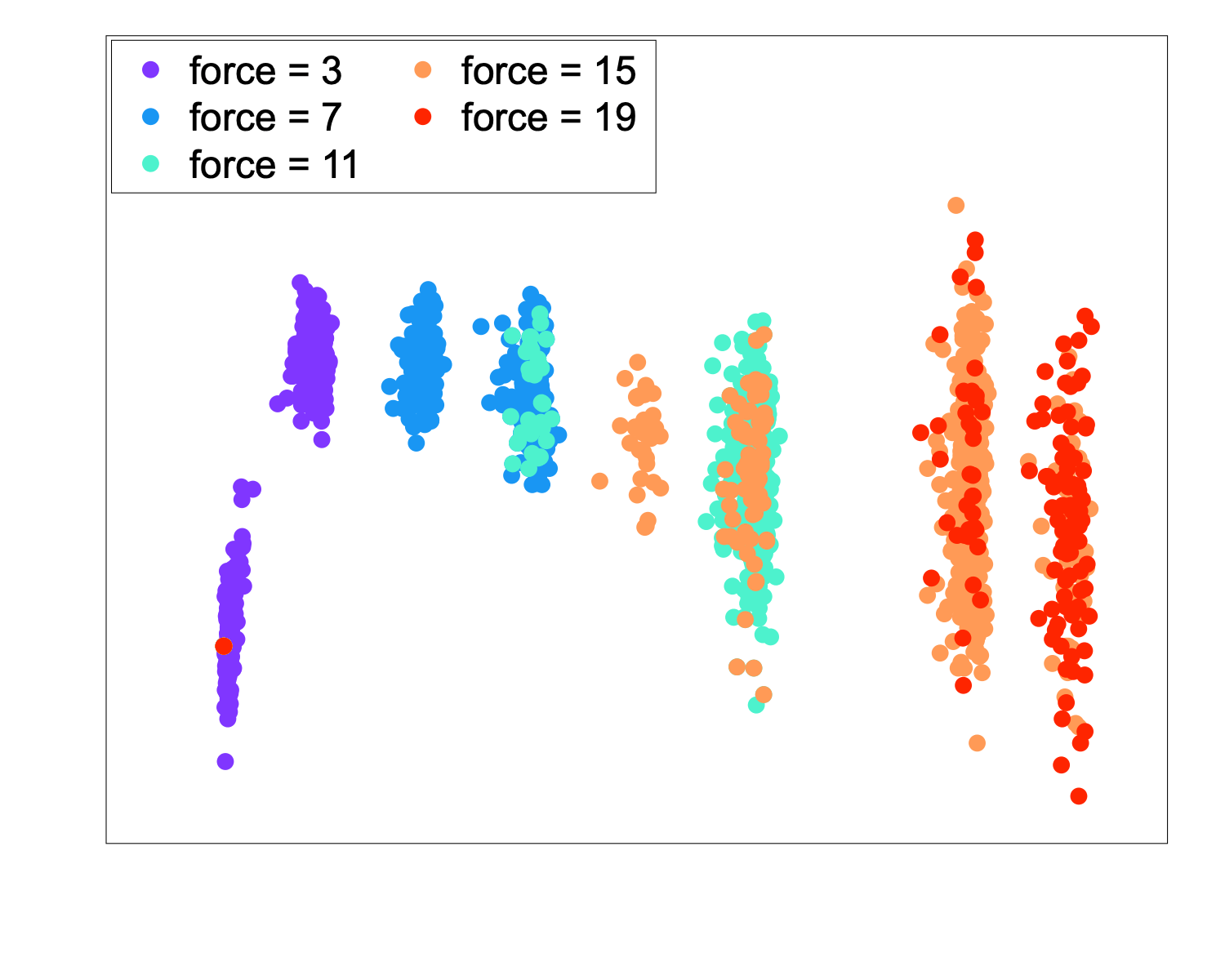} \label{fig:tsne_ours_context}}
\vspace{-0.1in}
\caption{
(a) Test performance of dynamics models optimized by variants of the proposed prediction objective in \eqref{eq:main_obj} on HalfCheetah environments.
(b) Prediction errors 
on the HalfCheetah task with varying mass values.
The pink shaded area represents the training range.
(c) PCA visualization of context latent vectors extracted from trajectories collected in the CartPole environments. Embedded points from environments with the same mass parameter have the same color. The full figures of all environments are in the supplementary material.}
\vspace{-0.07in}
\label{fig:embedding_analysis}
\end{figure*}


\begin{figure*} [th] \centering
\includegraphics[width=0.94\textwidth]{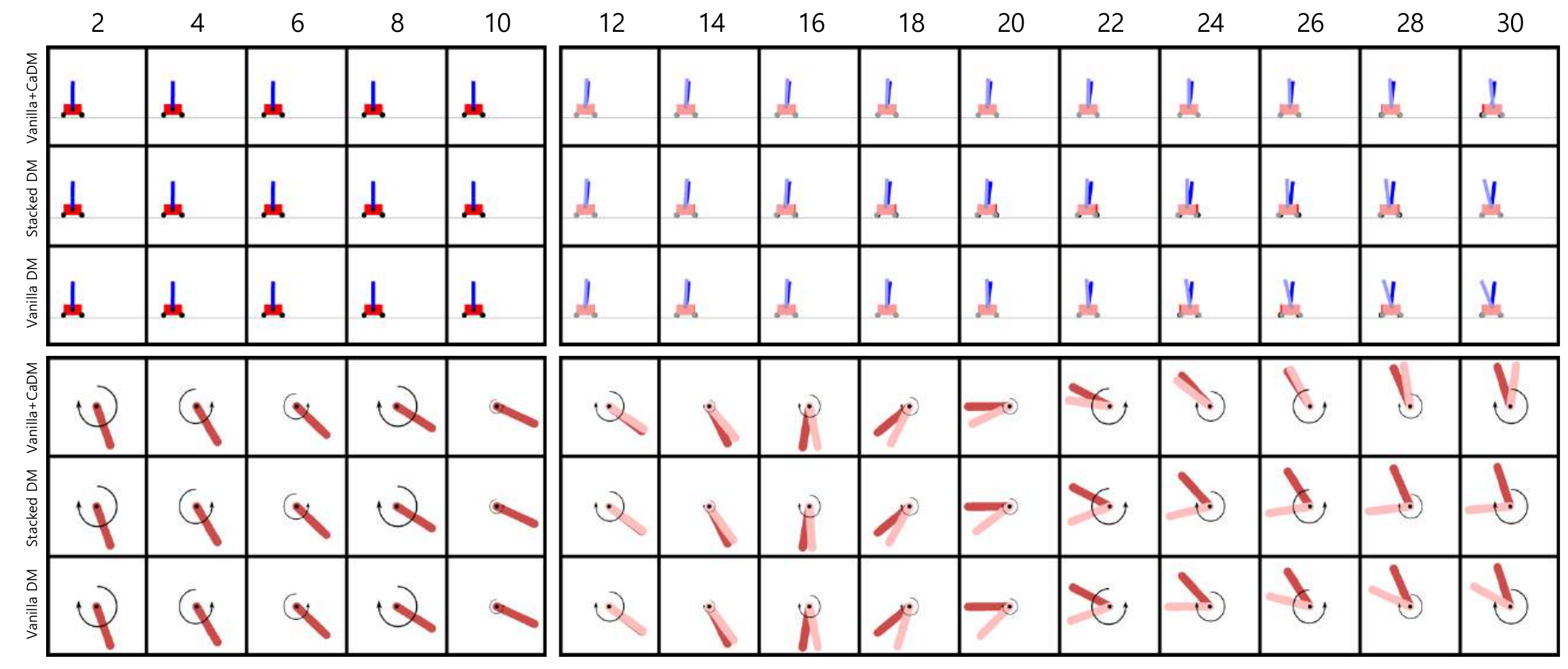}
\vspace{-0.1in}
\caption{Visualization of future state predictions from Vanilla + CaDM (ours), Stacked DM, and Vanilla DM on CartPole and Pendulum environments with unseen environment parameters.
Given 10 past states and actions,
we predict future states for the next 20 timesteps only using ground truth actions.
The colored and faded objects represent the ground truth states and the predicted states, respectively.}
\vspace{-0.15in}
\label{fig:modelling error}
\end{figure*}

\subsection{Comparison with the Model-free RL Methods}  \label{seq:comb_model_free}
We also verify whether the learned context latent vector is useful for improving the generalization performance of model-free RL methods.
Specifically, similar to \citet{henderson2018deep}, we use MLPs with two hidden layers of 64 units and tanh activations for the policy network, and the Proximal Policy Optimization (PPO; \citealt{schulman2017proximal}) method to train the agents.
Our proposed method, which takes the learned context latent vector (denoted PPO + CaDM), is compared with several context-conditional policies \cite{rakelly2019efficient, zhou2019environment}.
Specifically, we consider the stacked PPO, which takes the past $K \in \{5, 10, 15\}$ transitions as an additional input,
and PPO with probabilistic context (PPO + PC), which learns context variable by maximizing the expected returns \cite{rakelly2019efficient}.
We also consider PPO with environment probing policy (PPO + EP) that takes embeddings extracted from initial interaction with an environment as an additional input \cite{zhou2019environment}.
Due to space limitations, we provide more detailed explanations in the supplementary material.

Table~\ref{tbl:mfrl_test} shows the performance of various model-free RL methods on both training and test environments (see the supplementary for training curve plots).
In complex environments, such as HalfCheetah, Ant, CrippledHalfCheetah, and SlimHumanoid,
our PPO + CaDM shows better generalization performances than previous conditional policy methods (i.e., PPO + EP and PPO + PC), implying that the proposed CaDM method can extract contextual information more effectively. 
On the other hand, 
the performance gain from our method is marginal in simple environments, such as 
CartPole and Pendulum.


\subsection{Ablation Study}

{\bf Effects of prediction loss}.
In order to verify the individual effects of the suggested prediction losses \eqref{eq:main_obj},
we train three different dynamics models that optimize (a) one-step forward prediction loss, (b) future-step forward prediction loss, and (c) future-step forward and backward prediction loss, respectively. 
Figure~\ref{fig:mbrl_ablation_test} shows the test performances for the HalfCheetah environment in the extreme regime. One can observe that optimizing the future-step forward and backward prediction loss achieves the best performances. This shows that every component in the proposed loss function helps the context encoder extract the environment context.

{\bf Prediction errors.}
To show that our method indeed helps with a forward prediction,
we compare baseline methods with CaDM in terms of prediction error
across multiple HalfCheetah environments with varying mass values.
As shown in Figure~\ref{fig:prediction_error}, our model demonstrates superior prediction performance in both training and test environments.
In particular, PE-TS quickly becomes unreliable outside the training range (pink shaded area),
whereas our model maintains a tolerable level of prediction errors throughout.


{\bf Embedding analysis}. 
We analyze whether the learned latent vector encodes meaningful information about the environment. To this end, we collect trajectories from CartPole environments with different push force magnitudes, then visualize the latent encoding of the collected trajectory segments using principal component analysis (PCA) \cite{jolliffe1986principal}.
As shown in Figure~\ref{fig:tsne_ours_context}, latent vectors extracted from environments with different mass parameters are clearly separated in the embedding space.
In contrast, raw state vectors are scattered and disjointed (see Figure~\ref{fig:cartpole_ablation_raw_pca} in the supplementary material),
which implies that our context encoder captures useful contextual information.

{\bf Prediction visualization}. 
We also visualize the future state predictions in test environments from CartPole and Pendulum in Figure~\ref{fig:modelling error} with unseen environment parameters (i.e., force and mass).
Given 10 past states and actions, we generate 20 future state predictions from Vanilla DM, Stacked DM, and Vanilla + CaDM only using ground truth actions.
One can observe that Vanilla + CaDM gives much more accurate predictions compared to other models.
Notably, Vanilla DM and Stacked DM fail to provide accurate predictions for more distant future timesteps, while CaDM consistently gives accurate predictions across all timesteps,
which confirms our belief that the proposed method can capture contextual information about the transition dynamics.

\section{Conclusion}
In this paper, 
we propose a context-aware dynamics model that adapts to dynamics changes.
To learn a generalizable dynamics model,
we separate context encoding (i.e., capturing the contextual information) and transition inference (i.e., predicting the next state conditioned on the captured information).
By forcing the context latent vector to be useful for both forward dynamics and backward dynamics, our method aptly captures the contextual information. We also showed that the learned context vector can be utilized to improve the generalization performance of model-free methods. We believe our work can serve as a strong guideline in related topics.

\section*{Acknowledgements}
This work was funded in part by ONR PECASE N000141612723, Tencent, Sloan Research Fellowship, Institute of Information \& communications Technology Planning \& Evaluation (IITP) grant funded by the Korea government (MSIT) (No.2019-0-00075, Artificial Intelligence Graduate School Program (KAIST)), and Institute for Information \& communications Technology Planning \& Evaluation (IITP) grant funded by the Korea government (MSIT) (No. 2019-0-01396, Development of framework for analyzing, detecting, mitigating of bias in AI model and training data).
We thank Sungsoo Ahn, Wilka Carvalho, Hyuntak Cha, Chaewon Kim, Hankook Lee, Kibok Lee, Sangwoo Mo, and Jihoon Tack for providing helpful feedbacks and suggestions in preparing the early version of the manuscript.

\bibliography{example_paper}
\bibliographystyle{icml2020}


\appendix
\onecolumn

\begin{center}{\bf {\LARGE Supplementary Material}}
\end{center}

\section{Experimental Details}

\subsection{Environment}

\begin{figure} [htb] \centering
    \begin{center}
    \begin{subfigure}
            \centering
            \includegraphics[width=1.0\textwidth]{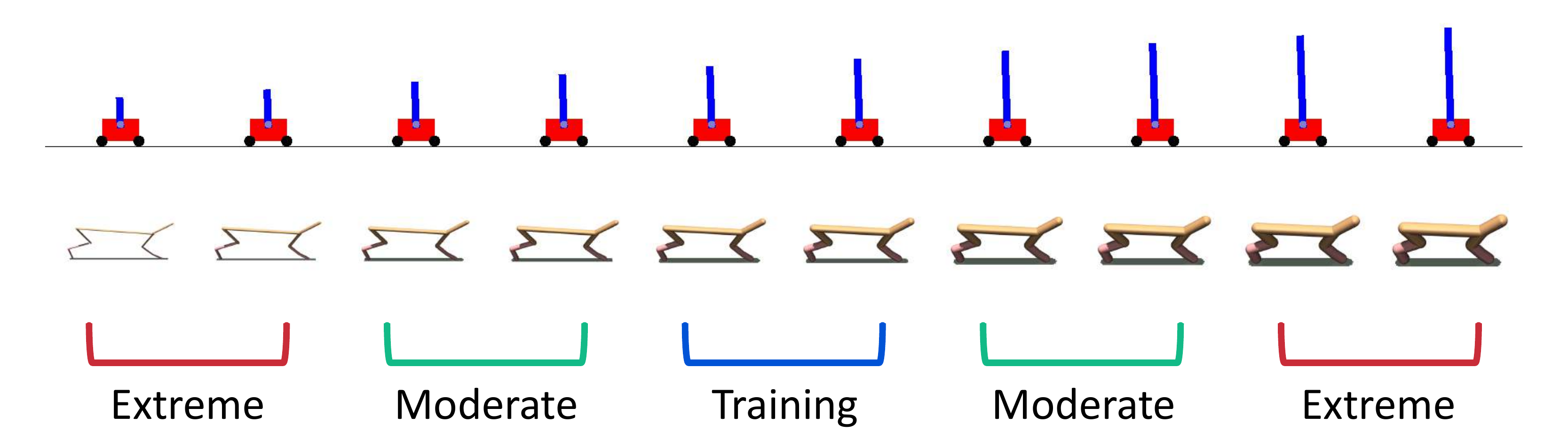}
    \end{subfigure}%
    \end{center}
    \caption{Illustration of training and test environments. }
    \label{fig:seen_unseen_environments}
\end{figure}

{\bf CartPole.} CartPole consists of a pushable cart and a pole attached on top of it by an unactuated joint. The goal is to push the cart left and right, so that the pole does not fall over. 
\begin{itemize}
    \item Observation. $(x_{t}, \dot{x}_{t}, \theta_{t}, \dot{\theta}_{t})$, where $x$ is the cart position, and $\theta$ is the pole's angle from the upright position.
    \item Action. $\{0,1\}$, where $0$ and $1$ correspond to pushing the cart left and right, respectively. 
    \item Reward. $r_{t} = \mathbbm{1}_{\{ |x_{t+1}|<2.4 \wedge |\theta_{t+1}|< 2\pi \cdot (\frac{14}{360})\}}$. That is, reward is $1$ if the cart position is within $\pm 2.4$ and pole angle is within $\pm 14^{\circ}$ during the next timestep, and $0$ otherwise.
    \item Modifications. In our experiments, we modify push force $f$ and pole length $l$ (see Table~\ref{tbl:environment}).
\end{itemize}

{\bf Pendulum.} Pendulum consists of a single rigid body, one end of which is attached to an actuated joint. The goal is to swing up the body and keep it in the upright position.

\begin{itemize}
    \item Observation. $(\cos\theta, \sin\theta, \dot{\theta})$, where $\theta \in [-\pi, \pi]$ is zero in the upright position, and increases in the counter-clockwise direction.
    \item Action. $a\in [-2.0, 2.0]$ represents counter-clockwise torque applied to the pendulum body.
    \item Reward. $r_{t} = -(\theta_{t}^{2} + 0.1\dot{\theta}_{t}^{2} + 0.001 a_{t}^{2})$, which penalises 1) deviation from the upright position, 2) non-zero angular velocity, and 3) joint actuation.
    \item Modifications. We modify pendulum mass $m$ and pendulum length $l$ (see Table~\ref{tbl:environment}).
\end{itemize}

{\bf Half-cheetah.}\footnotemark Half-cheetah agent is made up of 7 rigid links (1 for torso, 3 for forelimb, and 3 for hindlimb), connected by 6 joints. The goal is to move forward as fast as possible, while keeping the control cost minimal. We used additional pre-processing for observations as in \cite{chua2018deep}.
\begin{itemize}
    \item Observation. Observation is given by a 20-dimensional vector that includes 1) angular position and velocity of all 6 joints, 2) root joint's position (except for the x-coordinate) and velocity, and 3) center of mass of the torso. 
    \item Action. $a\in [-1.0, 1.0]^{6}$ represents torques applied at six joints. 
    \item Reward. $r_{t} = \dot{x}_{\text{torso}, t} - 0.05 \Vert a_{t} \Vert^{2}$, where $\dot{x}_{\text{torso}, t}$ represents forward velocity of the torso.
    \item Modification. As for dynamics modification, we 1) scale mass of every rigid link by a fixed scale factor $m$, and 2) scale damping of every joint by a fixed scale factor $d$ (see Table~\ref{tbl:environment}).
\end{itemize}

{\bf Ant.}\footnotemark[\value{footnote}] Ant agent consists of 13 rigid links connected by 8 joints. The goal is to move forward as fast as possible at a minimal control cost. 
We used additional pre-processing for observations as in \cite{chua2018deep}.
\begin{itemize}
    \item Observation. Observation is a 41-dimensional vector that includes 1) angular position and velocity of all 8 joints, 2) position and velocity of the root joint, 3) frame orientations (xmat) and 4) center of mass of the torso. 
    \item Action. $a \in [-1, 1]^{8}$ represents torques applied at eight joints. 
    \item Reward. $r_{t} = \dot{x}_{\text{torso}, t} - 0.005\Vert a_{t}\Vert^{2} + 0.05$, where $\dot{x}_{\text{torso}, t}$ denotes forward velocity of the torso.
    \item Modification. As for dynamics modification, we modify the mass of every leg. Specifically, given a scale factor $m$, we modify two legs to have mass multiplied by $m$, and the other two legs to have mass multiplied by $\frac1{m}$ (see Table~\ref{tbl:environment}).  
\end{itemize}

{\bf SlimHumanoid.}\footnotemark[\value{footnote}] 
SlimHumanoid \cite{wang2019benchmarking} consists of 13 rigid links with 17 actuators. The goal is to move forward as fast as possible, while keeping the control cost minimal. We used additional pre-processing for observations as in \cite{chua2018deep}.

\begin{itemize}
    \item Observation. Observation is a 45-dimensional vector that includes angular position and velocities.
    \item Action. $a \in [-0.4, 0.4]^{17}$.
    \item Reward. $r_{t} = 50/3 \times \dot{x}_{\text{torso}, t} - 0.1\Vert a_{t}\Vert^{2} + 5 \times \text{bool}(1.0 \leq z_{\text{torso}, t} \leq 2.0)$, where $\dot{x}_{\text{torso}, t}$ denotes forward velocity of the torso and $z_{t}$ is the height of the torso.
    \item Modification. As for dynamics modification, we 1) scale mass of every rigid link by a fixed scale factor $m$, and 2) scale damping of every joint by a fixed scale factor $d$ (see Table~\ref{tbl:environment}).  
\end{itemize}

\footnotetext{We modified the publicly available code at \url{https://github.com/iclavera/learning\_to\_adapt} for implementation of these environments.}

\begin{table}[t]
\centering
\resizebox{1.0\textwidth}{!}{
\begin{tabular}{@{}crlrlrlc@{}}
\toprule
\multicolumn{1}{l}{}                               
& \multicolumn{2}{c}{Train}                                                             
& \multicolumn{2}{c}{\begin{tabular}[c]{@{}c@{}}Test\\ (Moderate)\end{tabular}} 
& \multicolumn{2}{c}{\begin{tabular}[c]{@{}c@{}}Test\\ (Extreme)\end{tabular}} 
& \begin{tabular}[c]{@{}c@{}}Episode\\ Length\end{tabular} \\ \midrule
\multicolumn{1}{c|}{\multirow{2}{*}{CartPole}}     & $f\in$  & \multicolumn{1}{l|}{\begin{tabular}[c]{@{}l@{}}$\{5.0, 6.0, 7.0, 8.0, 9.0, 10.0,$\\ $\hspace{1mm} 11.0, 12.0, 13.0, 14.0, 15.0\}$\end{tabular}}    & $f\in$       & \multicolumn{1}{l|}{$\{3.0, 3.5, 16.5, 17.0\}$}               & $f\in$       & \multicolumn{1}{l|}{$\{2.0, 2.5, 17.5, 18.0\}$}               & \multirow{2}{*}{200}                                     \\
\multicolumn{1}{c|}{}                              & $\l\in$ & \multicolumn{1}{l|}{$\{0.40, 0.45, 0.50, 0.55, 0.60\}$}                                                                                            & $l\in$       & \multicolumn{1}{l|}{$\{0.25, 0.30, 0.70, 0.75\}$}             & $l\in$       & \multicolumn{1}{l|}{$\{0.15, 0.20, 0.80, 0.85\}$}             &                                                          \\ \midrule
\multicolumn{1}{c|}{\multirow{2}{*}{Pendulum}}     & $m\in$  & \multicolumn{1}{l|}{\begin{tabular}[c]{@{}l@{}}$\{0.75, 0.80, 0.85, 0.90, 0.95,$\\ $\hspace{1mm}1.0, 1.05, 1.10, 1.15, 1.20, 1.25\}$\end{tabular}} & $m\in$       & \multicolumn{1}{l|}{$\{0.50, 0.70, 1.30, 1.50\}$}             & $m\in$       & \multicolumn{1}{l|}{$\{0.20, 0.40, 1.60, 1.80\}$}             & \multirow{2}{*}{200}                                     \\
\multicolumn{1}{c|}{}                              & $l\in$  & \multicolumn{1}{l|}{\begin{tabular}[c]{@{}l@{}}$\{0.75, 0.80, 0.85, 0.90, 0.95,$\\ $\hspace{1mm}1.0, 1.05, 1.10, 1.15, 1.20, 1.25\}$\end{tabular}} & $l\in$       & \multicolumn{1}{l|}{$\{0.50, 0.70, 1.30, 1.50\}$}             & $l\in$       & \multicolumn{1}{l|}{$\{0.20, 0.40, 1.60, 1.80\}$}             &                                                          \\ \midrule
\multicolumn{1}{c|}{\multirow{2}{*}{Half-cheetah}} & $m\in$  & \multicolumn{1}{l|}{$\{0.75, 0.85, 1.0, 1.15, 1.25\}$}                                                                                             & $m\in$       & \multicolumn{1}{l|}{$\{0.40, 0.50, 1.50, 1.60\}$}             & $m\in$       & \multicolumn{1}{l|}{$\{0.20, 0.30, 1.70, 1.80\}$}             & \multirow{2}{*}{1000}                                     \\
\multicolumn{1}{c|}{}                              & $d\in$  & \multicolumn{1}{l|}{$\{0.75, 0.85, 1.0, 1.15, 1.25\}$}                                                                                             & $d\in$       & \multicolumn{1}{l|}{$\{0.40, 0.50, 1.50, 1.60\}$}             & $d\in$       & \multicolumn{1}{l|}{$\{0.20, 0.30, 1.70, 1.80\}$}             &                                                          \\ \midrule
\multicolumn{1}{c|}{Ant}                           & $m\in$  & \multicolumn{1}{l|}{$\{0.85, 0.90, 0.95, 1.0\}$}                                                                                                   & $m\in$       & \multicolumn{1}{l|}{$\{0.20, 0.25, 0.30, 0.35, 0.40\}$}       & $m\in$       & \multicolumn{1}{l|}{$\{0.40, 0.45, 0.50, 0.55, 0.60\}$}       & 1000 
\\ \midrule
\multicolumn{1}{c|}{\multirow{2}{*}{SlimHumanoid}} & $m\in$  & \multicolumn{1}{l|}{$\{0.80, 0.90, 1.0, 1.15, 1.25\}$}                                                                                             & $m\in$       & \multicolumn{1}{l|}{$\{0.60, 0.70, 1.50, 1.60\}$}             & $m\in$       & \multicolumn{1}{l|}{$\{0.40, 0.50, 1.70, 1.80\}$}             & \multirow{2}{*}{1000}                                     \\
\multicolumn{1}{c|}{}                              & $d\in$  & \multicolumn{1}{l|}{$\{0.80, 0.90, 1.0, 1.15, 1.25\}$}                                                                                             & $d\in$       & \multicolumn{1}{l|}{$\{0.60, 0.70, 1.50, 1.60\}$}             & $d\in$       & \multicolumn{1}{l|}{$\{0.40, 0.50, 1.70, 1.80\}$}             &                                                          \\ \bottomrule
\end{tabular}}
\caption{Environment parameters used for our experiments.}
\label{tbl:environment}
\end{table}

\newpage

\subsection{Training Details}
\textbf{Model-based RL.}
To train the dynamics model, we collect 10 trajectories with MPC controller from environments and train the model for 5 epochs at every iteration. We train the model for 20 iterations for every experiments. To report test performance, we evaluated trained models every iteration on environments with fixed random seeds. The Adam optimizer \cite{kingma2014adam} is used with the learning rate 0.001. 
For planning, we used the cross entropy method (CEM) with 200 candidate actions except for CartPole and Pendulum, where random shooting (RS) method with 1,000 candidate actions is used. The horizon of MPC is 30.

\textbf{Model-free RL.}
We trained PPO agents for 5 million timesteps on Pendulum and MuJoCo environments (i.e. Half-cheetah, Ant, Crippled Half-cheetah, Walker) and 0.5 million timesteps on CartPole. We evaluated trained agents every 10,000 timesteps on environments with fixed random seeds. We use a discount factor $\gamma = 0.99$, a generalized advantage estimator \cite{schulman2015high} parameter $\lambda = 0.95$ and an entropy bonus of 0.01 for exploration. We use 10 rollouts with 200 timesteps per each rollout, and then train agents for 8 epochs with 4 mini-batches. The Adam optimizer is used with the learning rate 0.0005.

\subsection{Implementation of Context-aware Dynamics Model}
For our method, the context encoder is modeled as multi-layer perceptrons (MLPs) with 3 hidden layers that produce a 10-dimensional vector.
Then, CaDM receives the context vector as an additional input, i.e., the input is given as a concatenation of state, action, and context vector.
We use $\beta \in \{0.25, 0.5, 1.0\}$ for the penalty parameter in \eqref{eq:main_obj}, $K \in \{5, 10\}$ for the number of past observations and $M \in \{5, 10\}$ for the number of future observations.

\subsection{Implementation of Model-based RL Baselines}

\textbf{Vanilla DM.}
We model the dynamics model as Gaussian, in which the mean is parameterized by MLPs with 4 hidden layers of 200 units each and Swish activations \cite{ramachandran2017searching} and the variance is fixed.
Note that maximum likelihood estimation, i.e., $\log f(s^\prime|s,a)$ corresponds to minimizing the mean squared error in this setup.

\textbf{Stacked DM.}
We implemented Stacked DM to take additional 10 observations as an additional input. 

\textbf{GrBAL and ReBAL.}
We used a reference implementation provided by the authors.\footnote{\url{https://github.com/iclavera/learning_to_adapt}}
For GrBAL, we model the dynamics model as MLPs with 3 layers of 512 units each and ReLU activations. The size of adaptive batch is 16.
For ReBAL, we use 1-layer LSTM with 256 cells. We use planning horizon $h \in \{10, 30\}$ and learning rate 0.01.

\textbf{PE-TS.}
We implemented PE-TS based on the code provided by the authors.\footnote{\url{https://github.com/kchua/handful-of-trials}} 
Following the setup in \citet{chua2018deep}, we used 5 bootstrap models for ensemble and 20 particles for trajectory sampling. 
We model the dynamics model as MLPs with 4 hidden layers of 200 units each and Swish activations. 

\subsection{Implementation of Model-free RL Baselines}
\textbf{Stacked PPO.}
We implemented Stacked PPO to take additional 10 observations as an additional input. 

\textbf{PPO + PC.}
This baseline is implemented based on \citet{rakelly2019efficient}.
We train a PPO agent with probabilistic context encoder that takes $K \in \{5, 10, 15\}$ observations as an additional input. 
We model the context encoder as the product of independent Gaussian factors, in which the mean and the variance are parameterized by MLPs with 3 hidden layers of (256, 128, 64) units that produce a 10-dimensional vector.
Note that this probabilistic context encoder is permutation invariant as in the original paper.
For $\beta$ in KL divergence term from variational lower bound, we select $\beta$ from $\{0.5, 1.0\}$.

\textbf{PPO + EP.}
This baseline is implemented based on \citet{zhou2019environment}.
We model the context encoder as the product of independent Gaussian factors, in which the mean and the variance are parameterized by MLPs with 3 hidden layers of (256, 128, 64) units that produce a 10-dimensional vector.
We first train interaction policy using two dynamics models for $\{100\text{K}, 125\text{K}\}$ timesteps, which are the same as the number of samples used for training Vanilla + CaDM, and then train a context-conditional policy.



\newpage
\section{Effects of Training Environments}
\begin{figure*} [h!] \centering
\subfigure[Vanilla DM]
{
\includegraphics[width=0.9\textwidth]{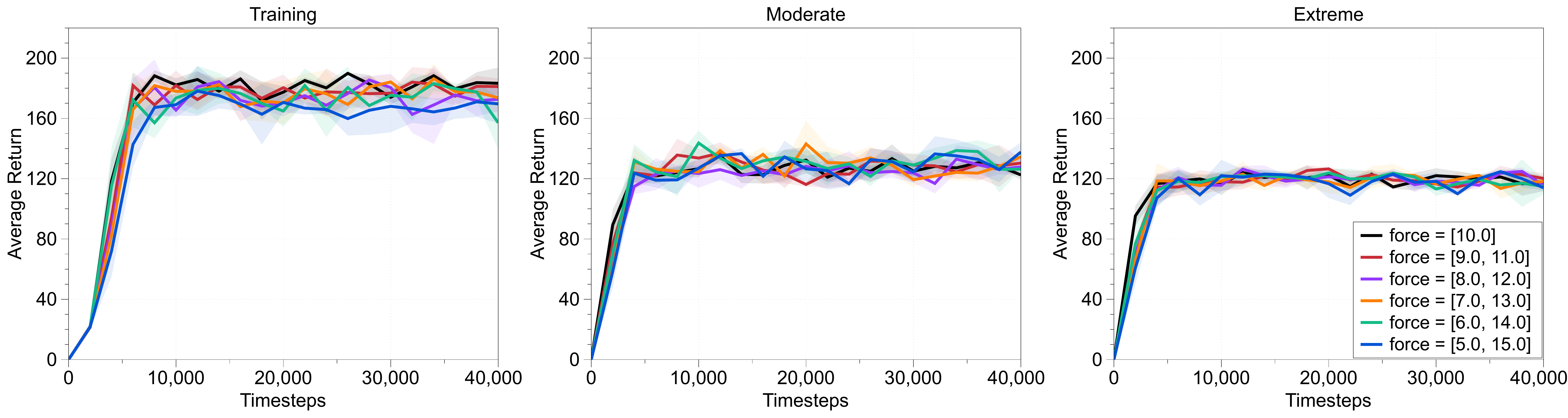} \label{fig:sup_cartpole_range_normal}}
\vspace{-0.1in}
\\
\subfigure[Vanilla DM + CaDM]
{
\includegraphics[width=0.9\textwidth]{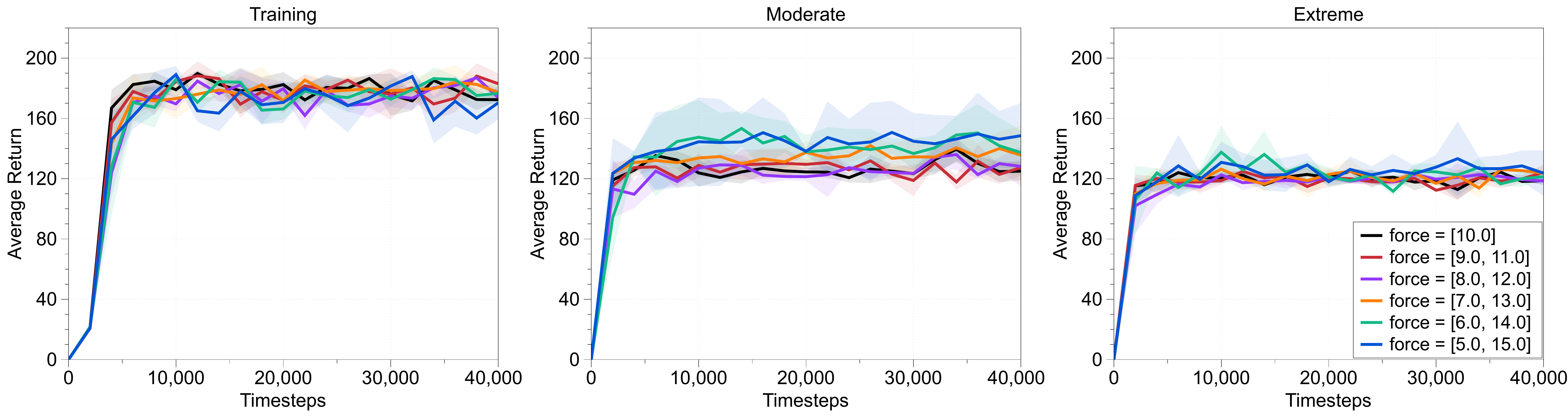} \label{fig:sup_cartpole_range_cadm}}
\vspace{-0.1in}
\caption{
The performance (average returns) of (a) Vanilla DM and (b) Vanilla DM + CaDM on CartPole environment with varying training ranges.
The transition dynamics of environments are changing in both training and test environments. 
The solid line and shaded regions represent the mean and standard deviation, respectively, across five runs.}
\label{fig:sup_cartpole_range}
\vspace{-0.15in}
\end{figure*}

\begin{figure*} [h!] \centering
\includegraphics[width=0.4\textwidth]{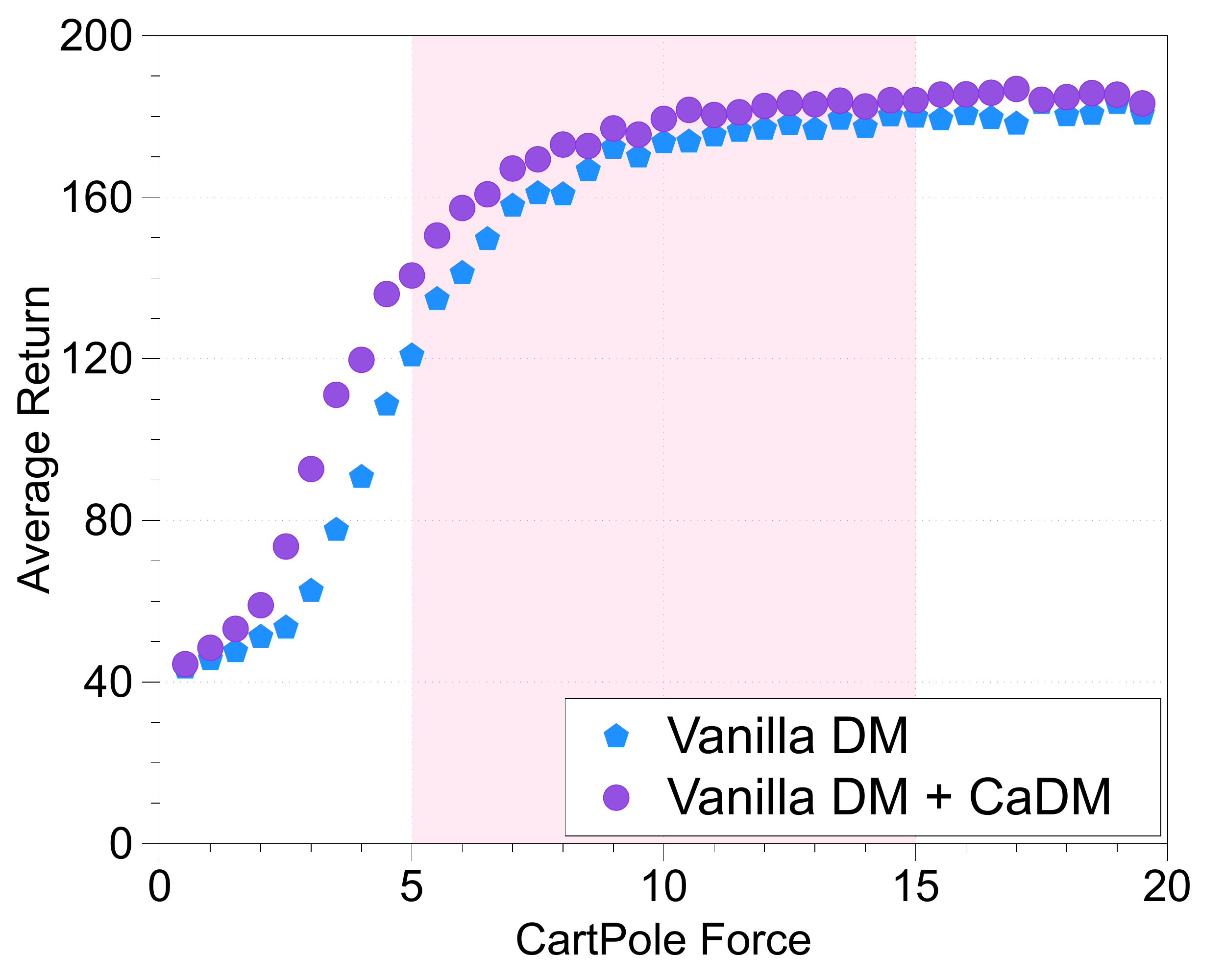}
\,
\includegraphics[width=0.4\textwidth]{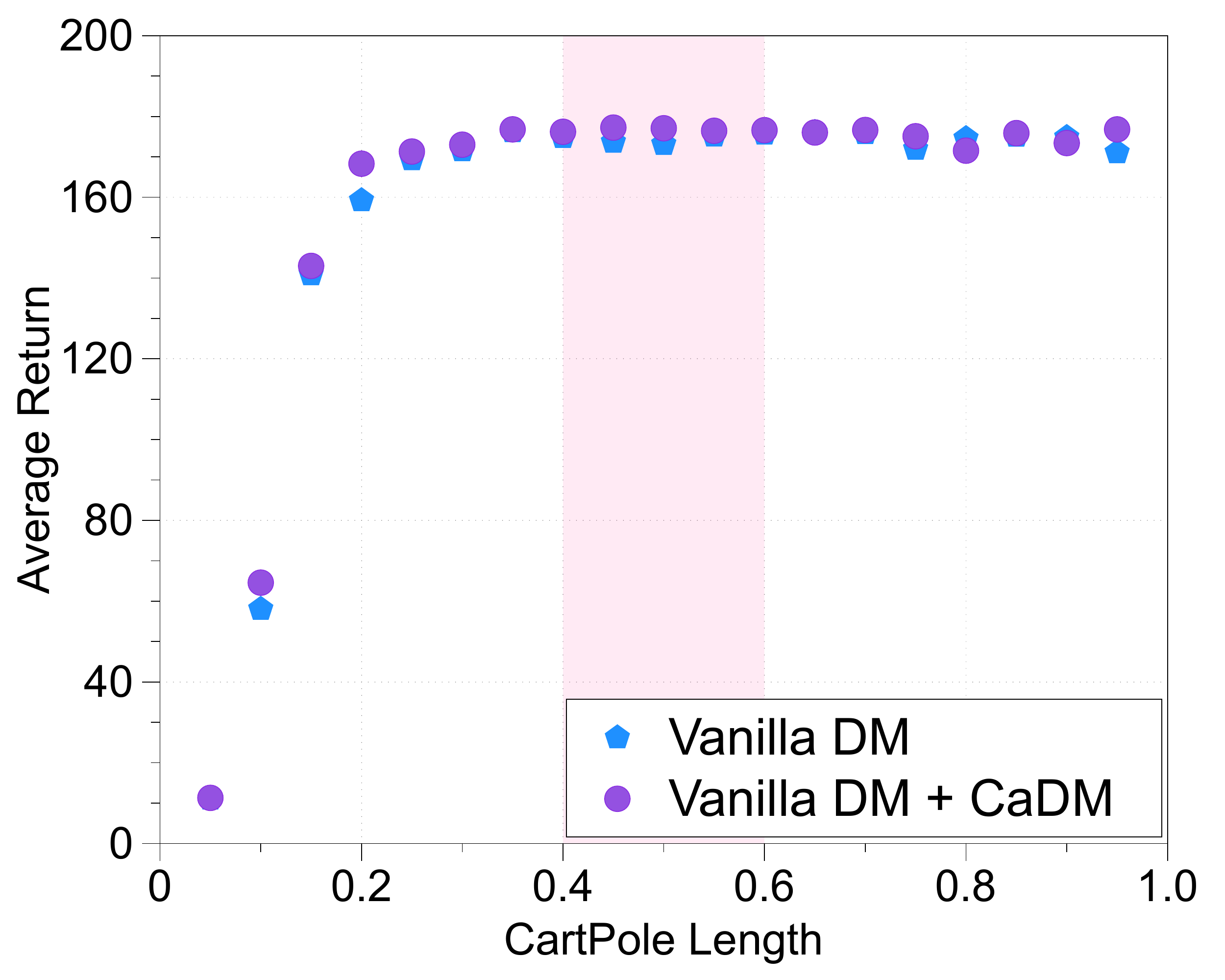}
\vspace{-0.1in}
\caption{
The performance of Vanilla DM and Vanilla DM + CaDM on unseen (moderate) CartPole environments with varying force and length values. The pink shaded area represents the training range. The results show the mean of returns averaged over three runs.}
\label{fig:sup_cartpole_range_varying}
\vspace{-0.15in}
\end{figure*}

In this section, we analyze the effects of training environments on the generalization performance of trained dynamics models.
First, we compare the performance of Vanilla DM with CaDM when the range of training environments gets wider.
As shown in Figure~\ref{fig:sup_cartpole_range}, the generalization performance of Vanilla DM + CaDM on CartPole improves when the model is trained on more wider range of environments while the performance of Vanilla DM does not improve, which implies that our CaDM can indeed utilize contextual information.
We also measure the generalization performance of trained Vanilla DM + CaDM on a wide range of simulation parameters of CartPole (see Figure~\ref{fig:sup_cartpole_range_varying}). As expected, the test performance degrades as the simulation parameter deviates from training environments (pink shaded area). However, Vanilla DM + CaDM still consistently outperforms Vanilla DM in most simulation parameters.

\newpage

\section{Learning Curves for Model-Based RL}
\begin{figure*} [h!] \centering
\subfigure[CartPole]
{
\includegraphics[width=0.9\textwidth]{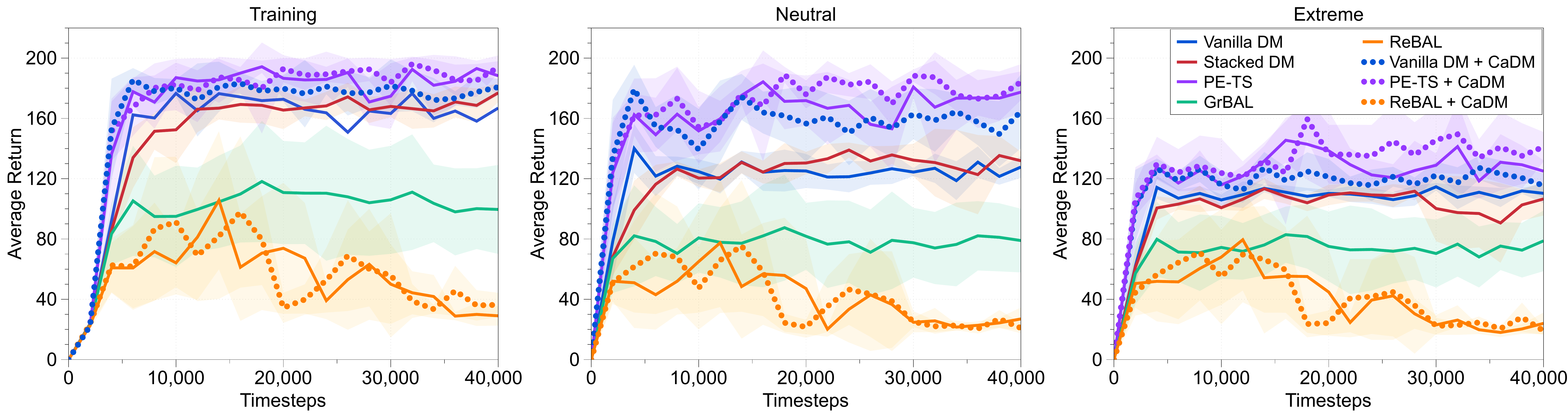} \label{fig:mbrl_sup_cartpole}}
\\
\subfigure[Pendulum]
{
\includegraphics[width=0.9\textwidth]{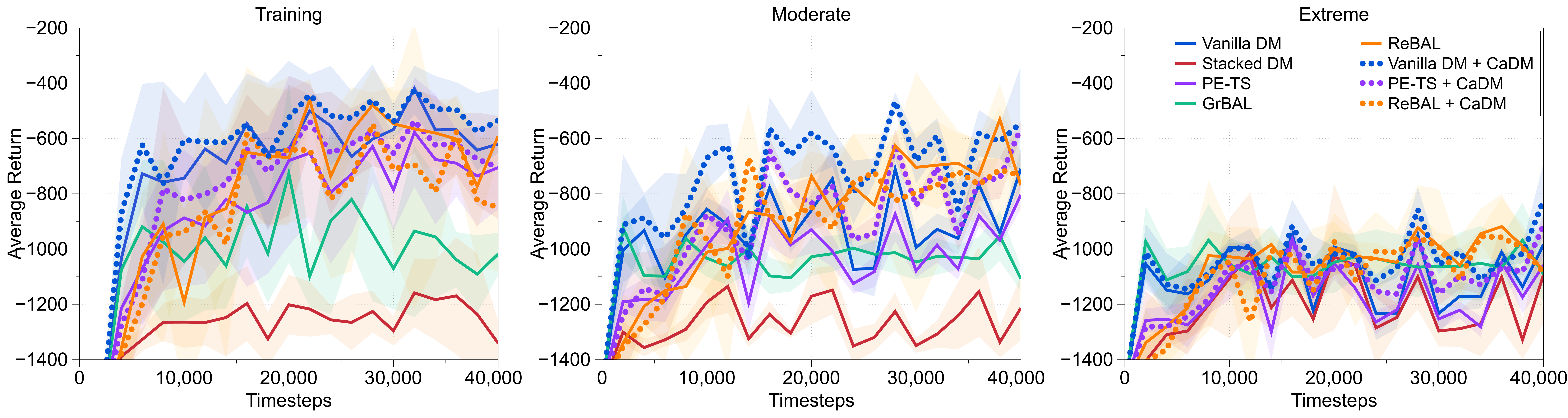} \label{fig:mbrl_sup_pendulum}}
\subfigure[HalfCheetah]
{
\includegraphics[width=0.9\textwidth]{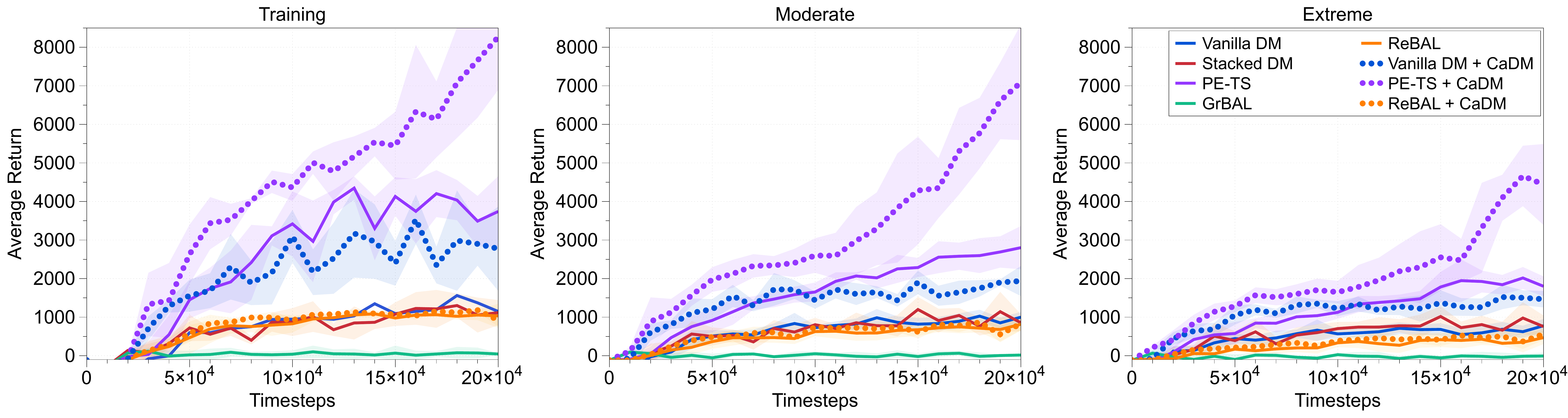} \label{fig:mbrl_sup_halfcheetah}}
\\
\subfigure[Ant]
{
\includegraphics[width=0.9\textwidth]{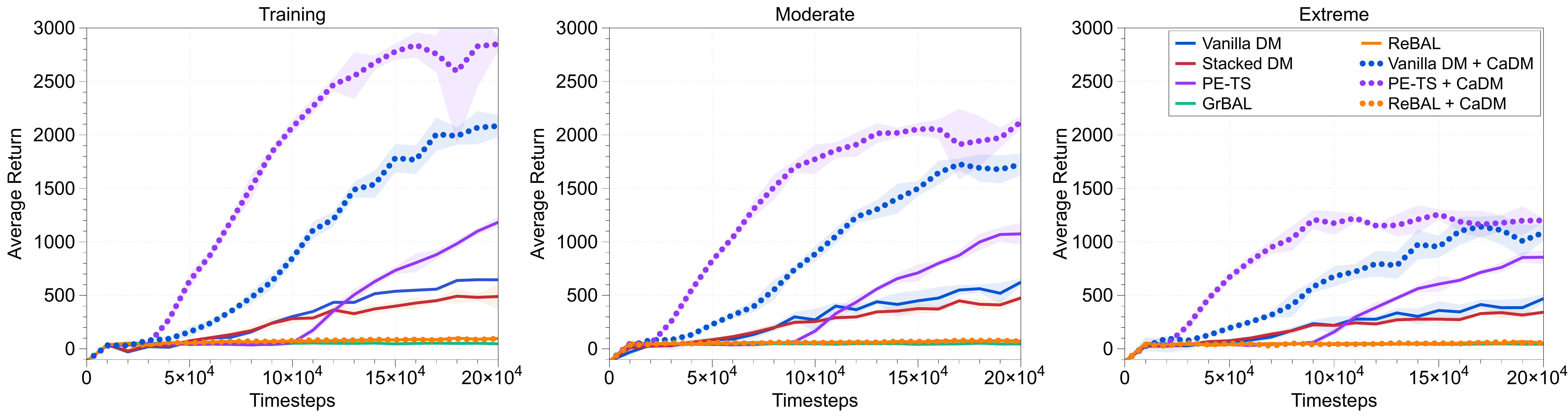} \label{fig:mbrl_sup_ant}}
\caption{
The performance (average returns) of trained dynamics models on (a) CartPole, (b) Pendulum, (c) HalfCheetah, and (d) Ant. The transition dynamics of environments are changing in both training and test environments. 
We remark that test environments consist of moderate and extreme environments, where the former draws environment parameters from a closer (yet different) range to the training one, compared to the latter. The solid line and shaded regions represent the mean and standard deviation, respectively, across five runs.}
\label{fig:mbrl_sup_first}
\end{figure*}

\begin{figure*} [h!] \centering
\subfigure[CrippledHalfCheetah]
{
\includegraphics[width=0.9\textwidth]{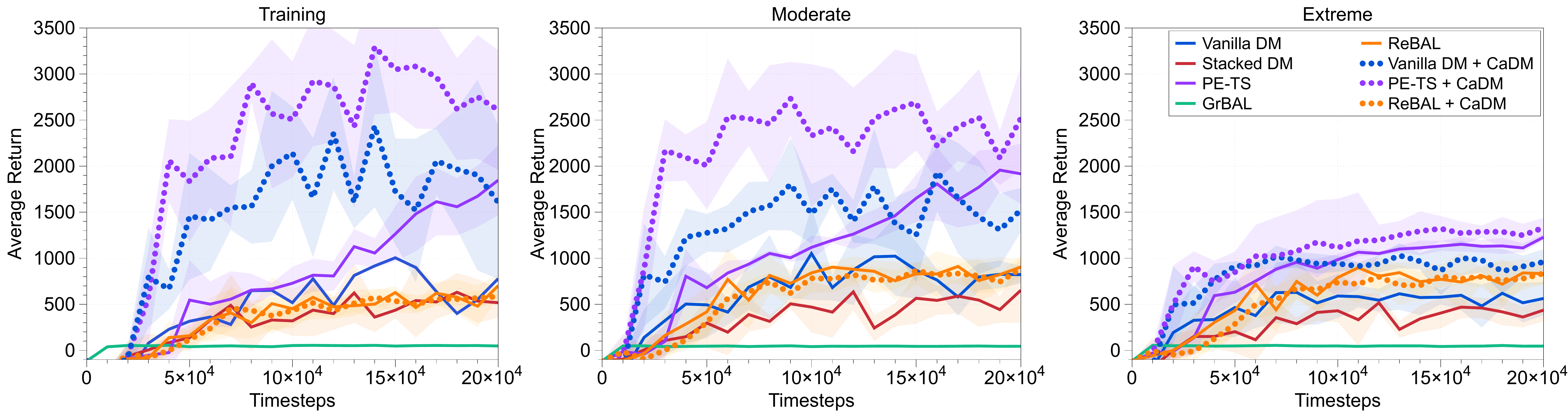} \label{fig:mbrl_sup_cripple_halfcheetah}}
\\
\subfigure[SlimHumanoid]
{
\includegraphics[width=0.9\textwidth]{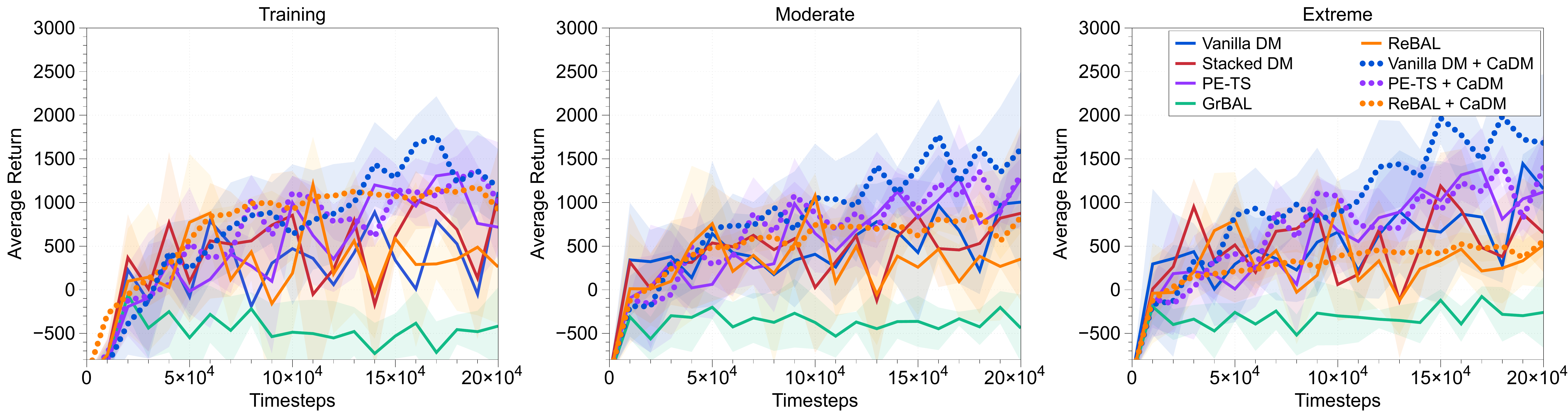} \label{fig:mbrl_sup_slim_humanoid}}
\caption{
The performance (average returns) of trained dynamics models on (a) CrippledHalfCheetah, and (b) SlimHumanoid. The transition dynamics of environments are changing in both training and test environments. 
We remark that test environments consist of moderate and extreme environments, where the former draws environment parameters from a closer (yet different) range to the training one, compared to the latter. The solid line and shaded regions represent the mean and standard deviation, respectively, across five runs.}
\label{fig:mbrl_sup_second}
\end{figure*}

\newpage
\mbox{}
\newpage

\section{Learning Curves for Model-Free RL}
\begin{figure*} [h!] \centering
\subfigure[CartPole]
{
\includegraphics[width=0.9\textwidth]{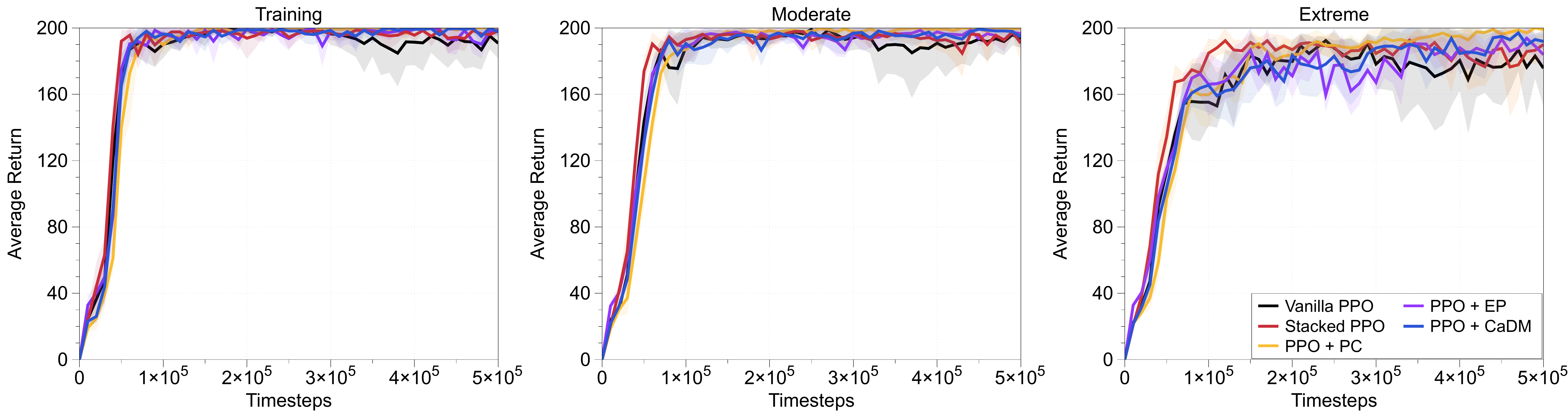} \label{fig:mfrl_sup_cartpole}}
\\
\subfigure[Pendulum]
{
\includegraphics[width=0.9\textwidth]{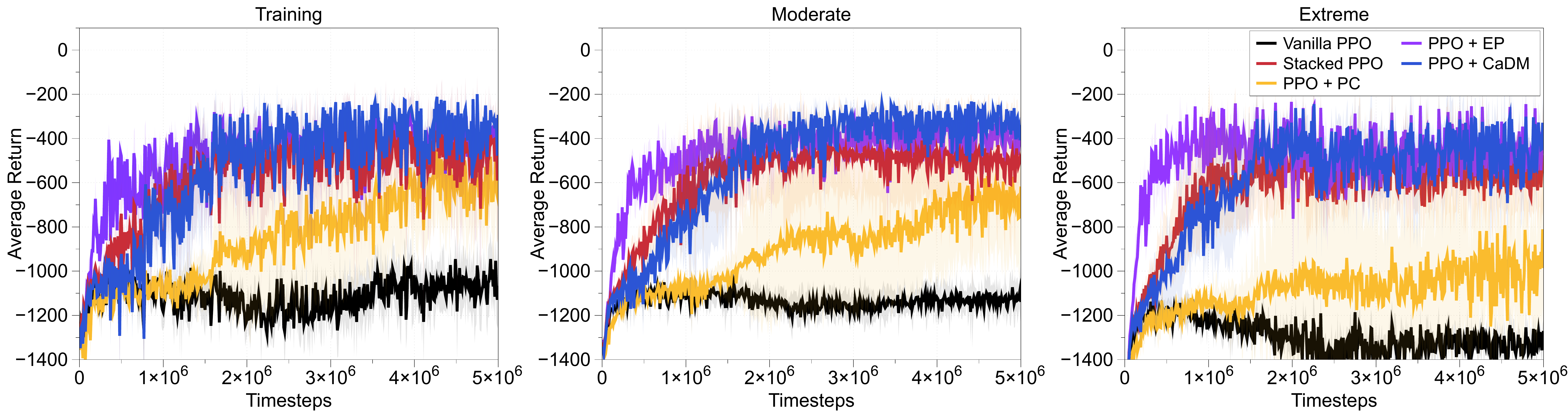} \label{fig:mfrl_sup_pendulum}}
\\
\subfigure[Half-cheetah]
{
\includegraphics[width=0.9\textwidth]{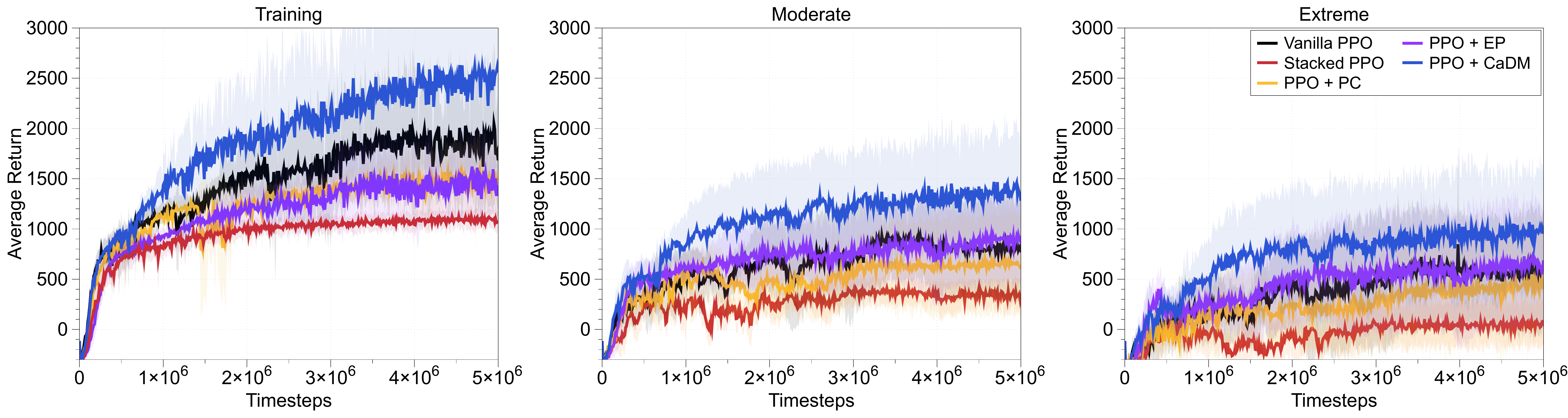} \label{fig:mfrl_sup_halfcheetah}}
\\
\subfigure[Ant]
{
\includegraphics[width=0.9\textwidth]{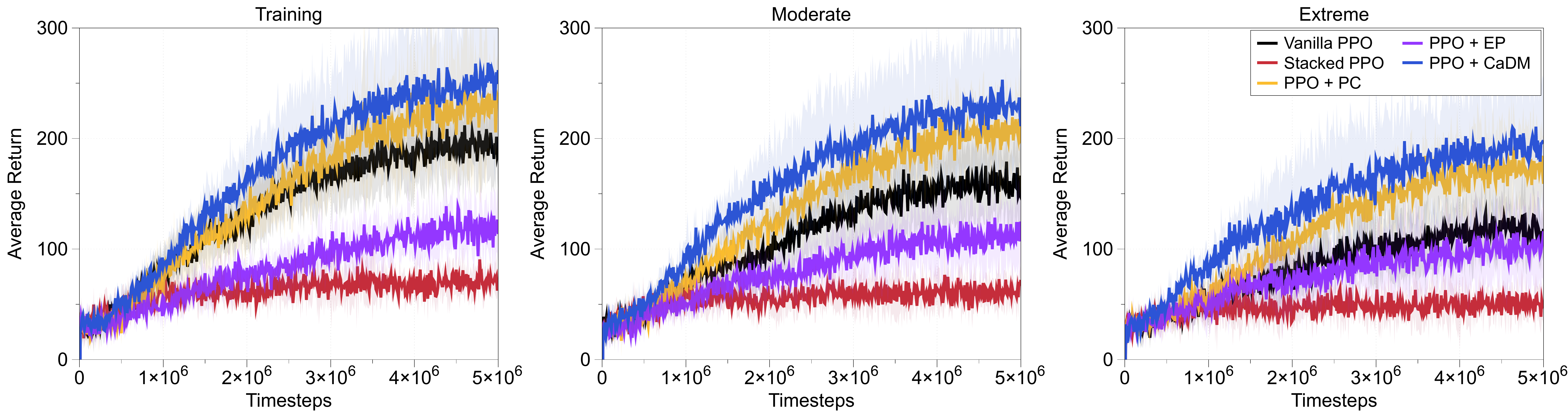} \label{fig:mfrl_sup_ant}}
\caption{
The performance (average returns) of model-free methods on (a) CartPole, (b) Pendulum, (c) Half-cheetah and (d) Ant. The transition dynamics of environments are changing in both training and test environments. We remark that test environments consist of moderate and extreme environments, where the former draws environment parameters from a closer (yet different) range to the training one, compared to the latter. The solid line and shaded regions represent the mean and standard deviation, respectively, across five runs.}
\label{fig:mfrl_sup_first}
\end{figure*}

\begin{figure*} [h!] \centering
\subfigure[CrippledHalfCheetah]
{
\includegraphics[width=0.9\textwidth]{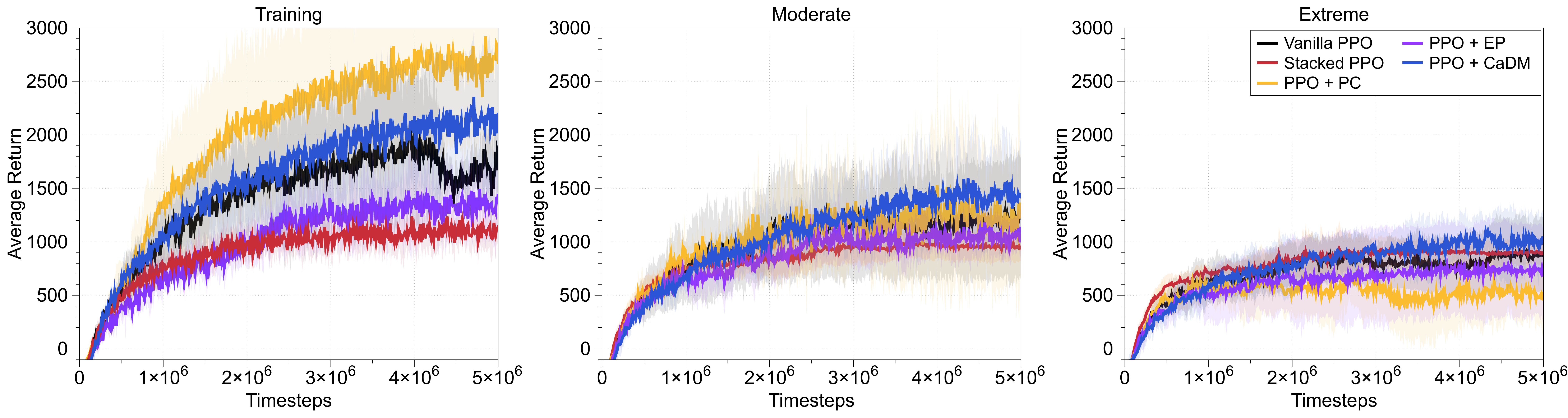} \label{fig:mfrl_sup_cripple_halfcheetah}}
\\
\subfigure[SlimHumanoid]
{
\includegraphics[width=0.9\textwidth]{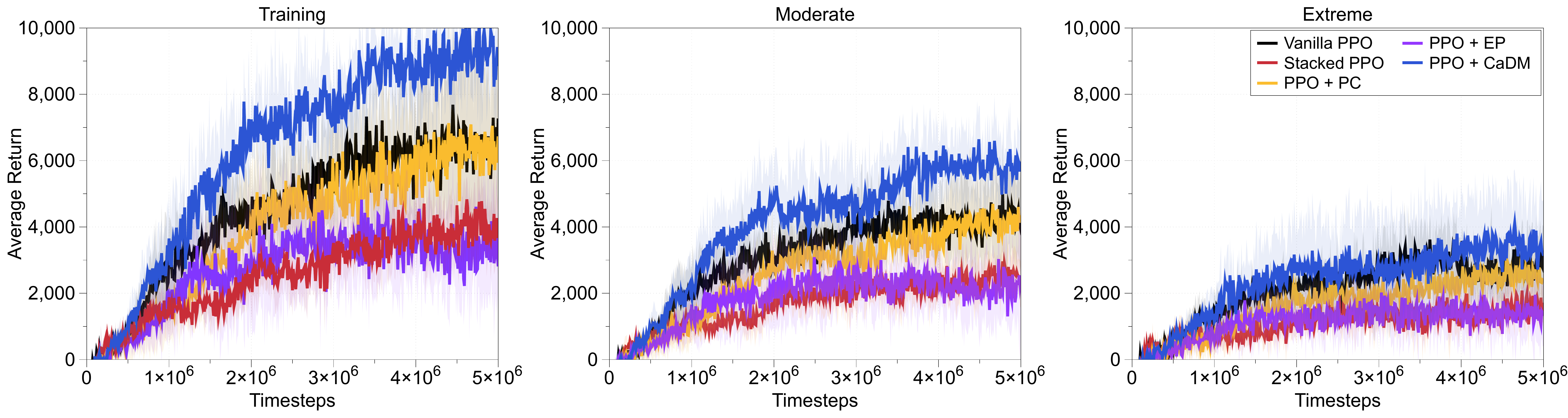} \label{fig:mfrl_sup_slim_humanoid}}
\caption{
The performance (average returns) of model-free methods on (a) CrippledHalfCheetah, and (b) SlimHumanoid. The transition dynamics of environments are changing in both training and test environments. We remark that test environments consist of moderate and extreme environments, where the former draws environment parameters from a closer (yet different) range to the training one, compared to the latter. The solid line and shaded regions represent the mean and standard deviation, respectively, across five runs.}
\label{fig:mfrl_sup_second}
\end{figure*}

\newpage
\mbox{}
\newpage

\section{Effects of Prediction Loss}

\begin{figure*} [h!] \centering
\subfigure[CartPole]
{
\includegraphics[width=0.9\textwidth]{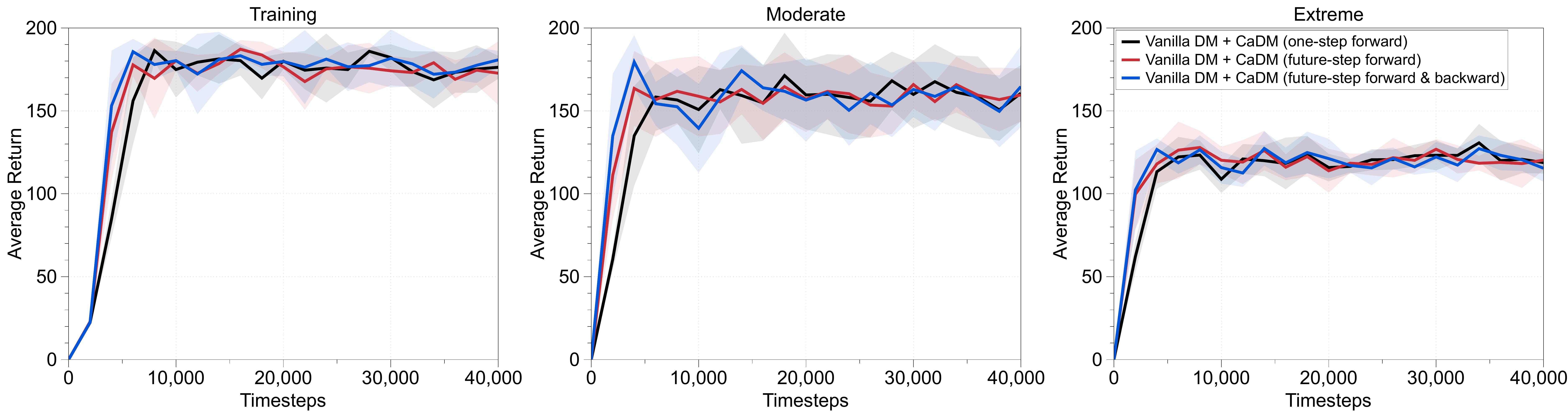} \label{fig:cartpole_sup_ablation}}
\\
\subfigure[Pendulum]
{
\includegraphics[width=0.9\textwidth]{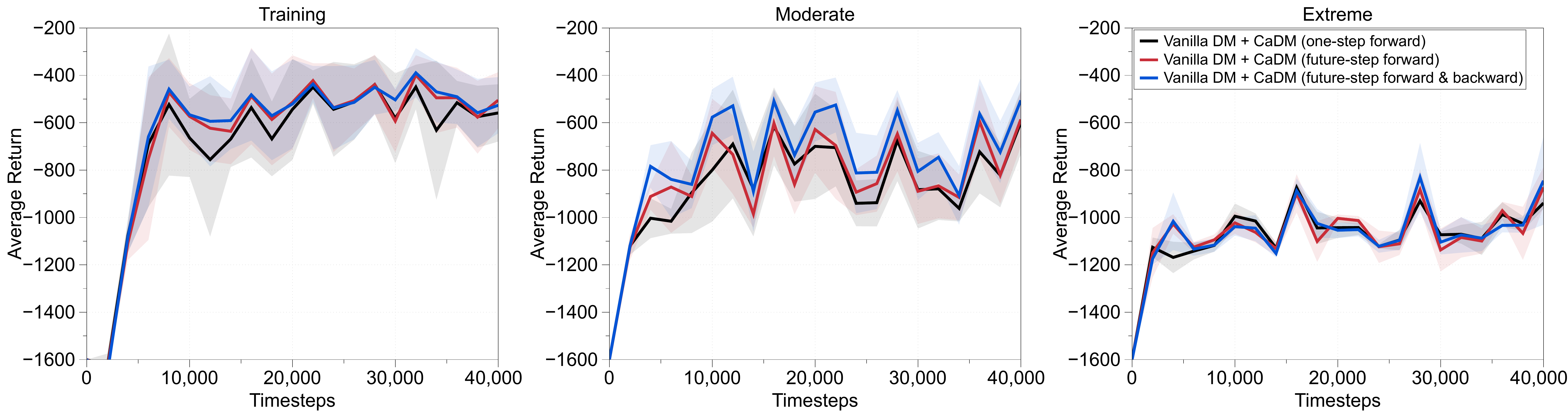} \label{fig:pendulum_sup_ablation}}
\\
\subfigure[Half-cheetah]
{
\includegraphics[width=0.9\textwidth]{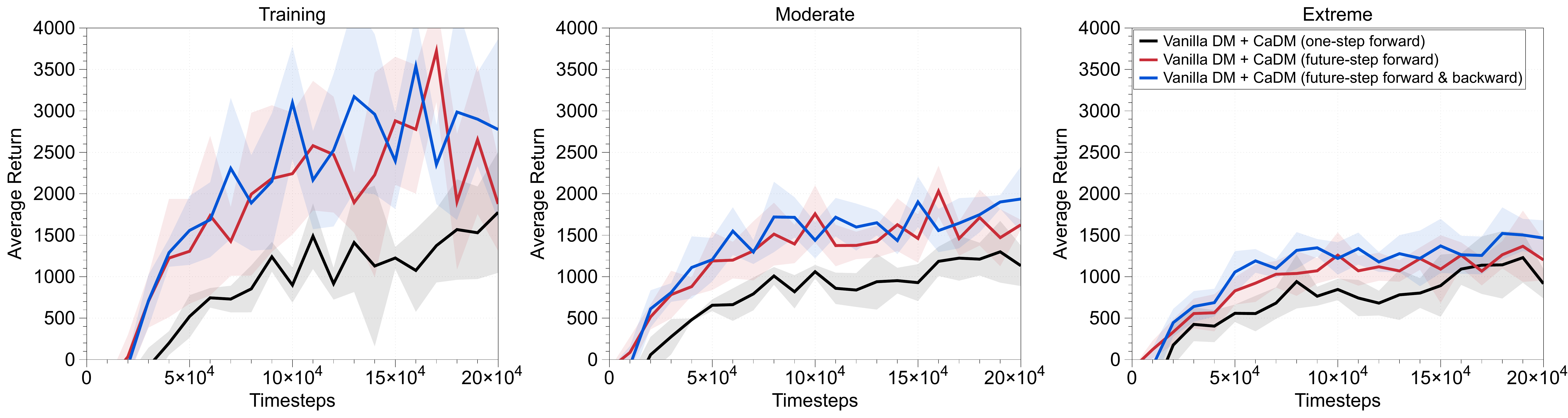} \label{fig:halfcheetah_sup_ablation}}
\\
\subfigure[Ant]
{
\includegraphics[width=0.9\textwidth]{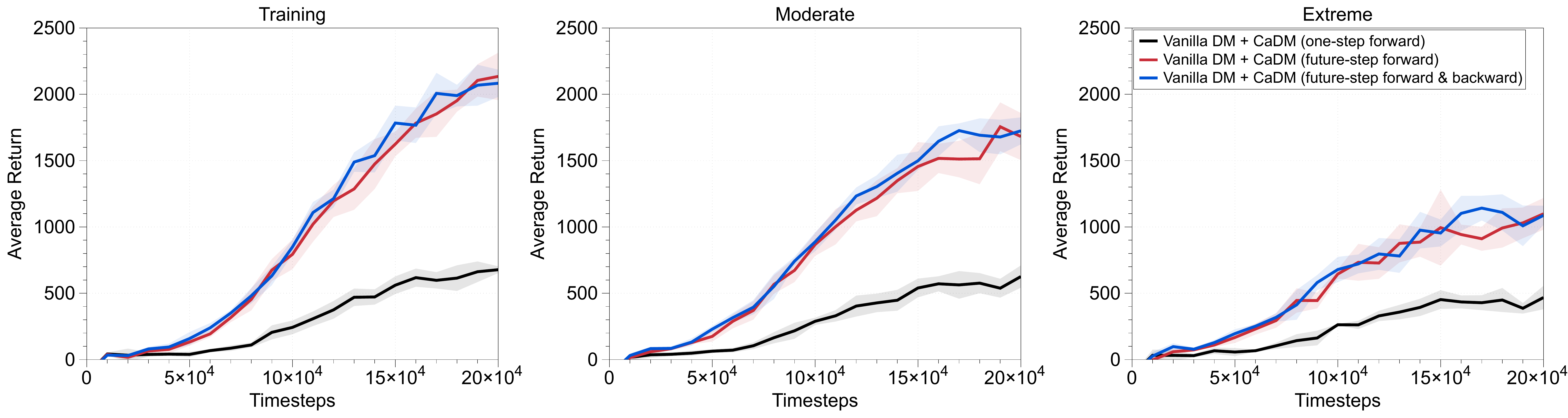} \label{fig:ant_sup_ablation}}
\caption{
The performance (average returns) of dynamics models optimized by variants of the proposed prediction objective in (1) on (a) CartPole, (b) Pendulum, (c) HalfCheetah, and (d) Ant. The transition dynamics of environments are changing in both training and test environments. We remark that test environments consist of moderate and extreme environments, where the former draws environment parameters from a closer (yet different) range to the training one, compared to the latter. The solid line and shaded regions represent the mean and standard deviation, respectively, across five runs.}
\label{fig:prediction_loss_ablation_first}
\end{figure*}

\begin{figure*} [h!] \centering
\subfigure[CrippledHalfCheetah]
{
\includegraphics[width=0.9\textwidth]{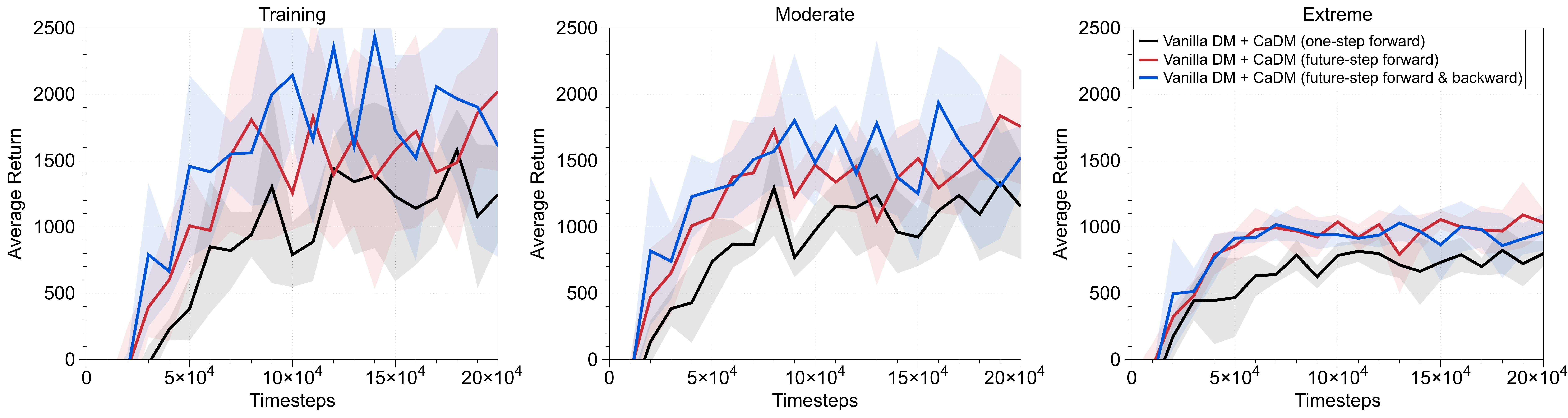} \label{fig:cripple_halfcheetah_sup_ablation}}
\\
\subfigure[SlimHumanoid]
{
\includegraphics[width=0.9\textwidth]{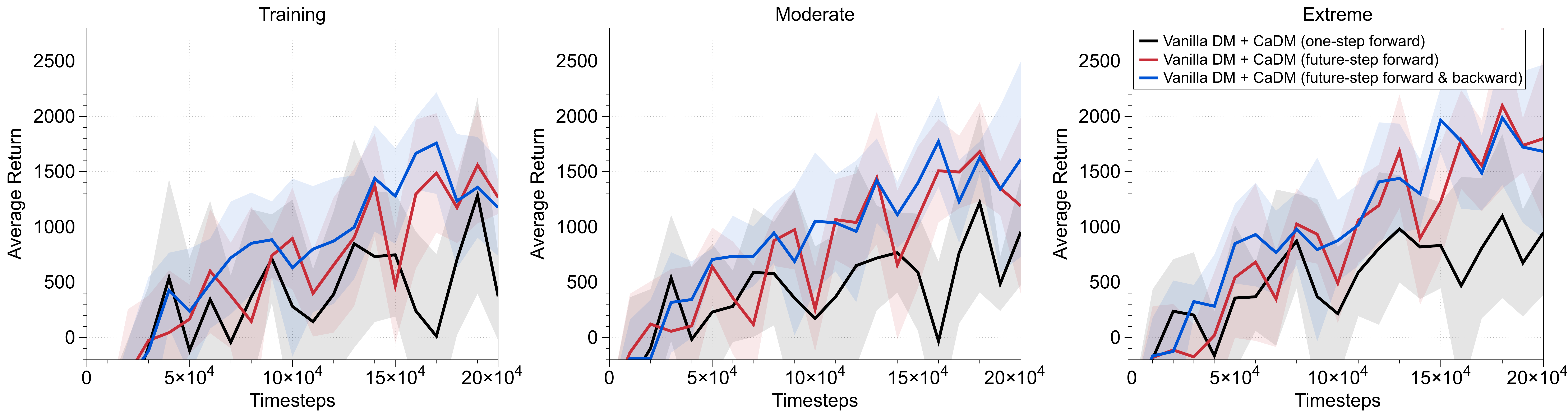} \label{fig:slim_humanoid_sup_ablation}}
\caption{
The performance (average returns) of dynamics models optimized by variants of the proposed prediction objective in (1) on (a) CrippledHalfCheetah, and (b) SlimHumanoid. The transition dynamics of environments are changing in both training and test environments. We remark that test environments consist of moderate and extreme environments, where the former draws environment parameters from a closer (yet different) range to the training one, compared to the latter. The solid line and shaded regions represent the mean and standard deviation, respectively, across five runs.}
\label{fig:prediction_loss_ablation_second}
\end{figure*}

\newpage
\mbox{}
\newpage

\section{Prediction Error}

\begin{figure*} [h!] \centering
\subfigure[CartPole]
{
\includegraphics[width=0.42\textwidth]{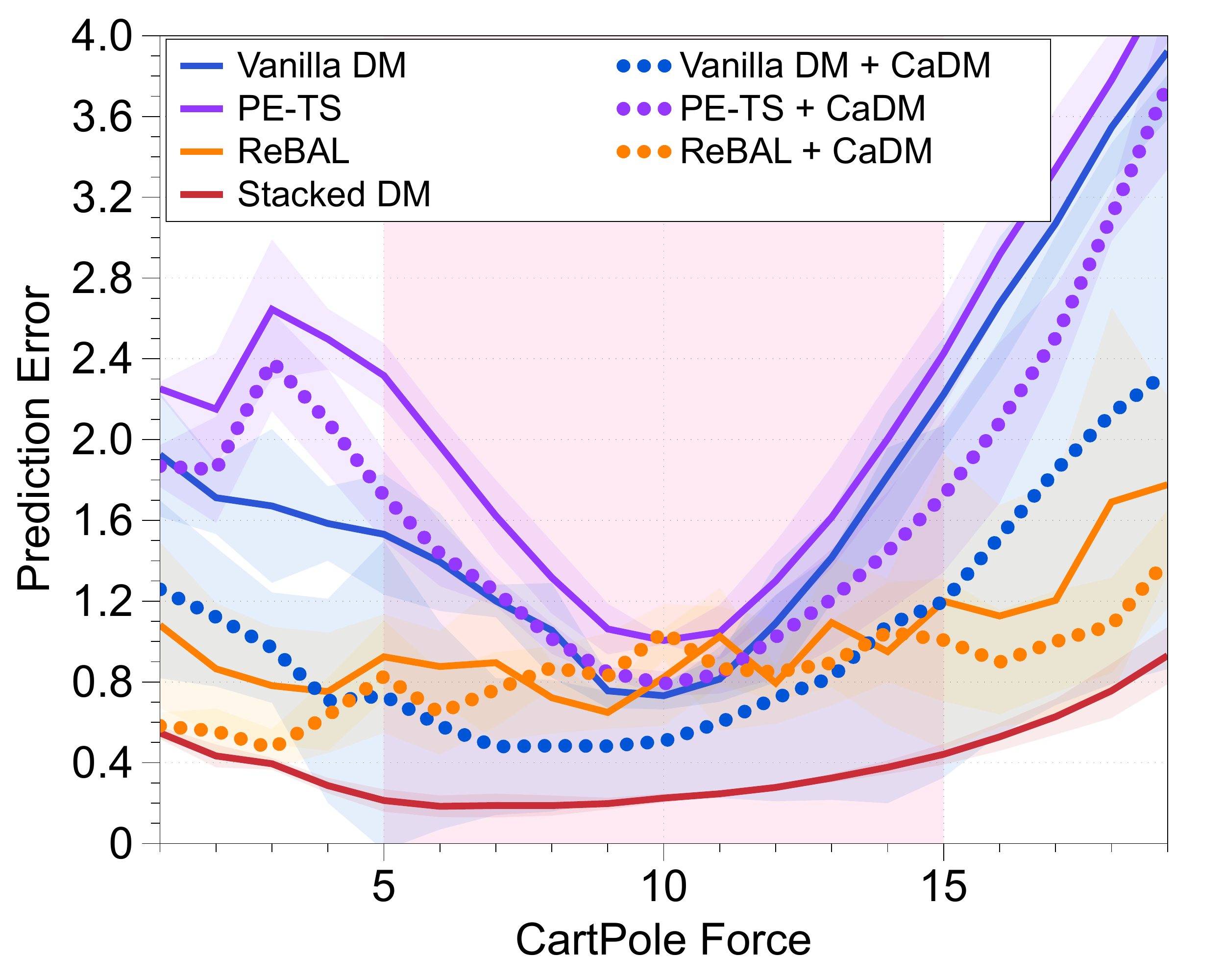} \label{fig:cartpole_sup_prediction_error}}
\,
\subfigure[Pendulum]
{
\includegraphics[width=0.42\textwidth]{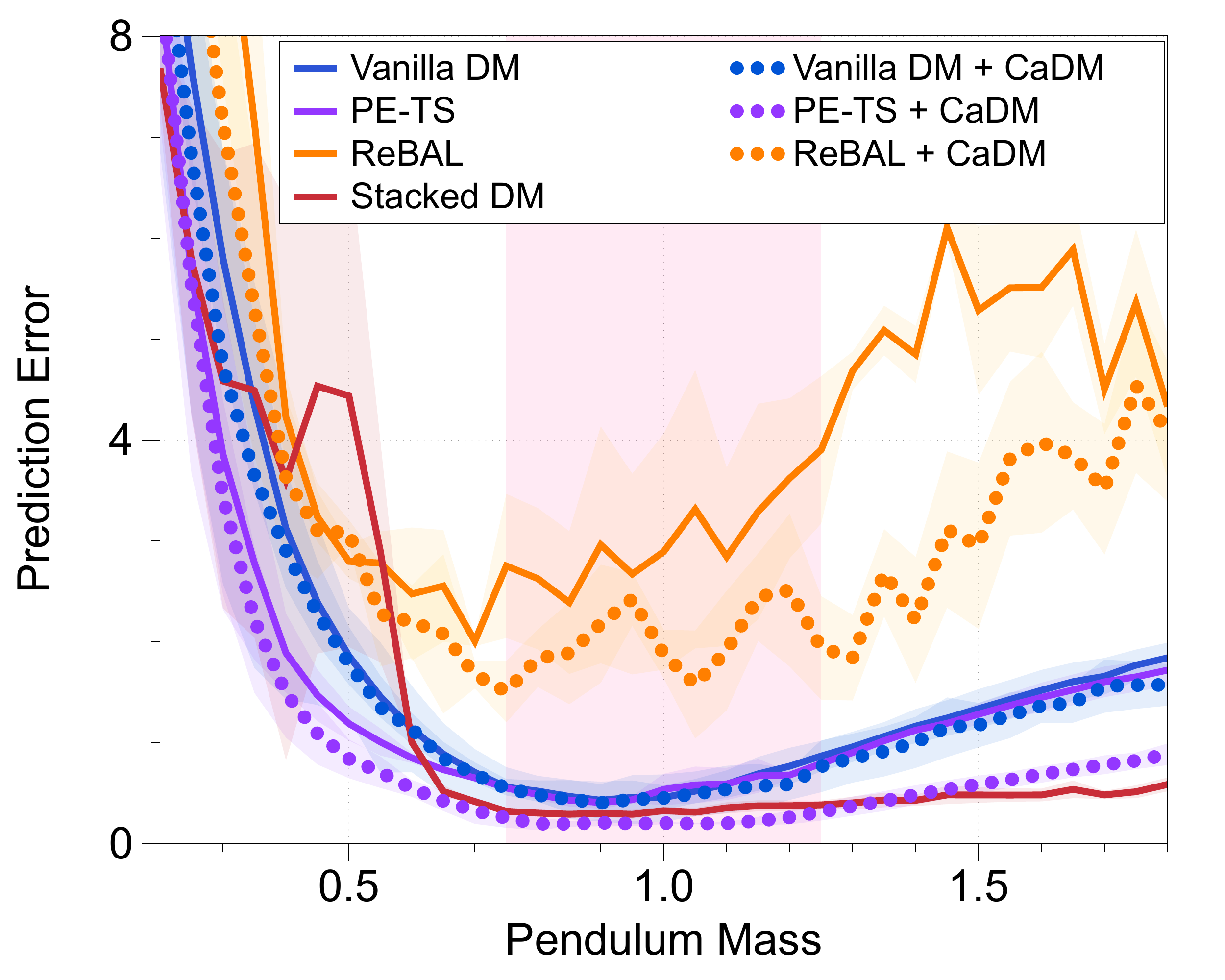} \label{fig:pendulum_sup_prediction_error}}
\\
\subfigure[Half-cheetah]
{
\includegraphics[width=0.42\textwidth]{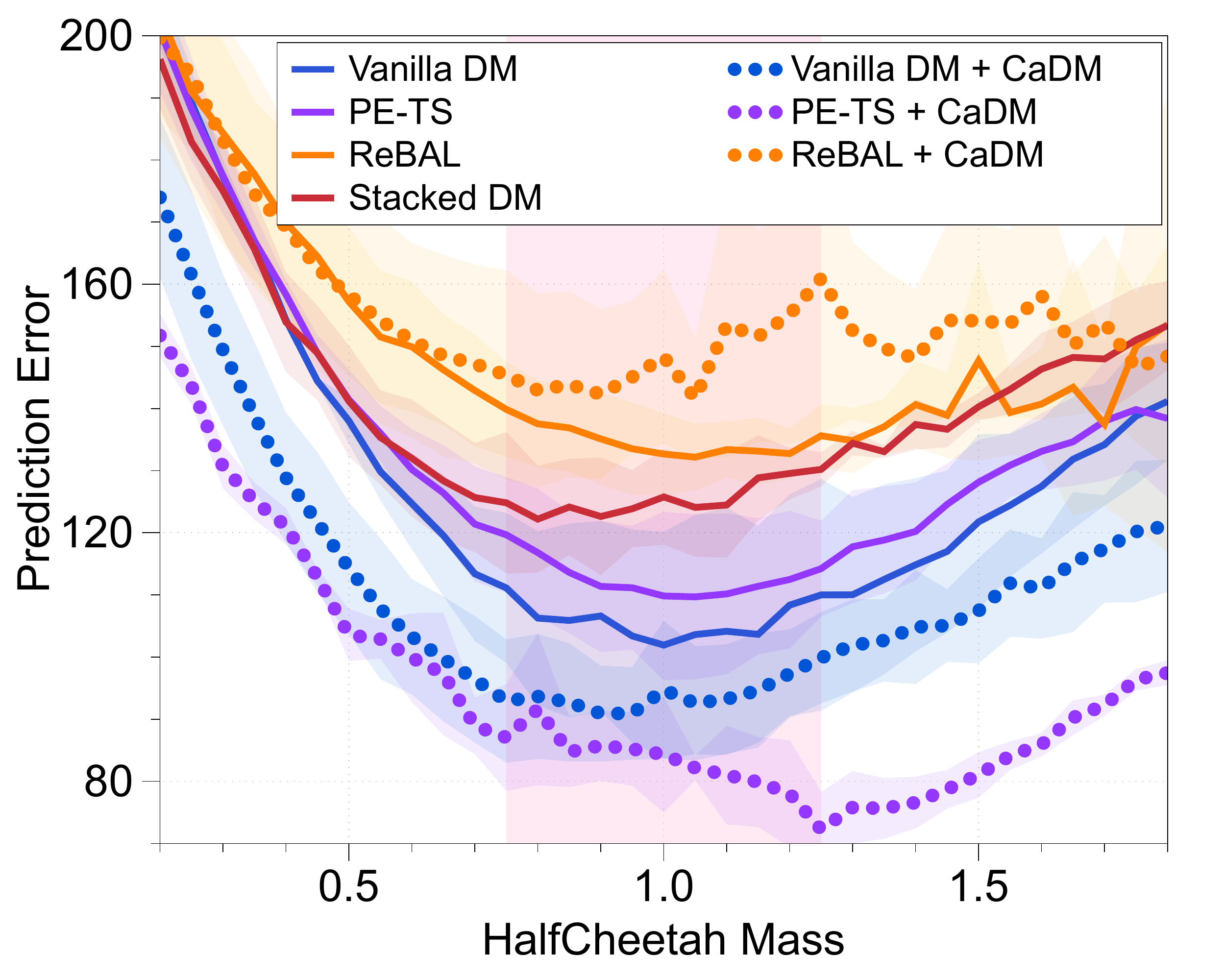} \label{fig:halfcheetah_sup_prediction_error}}
\,
\subfigure[Ant]
{
\includegraphics[width=0.42\textwidth]{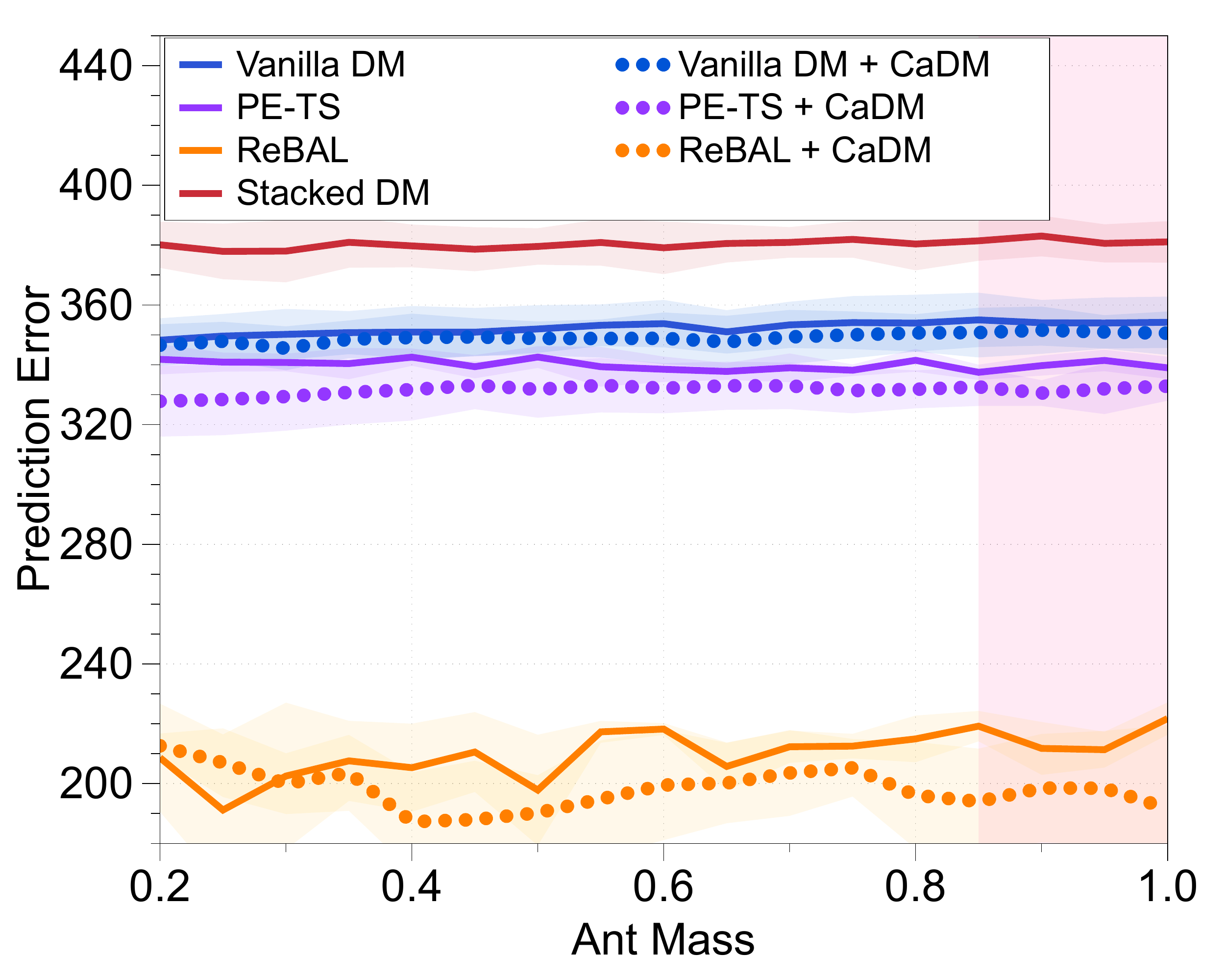} \label{fig:ant_sup_prediction_error}}
\\
\subfigure[CrippledHalfCheetah]
{
\includegraphics[width=0.42\textwidth]{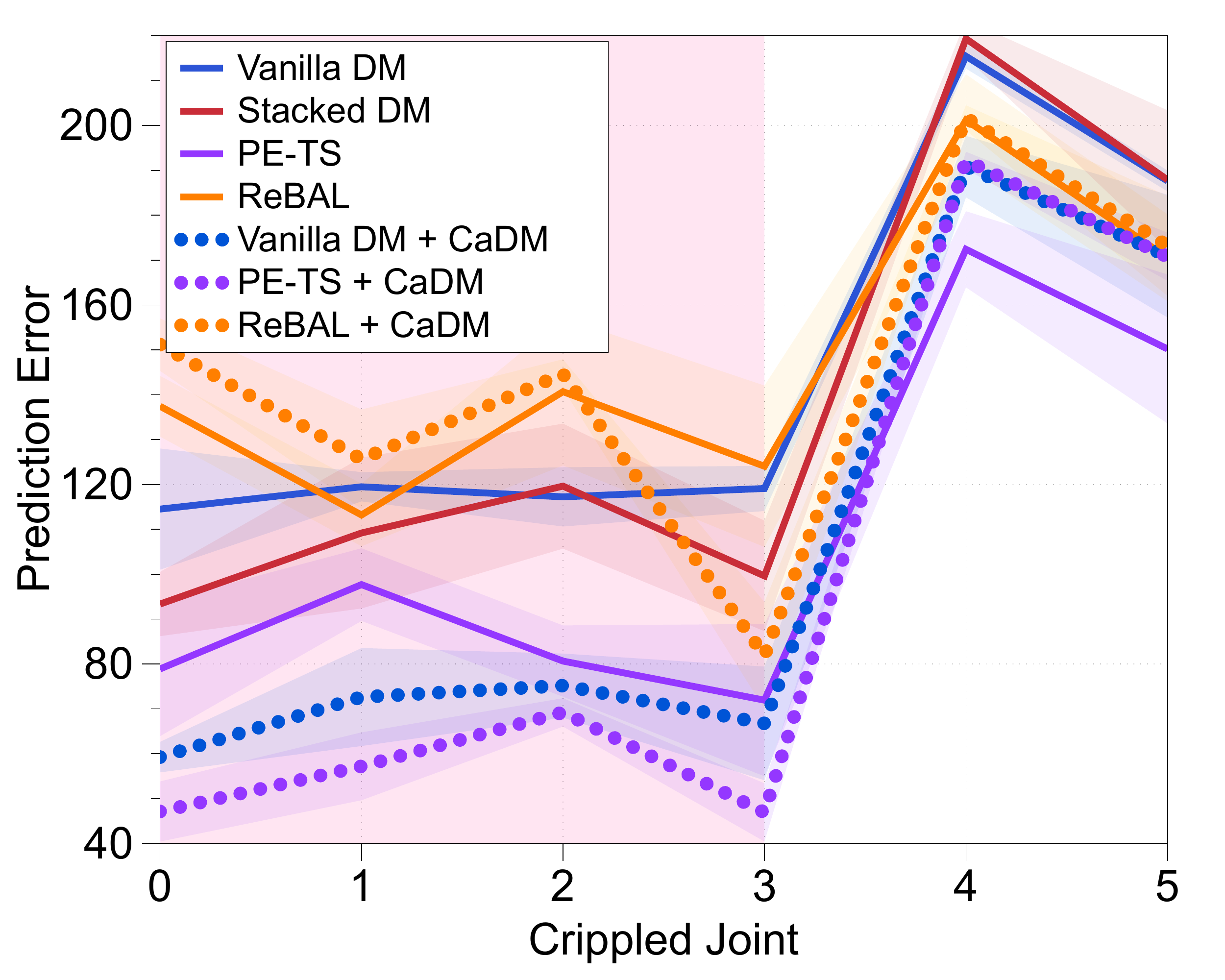} \label{fig:cripple_halfcheetah_sup_prediction_error}}
\,
\subfigure[SlimHumanoid]
{
\includegraphics[width=0.42\textwidth]{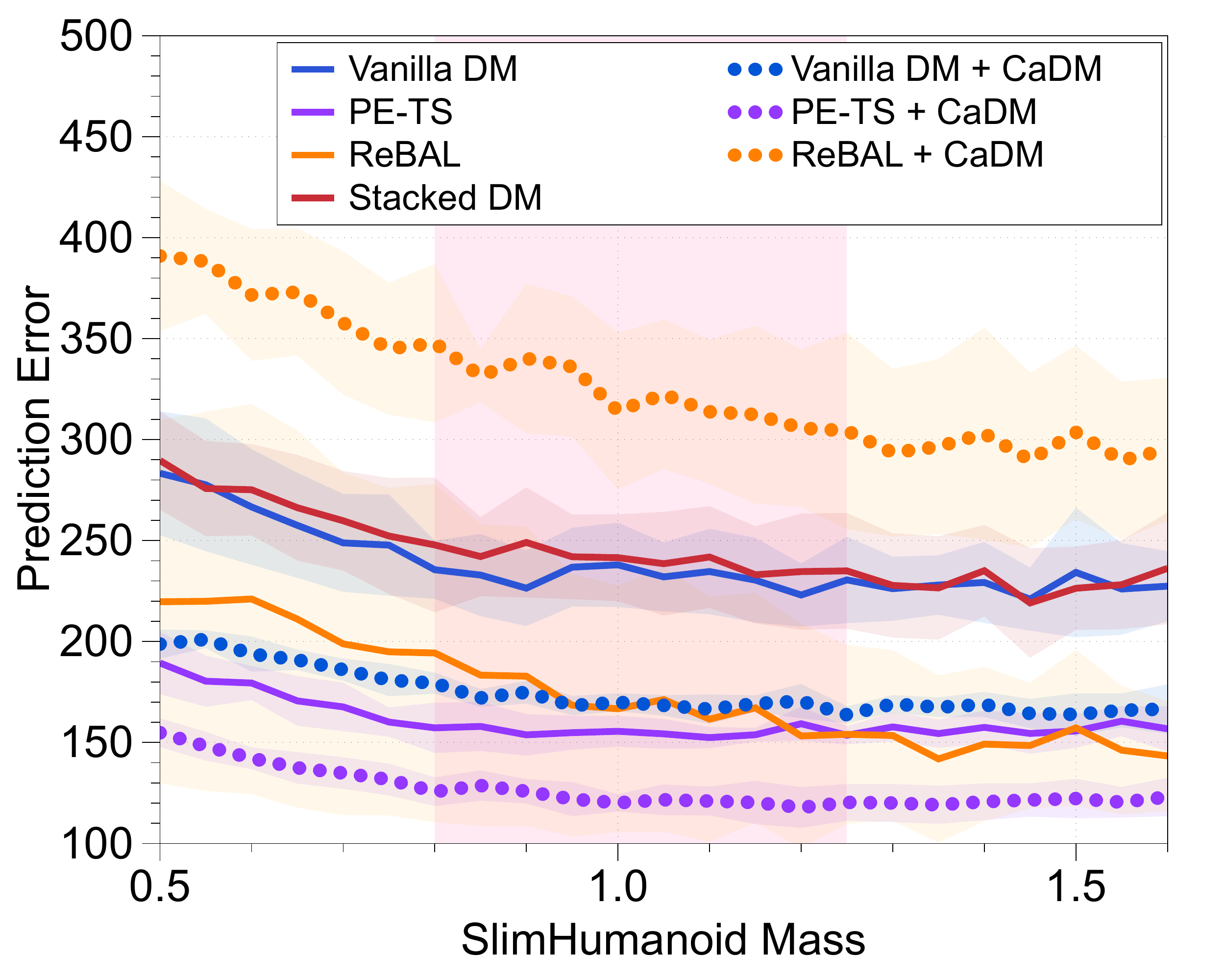} \label{fig:slim_humanoid_sup_prediction_error}}
\caption{
Prediction errors on (a) CartPole, (b) Pendulum, (c) HalfCheetah, (d) Ant, (e) CrippledHalfCheetah, and (f) SlimHumanoid with varying simulation parameters. The solid line and shaded regions represent the mean and standard deviation, respectively, across three runs.}
\label{fig:prediction_error_ablation_first}
\end{figure*}

\section {Embedding Analysis}

\begin{figure*} [h] \centering
\subfigure[CartPole]
{
\includegraphics[width=0.3\textwidth]{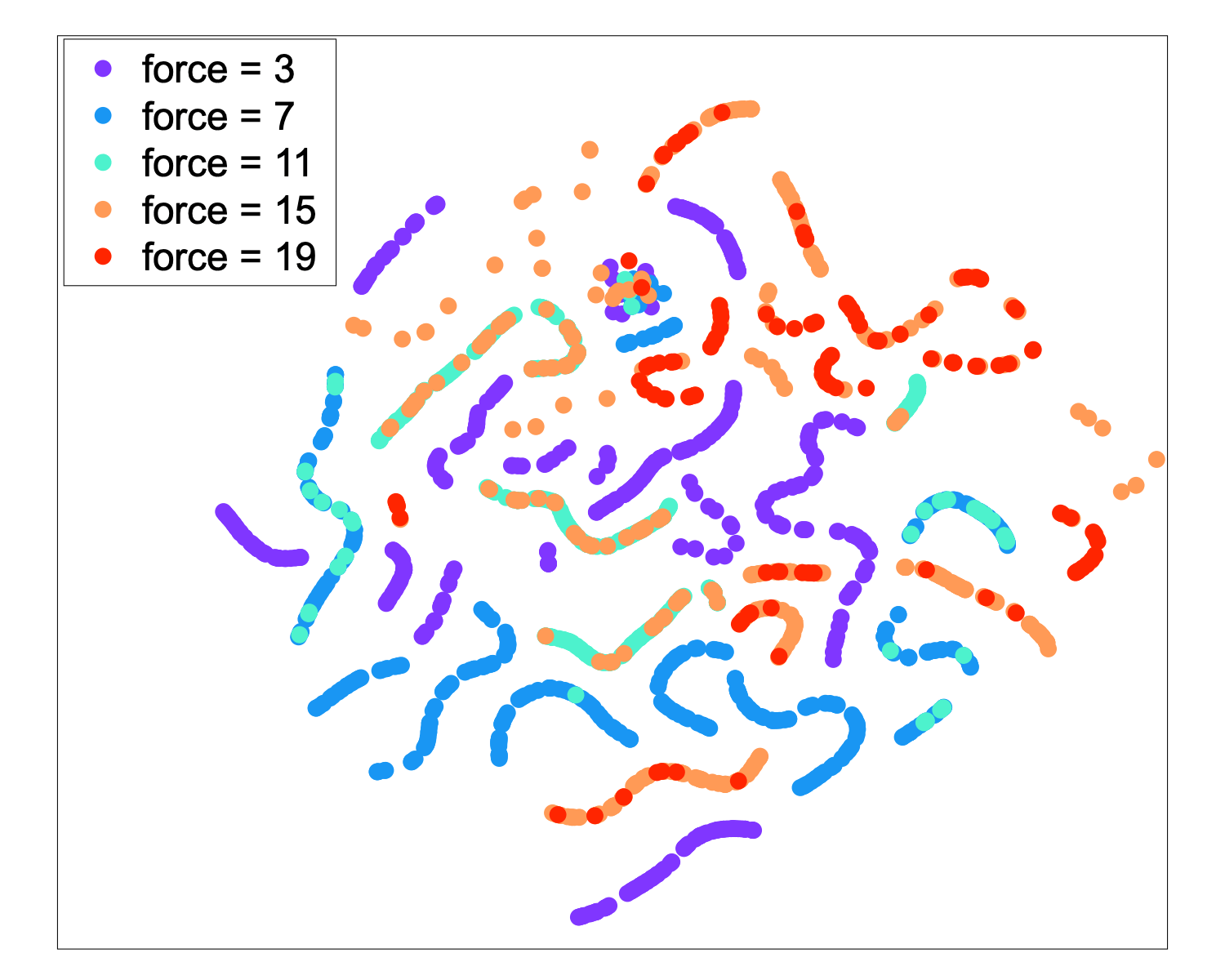} \label{fig:cartpole_ablation_tsne}} 
\,
\subfigure[Pendulum]
{
\includegraphics[width=0.3\textwidth]{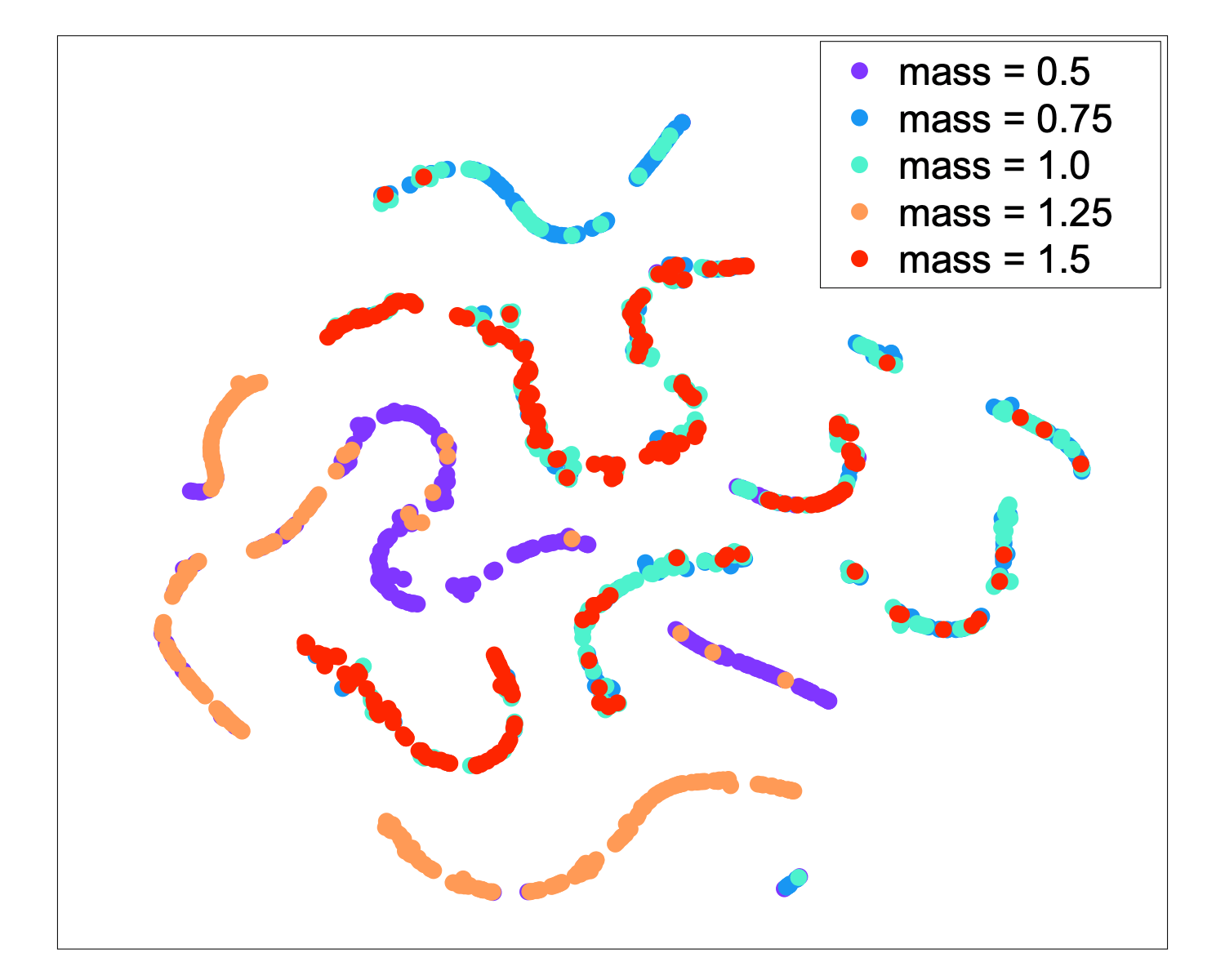} \label{fig:pendulum_ablation_tsne}}
\,
\subfigure[HalfCheetah]
{
\includegraphics[width=0.3\textwidth]{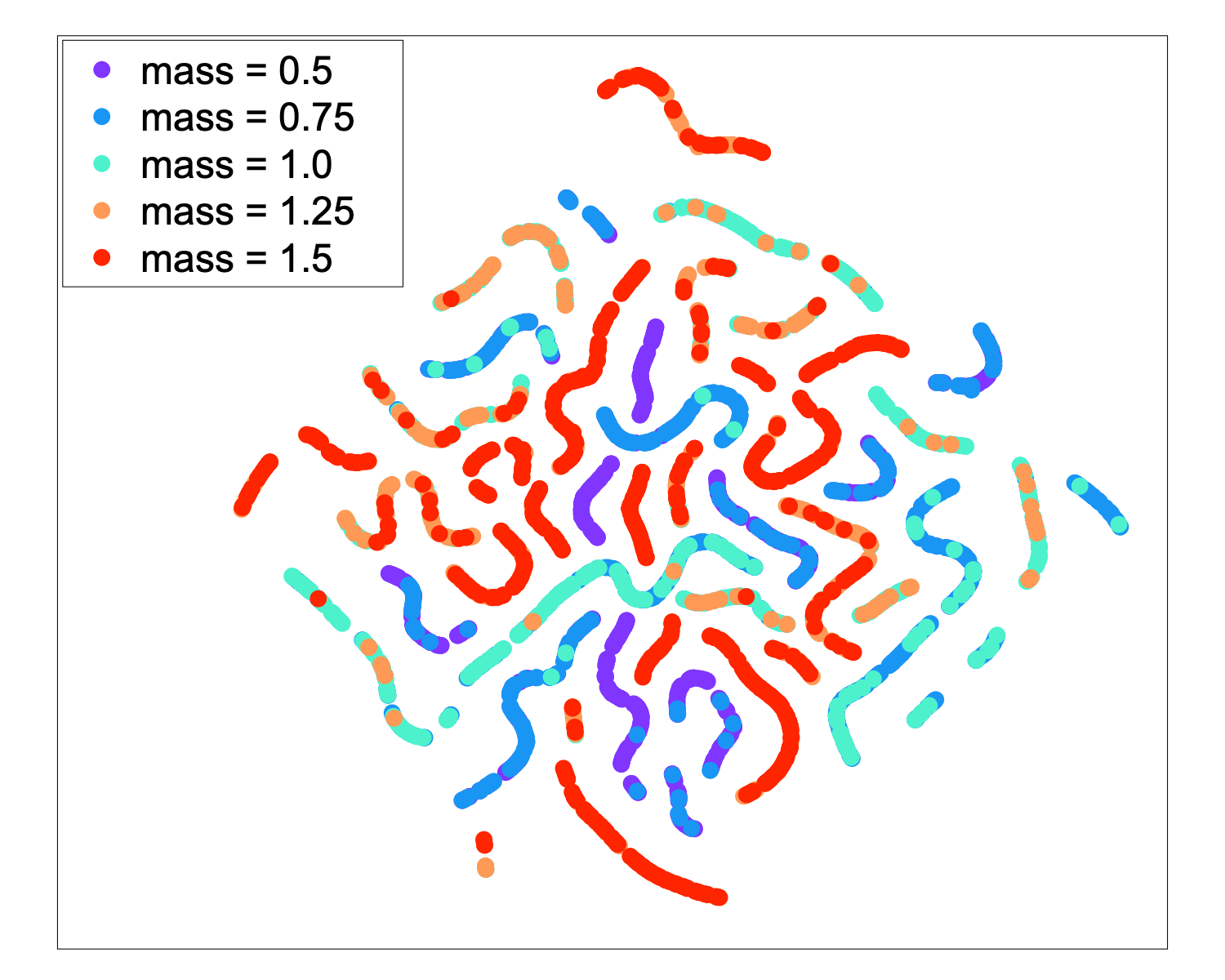} \label{fig:halfcheetah_ablation_tsne}}
\vspace{-0.1in}
\\
\subfigure[Ant]
{
\includegraphics[width=0.3\textwidth]{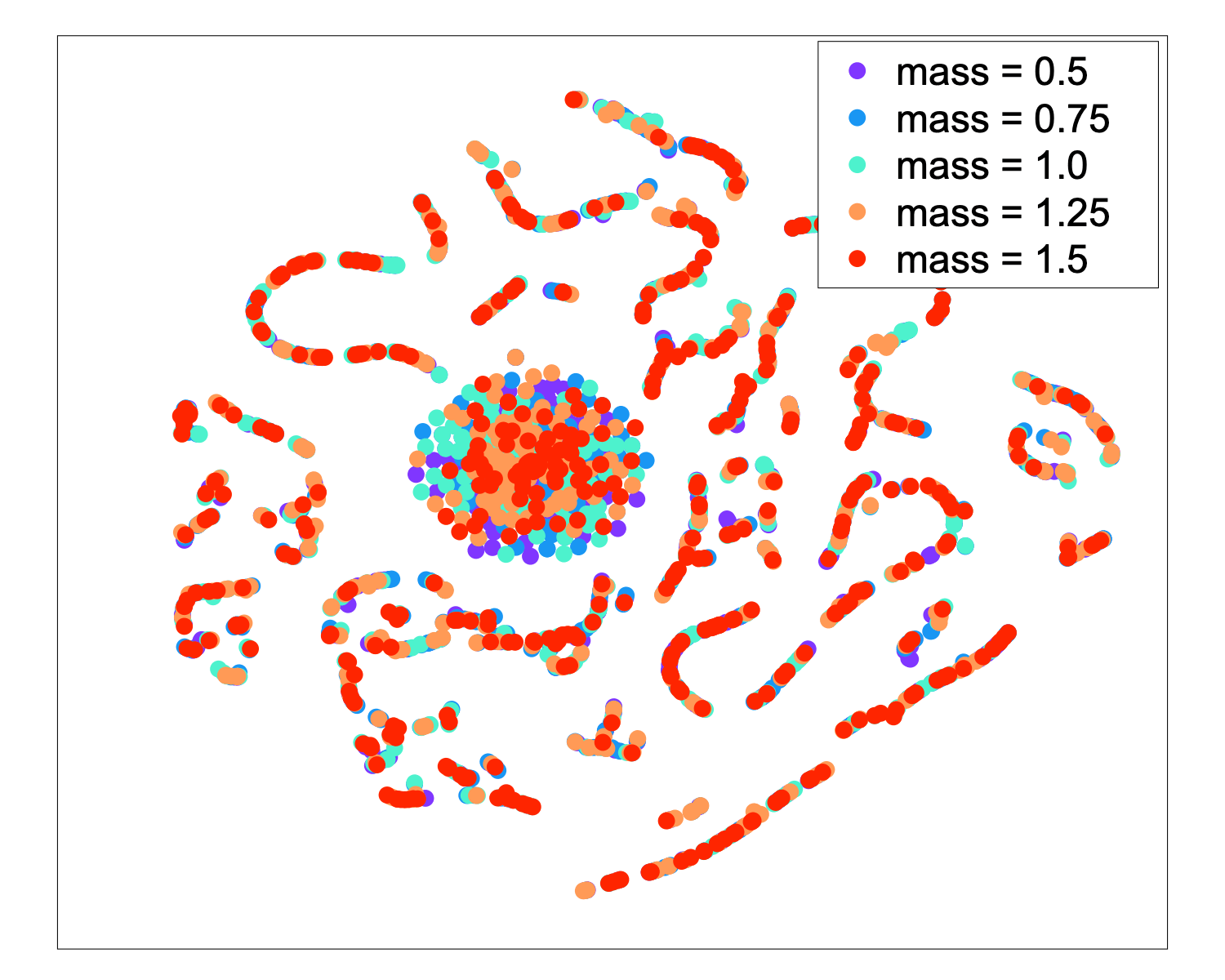} \label{fig:ant_ablation_tsne}} 
\,
\subfigure[CrippledHalfcheetah]
{
\includegraphics[width=0.3\textwidth]{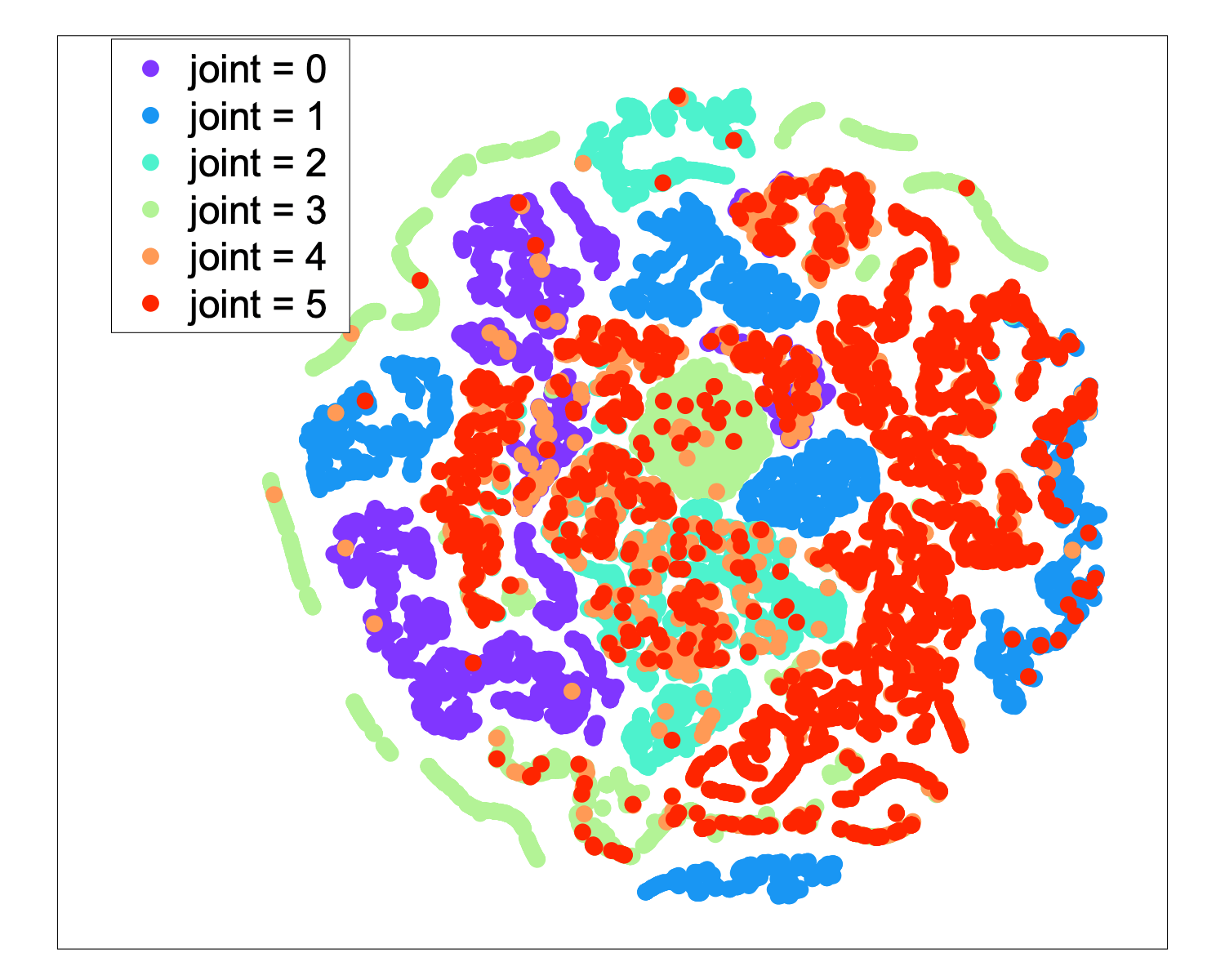} \label{fig:crippled_halfcheetah_ablation_tsne}}
\,
\subfigure[SlimHumanoid]
{
\includegraphics[width=0.3\textwidth]{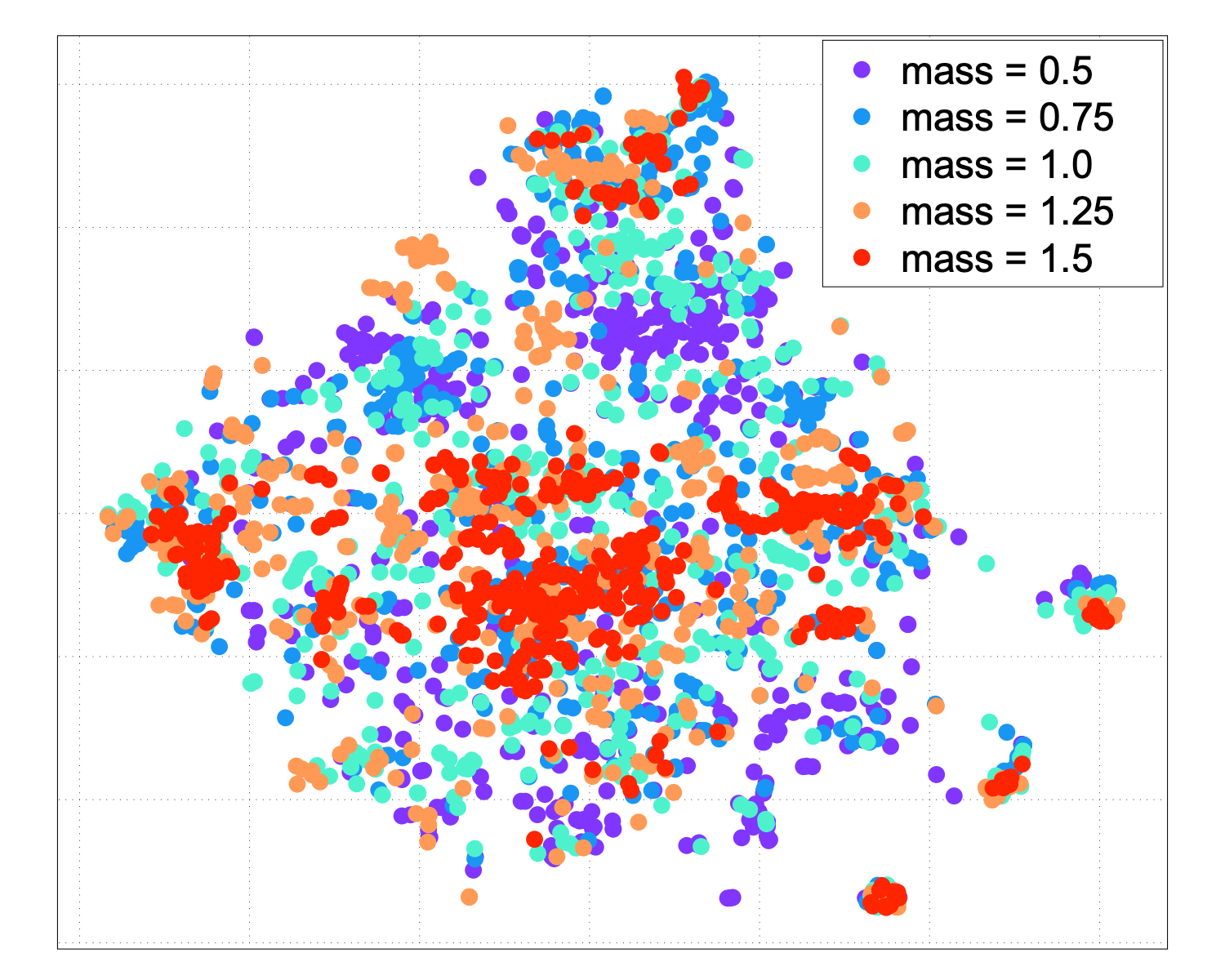} \label{fig:slim_humanoid_ablation_tsne}}
\vspace{-0.1in}
\caption{
t-SNE \cite{maaten2008visualizing} visualization of context latent vectors extracted from trajectories collected in various control tasks. Embedded points from environments with the same parameter have the same color.}
\label{fig:embedding_analysis_tsne}
\vspace{-0.1in}
\end{figure*}

\begin{figure*} [h!] \centering
\subfigure[CartPole]
{
\includegraphics[width=0.3\textwidth]{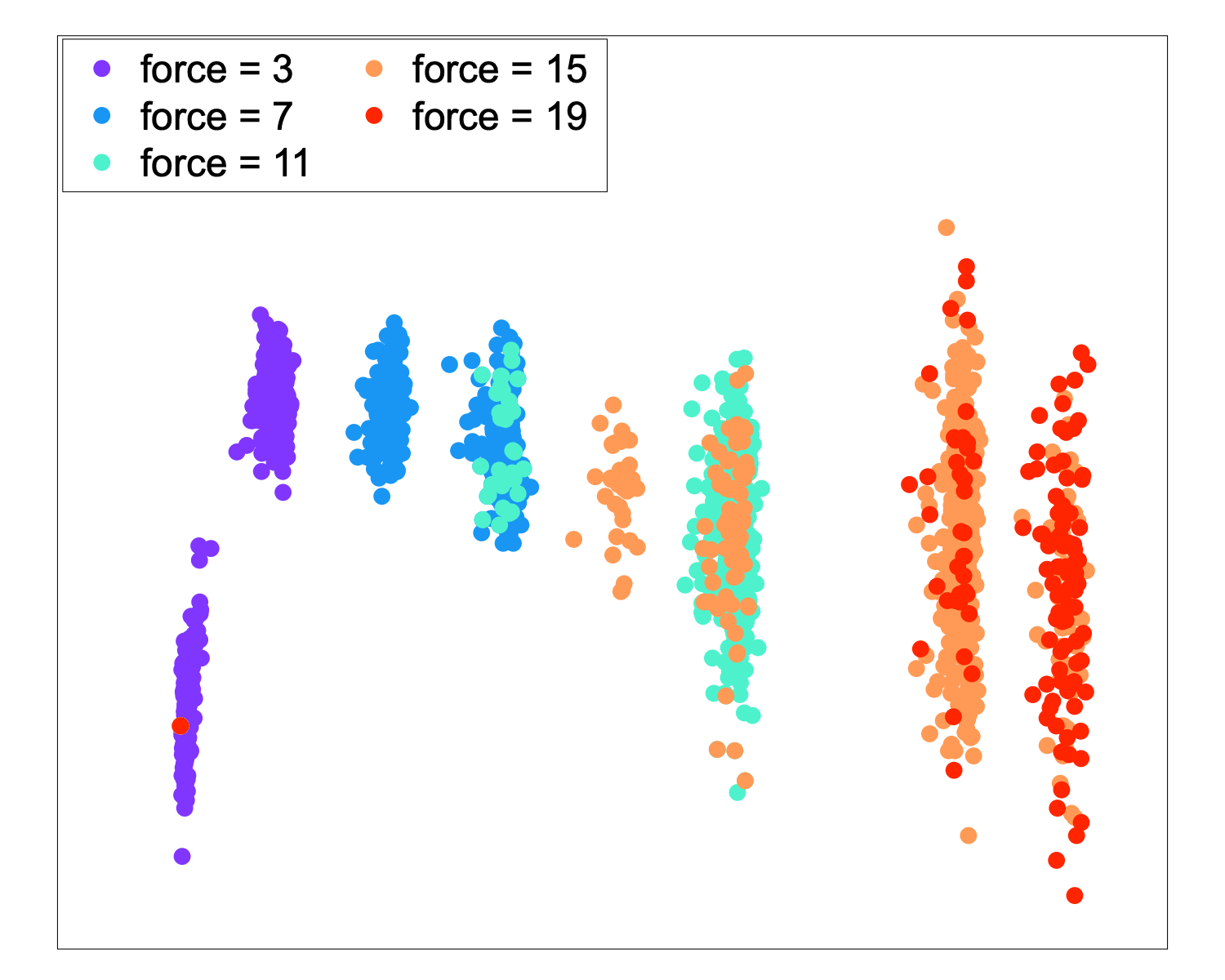} \label{fig:cartpole_ablation_pca}} 
\,
\subfigure[Pendulum]
{
\includegraphics[width=0.3\textwidth]{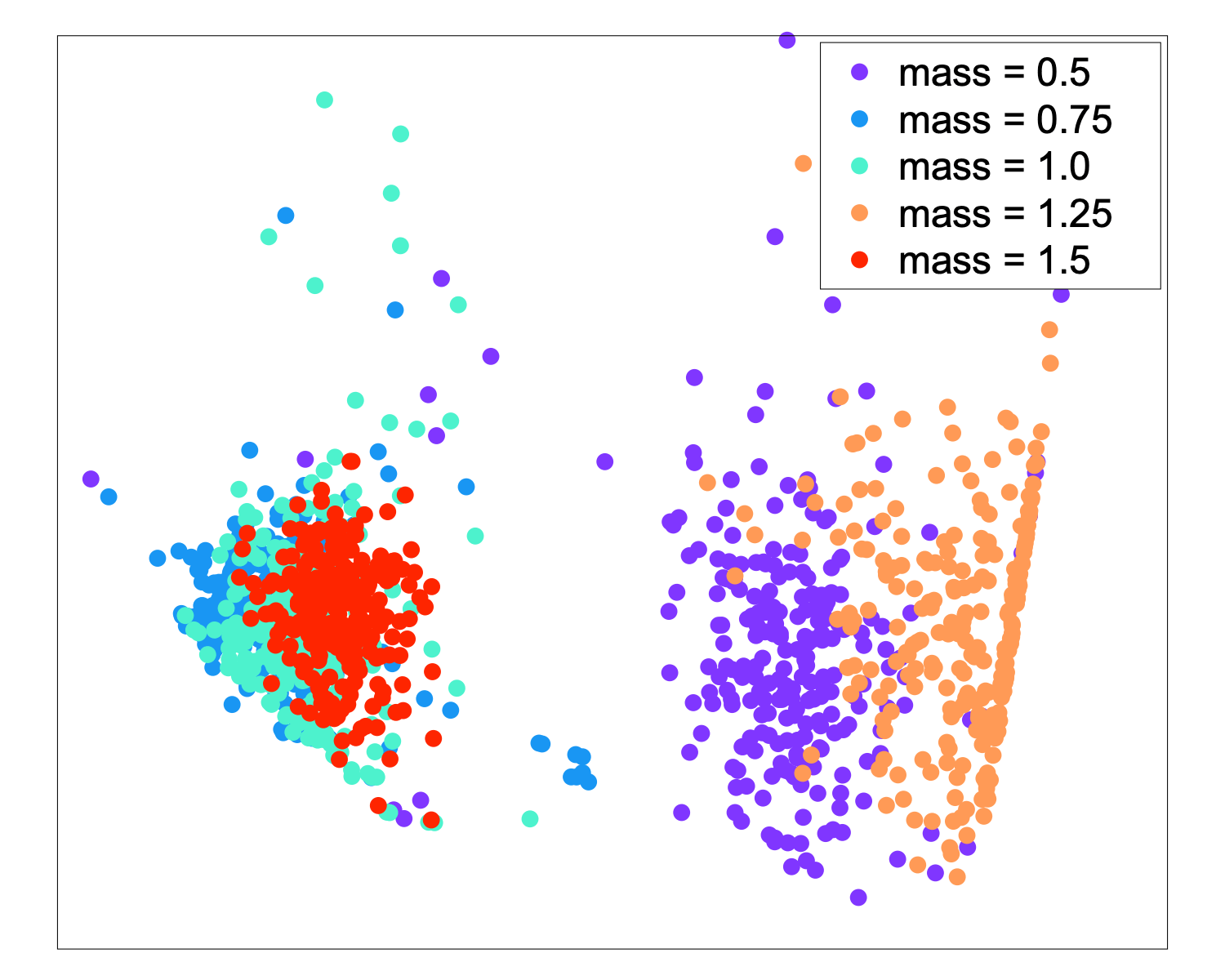} \label{fig:pendulum_ablation_pca}}
\,
\subfigure[HalfCheetah]
{
\includegraphics[width=0.3\textwidth]{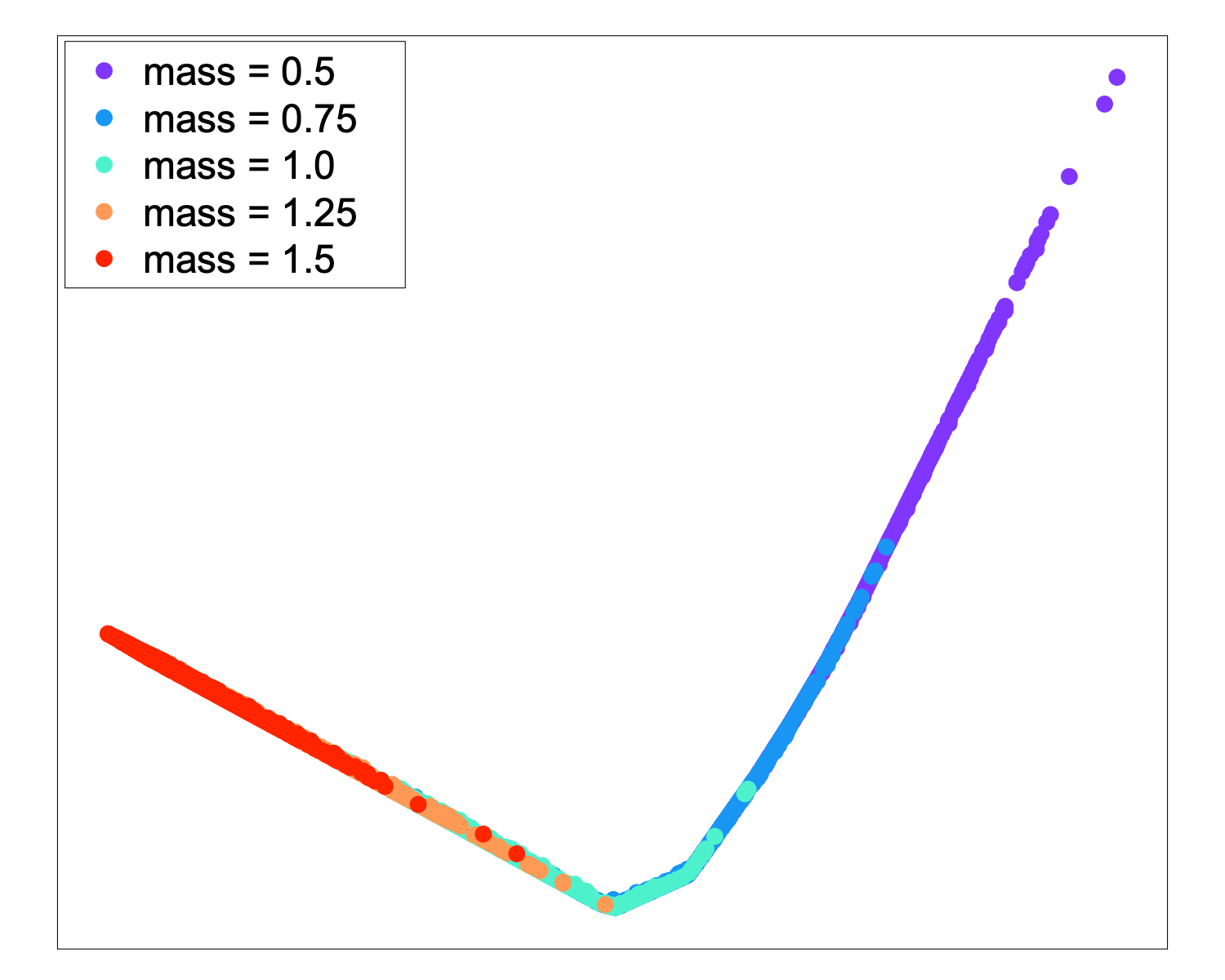} \label{fig:halfcheetah_ablation_pca}}
\vspace{-0.1in}
\\
\subfigure[Ant]
{
\includegraphics[width=0.3\textwidth]{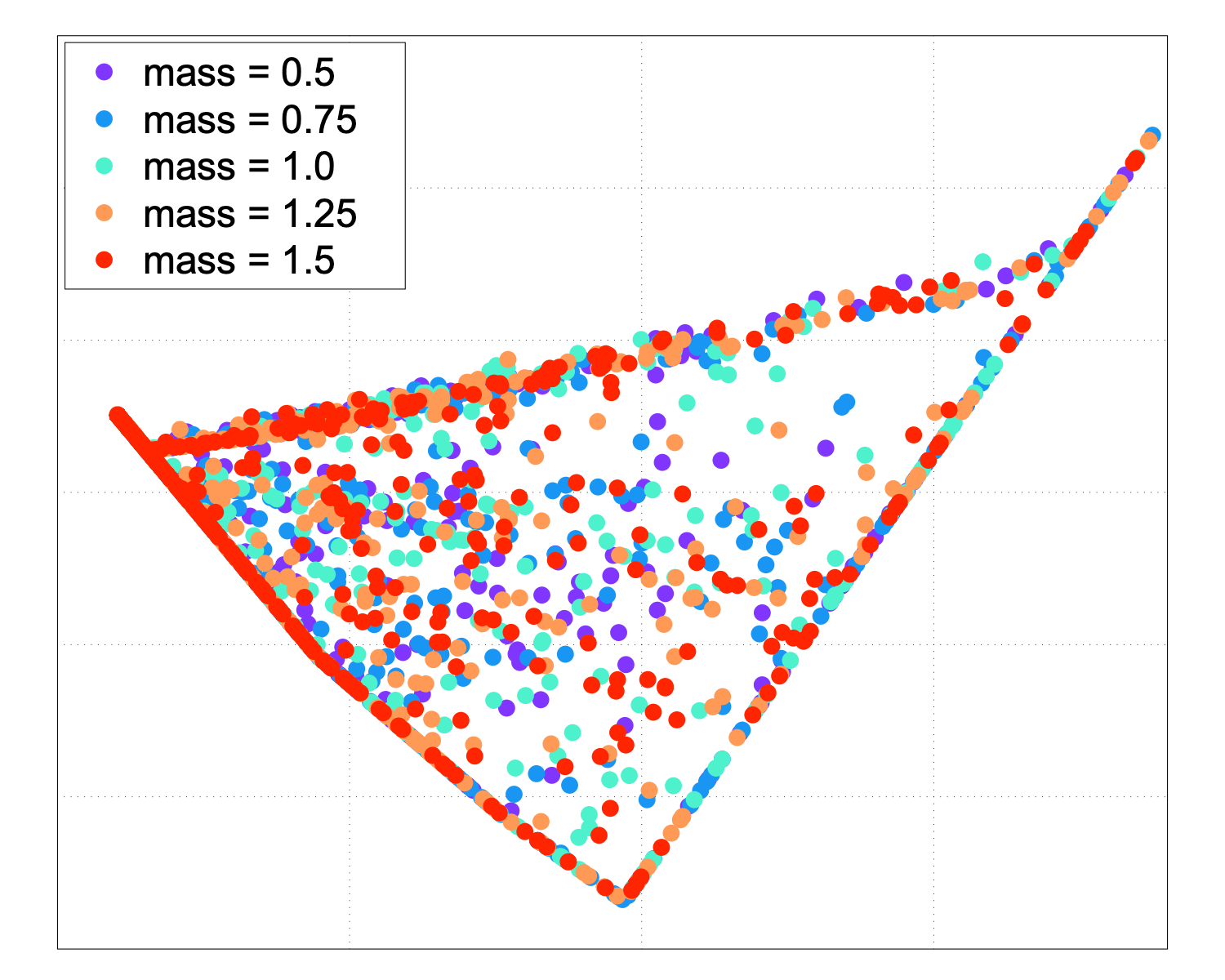} \label{fig:ant_ablation_pca}} 
\,
\subfigure[CrippledHalfcheetah]
{
\includegraphics[width=0.3\textwidth]{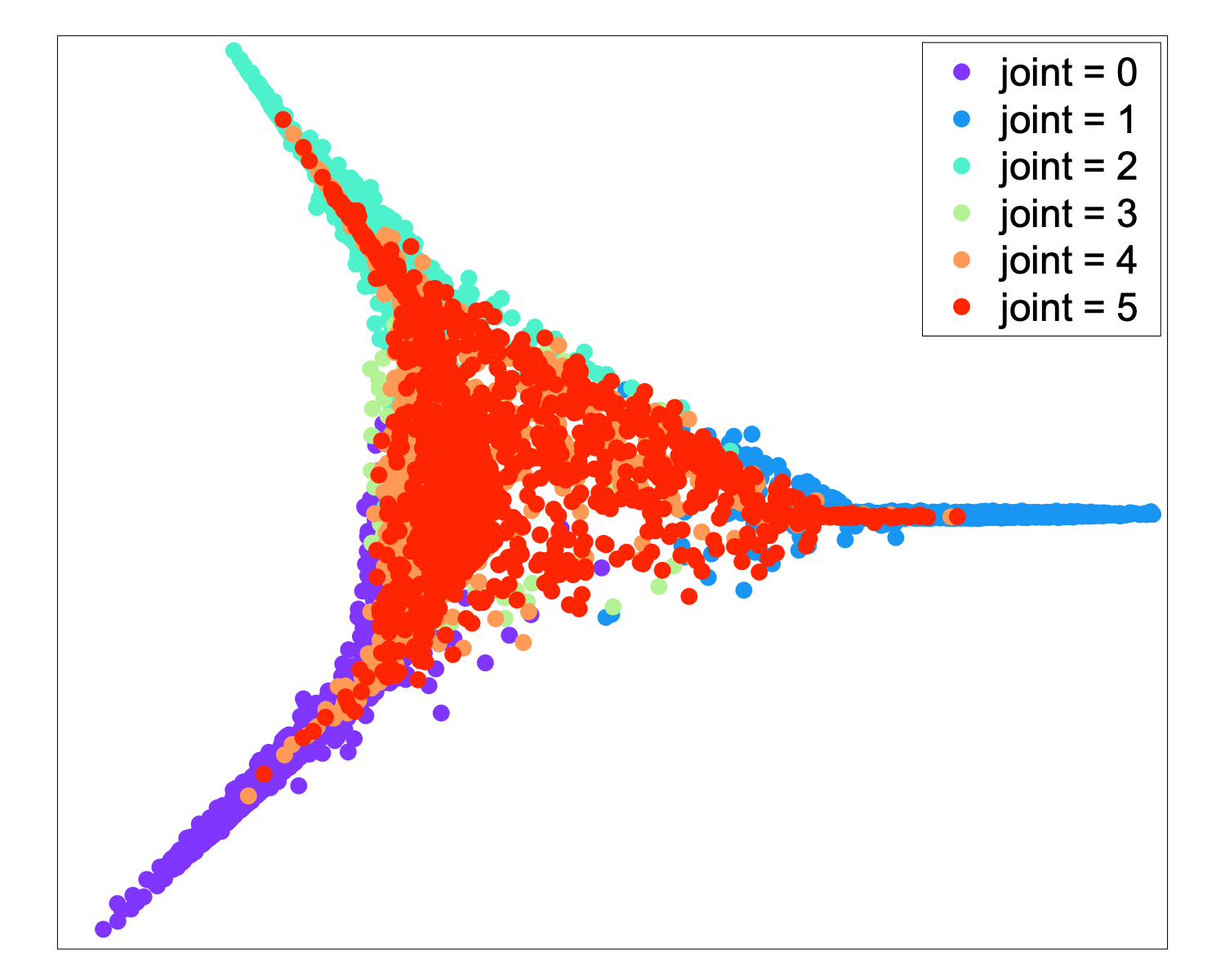} \label{fig:crippled_halfcheetah_ablation_pca}}
\,
\subfigure[SlimHumanoid]
{
\includegraphics[width=0.3\textwidth]{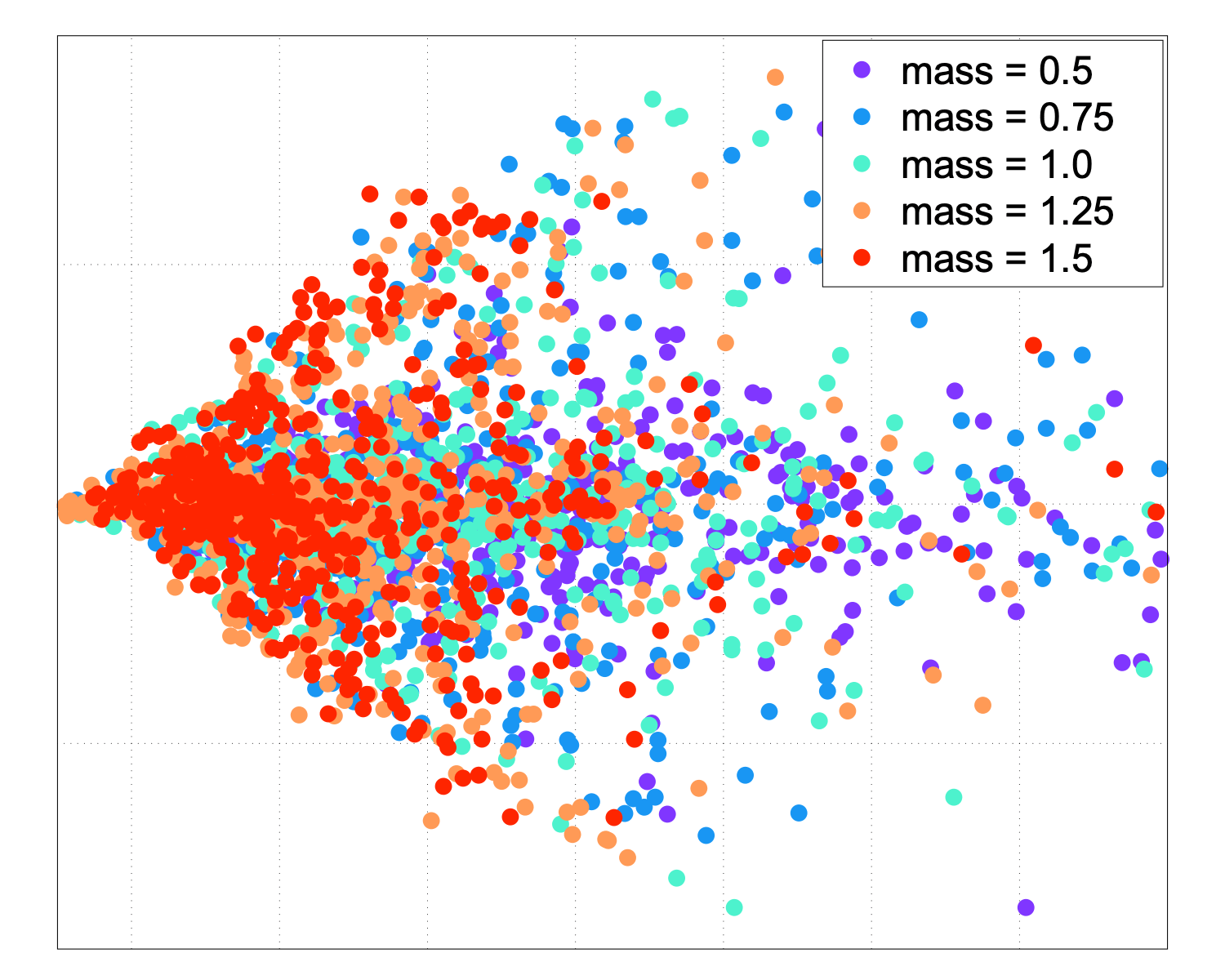} \label{fig:slim_humanoid_ablation_pca}}
\vspace{-0.1in}
\caption{
PCA visualization of context latent vectors extracted from trajectories collected in various control tasks. Embedded points from environments with the same parameter have the same color.}
\label{fig:embedding_analysis_pca}
\end{figure*}

\begin{figure*} [h] \centering
\subfigure[CartPole]
{
\includegraphics[width=0.3\textwidth]{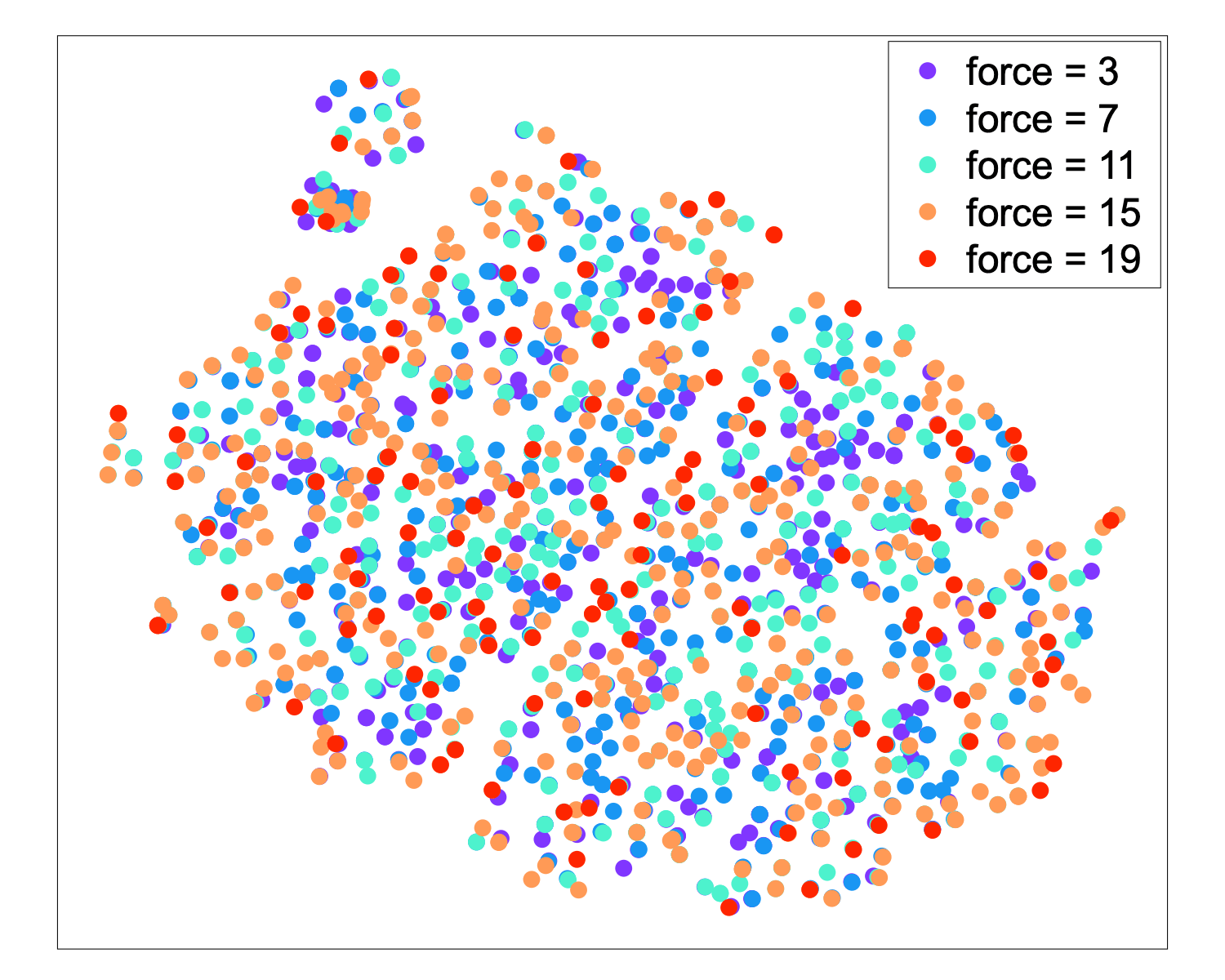} \label{fig:cartpole_ablation_raw_tsne}} 
\,
\subfigure[Pendulum]
{
\includegraphics[width=0.3\textwidth]{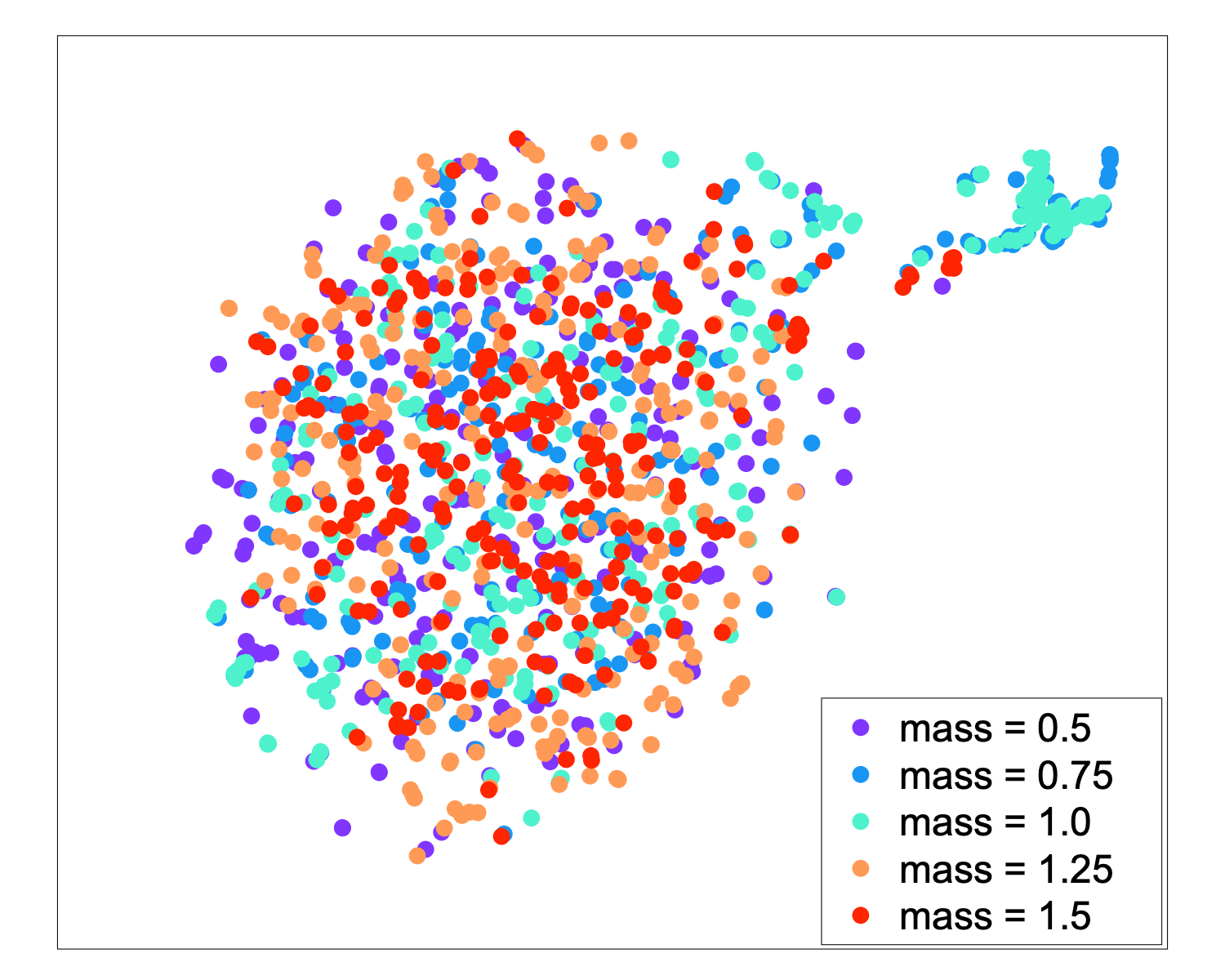} \label{fig:pendulum_ablation_raw_tsne}}
\,
\subfigure[HalfCheetah]
{
\includegraphics[width=0.3\textwidth]{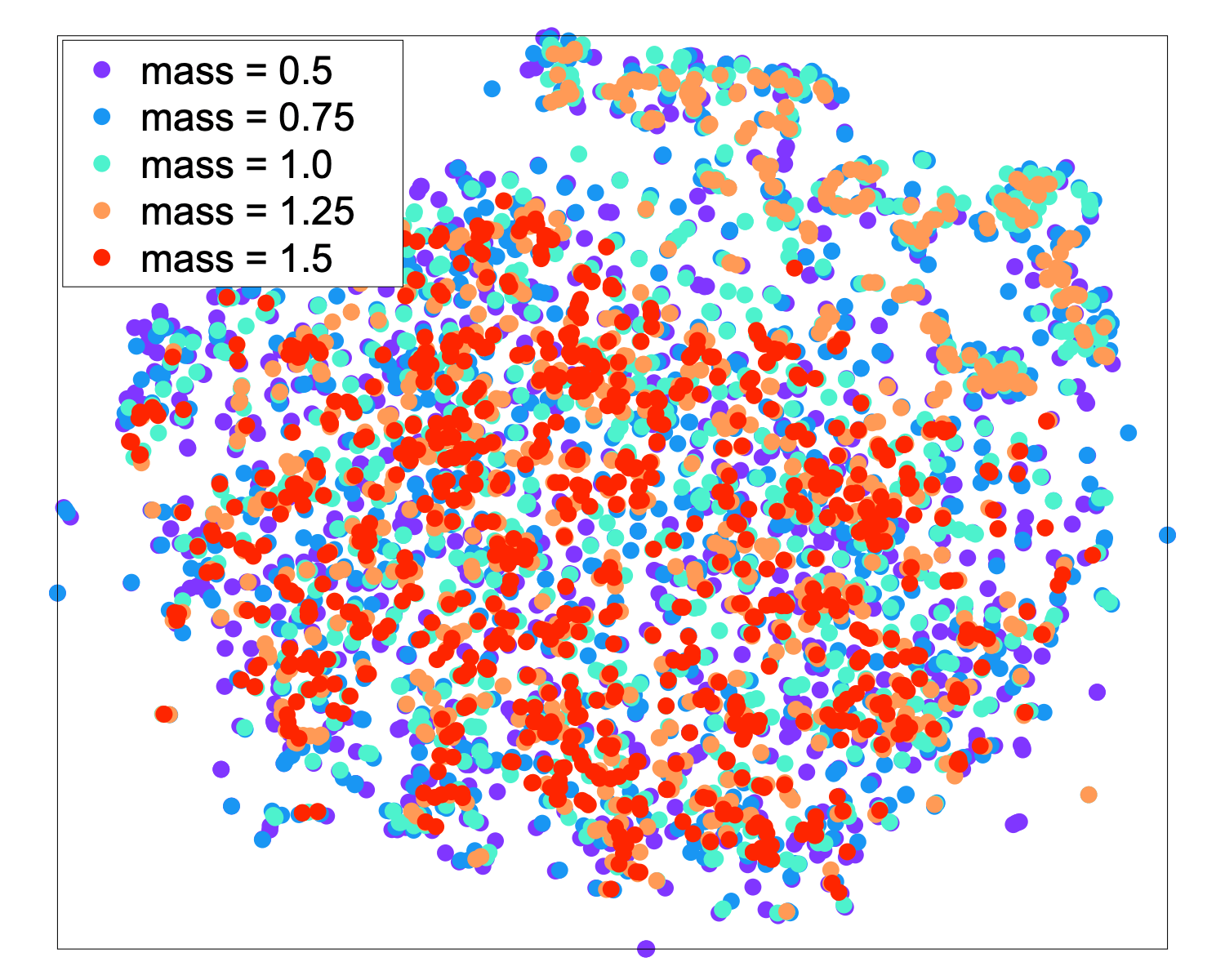} \label{fig:halfcheetah_ablation_raw_tsne}}
\vspace{-0.1in}
\\
\subfigure[Ant]
{
\includegraphics[width=0.3\textwidth]{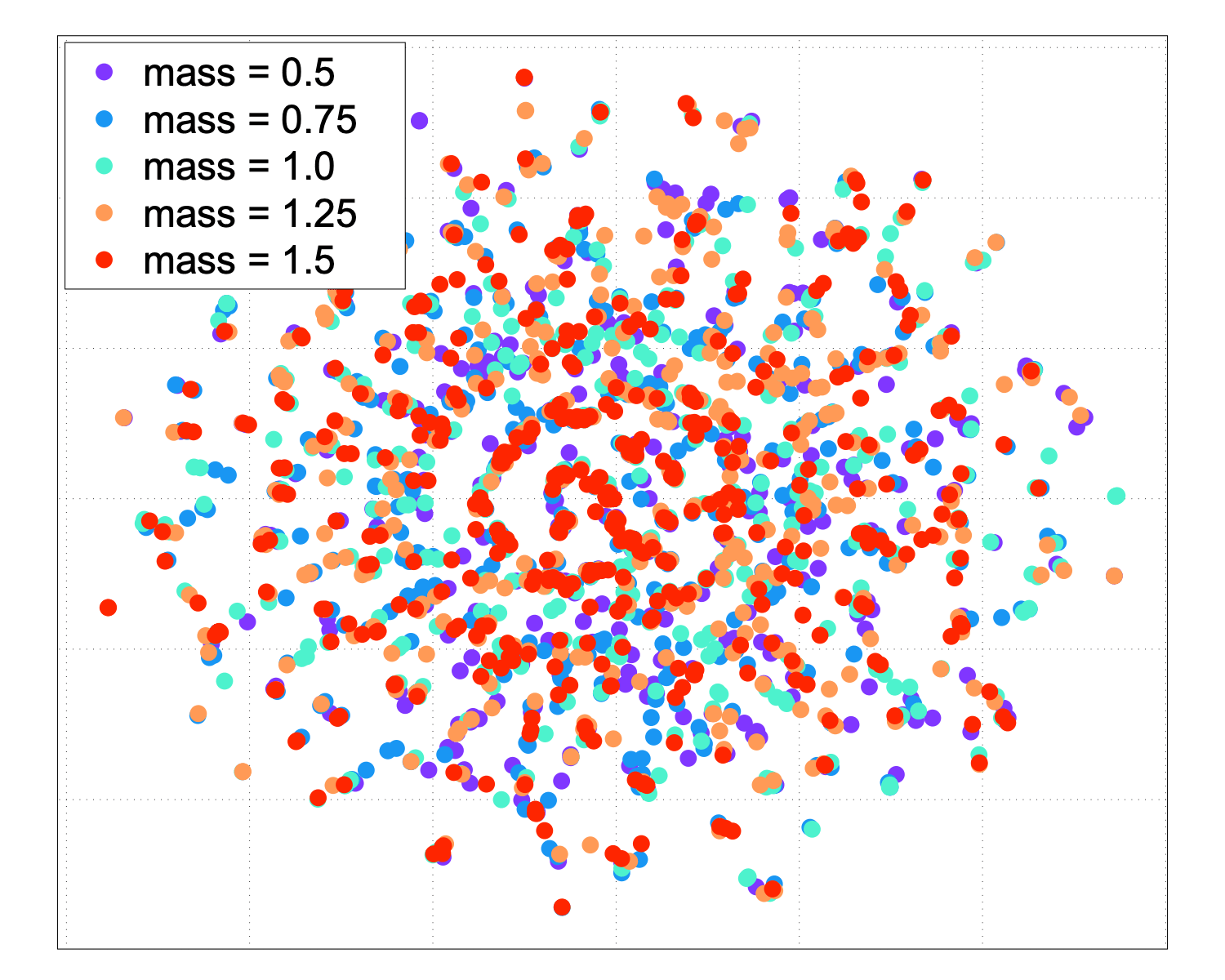} \label{fig:ant_ablation_raw_tsne}} 
\,
\subfigure[CrippledHalfcheetah]
{
\includegraphics[width=0.3\textwidth]{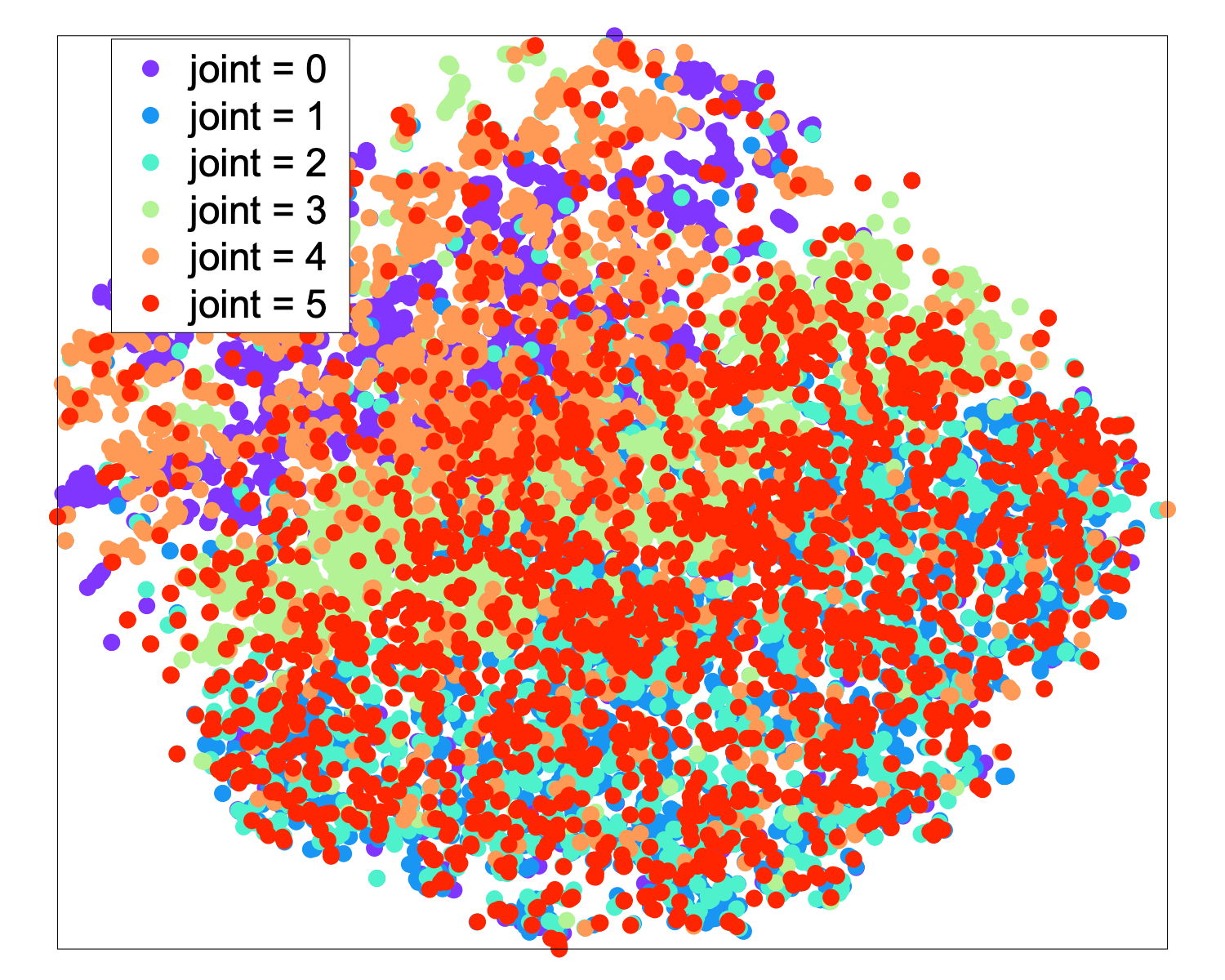} \label{fig:crippled_halfcheetah_ablation_raw_tsne}}
\,
\subfigure[SlimHumanoid]
{
\includegraphics[width=0.3\textwidth]{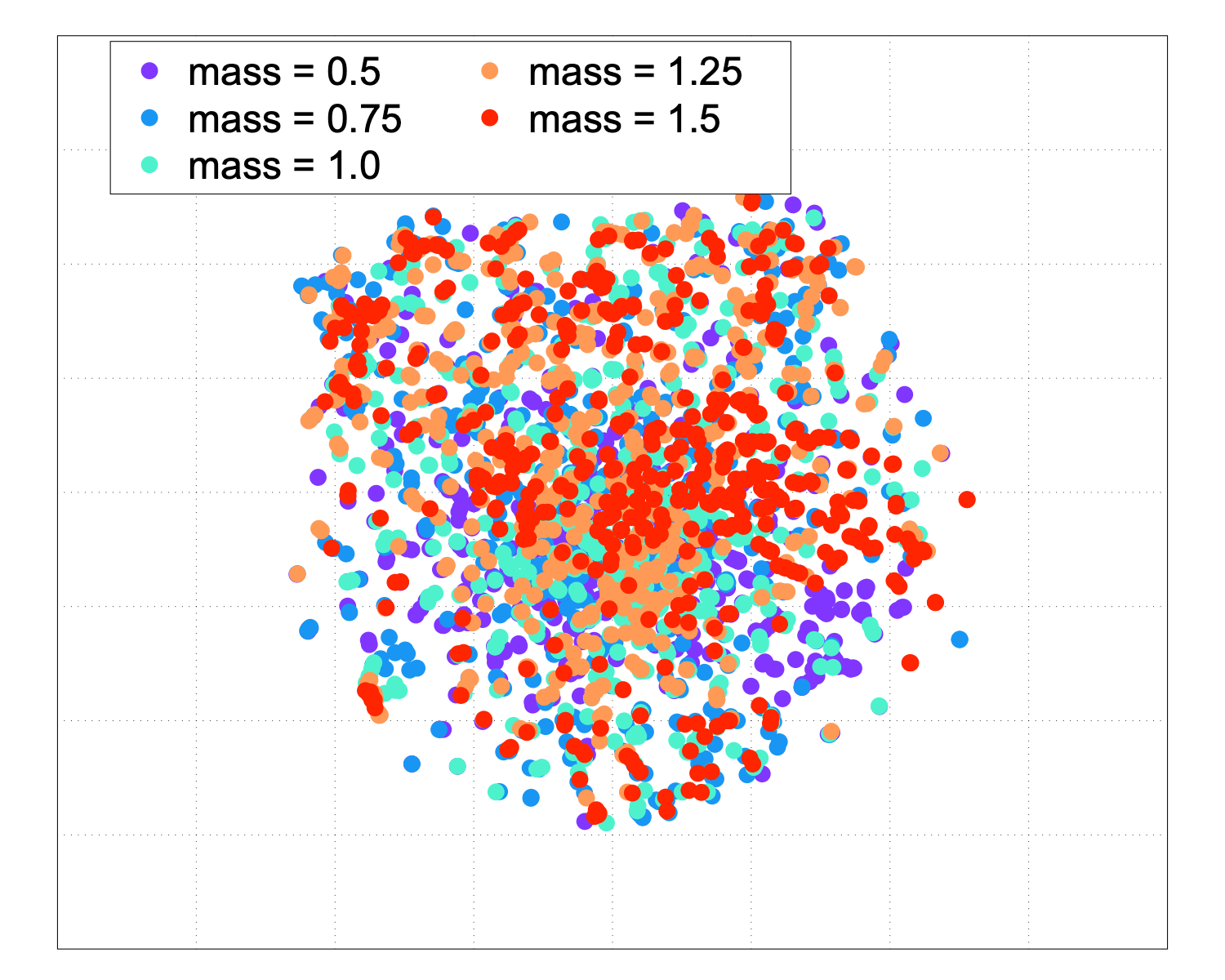} \label{fig:slim_humanoid_ablation_raw_tsne}}
\vspace{-0.1in}
\caption{
t-SNE \cite{maaten2008visualizing} visualization of raw state-action vectors extracted from trajectories collected in various control tasks. Embedded points from environments with the same parameter have the same color.}
\label{fig:embedding_analysis_raw_tsne}
\vspace{-0.1in}
\end{figure*}

\begin{figure*} [h!] \centering
\subfigure[CartPole]
{
\includegraphics[width=0.3\textwidth]{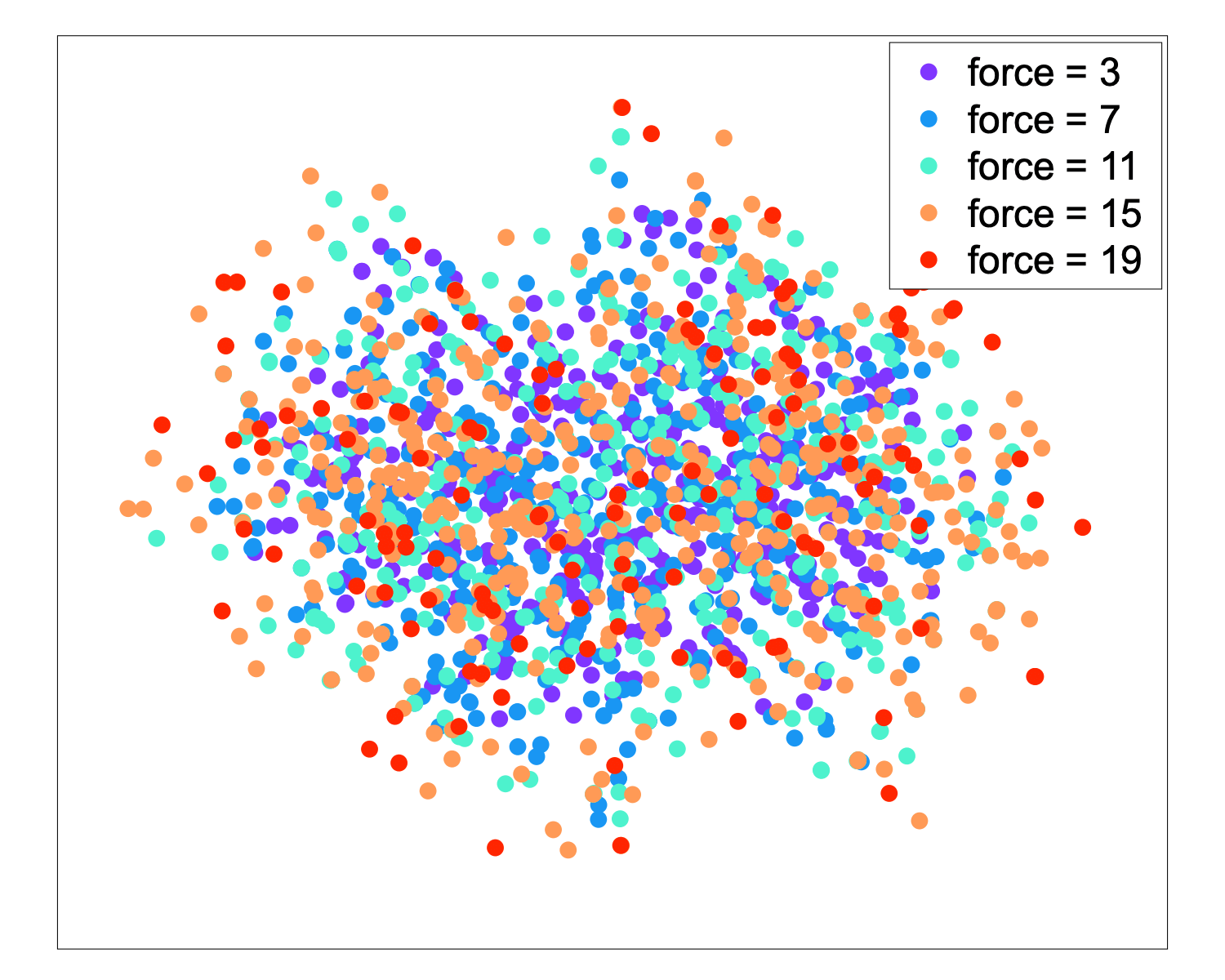} \label{fig:cartpole_ablation_raw_pca}} 
\,
\subfigure[Pendulum]
{
\includegraphics[width=0.3\textwidth]{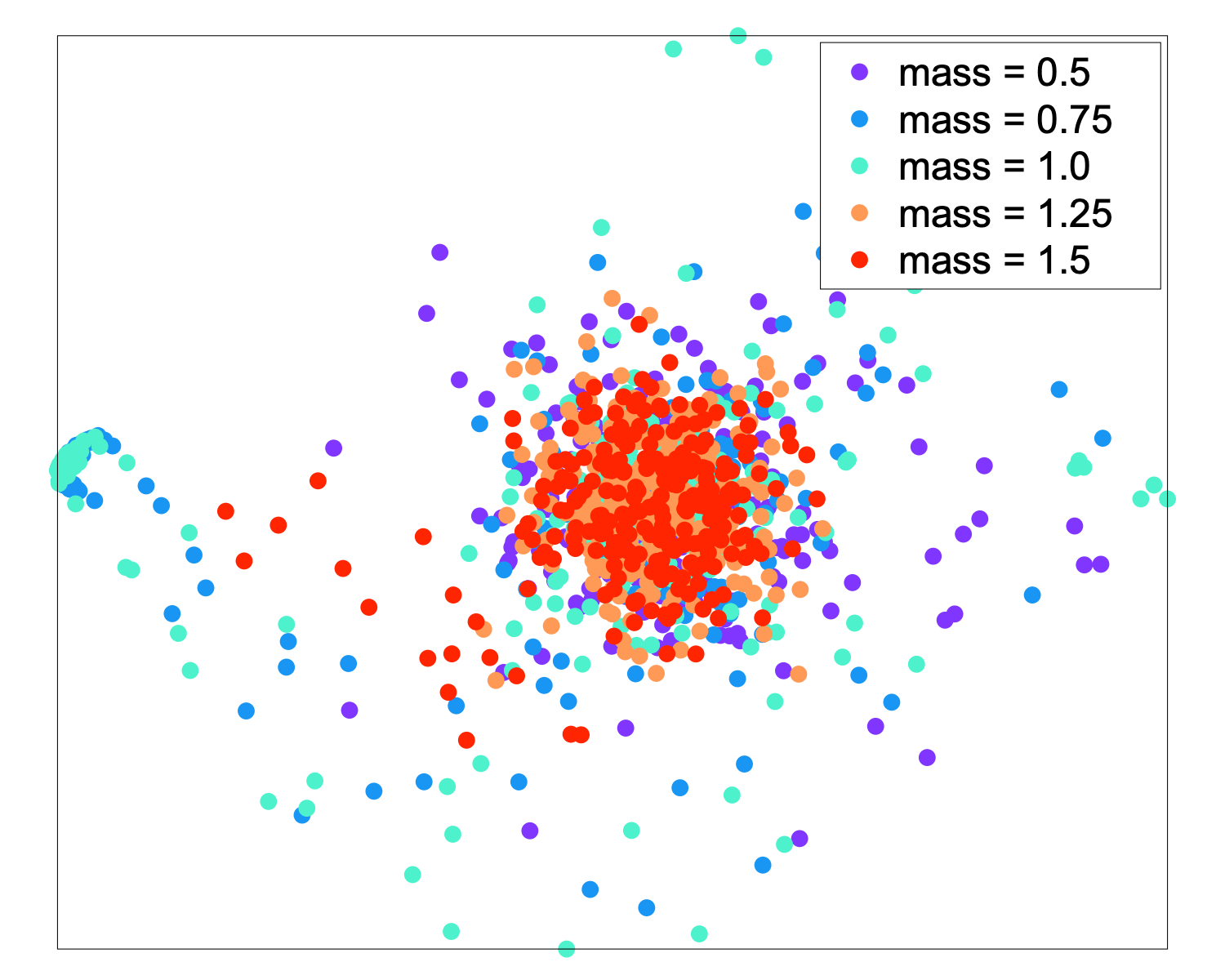} \label{fig:pendulum_ablation_raw_pca}}
\,
\subfigure[HalfCheetah]
{
\includegraphics[width=0.3\textwidth]{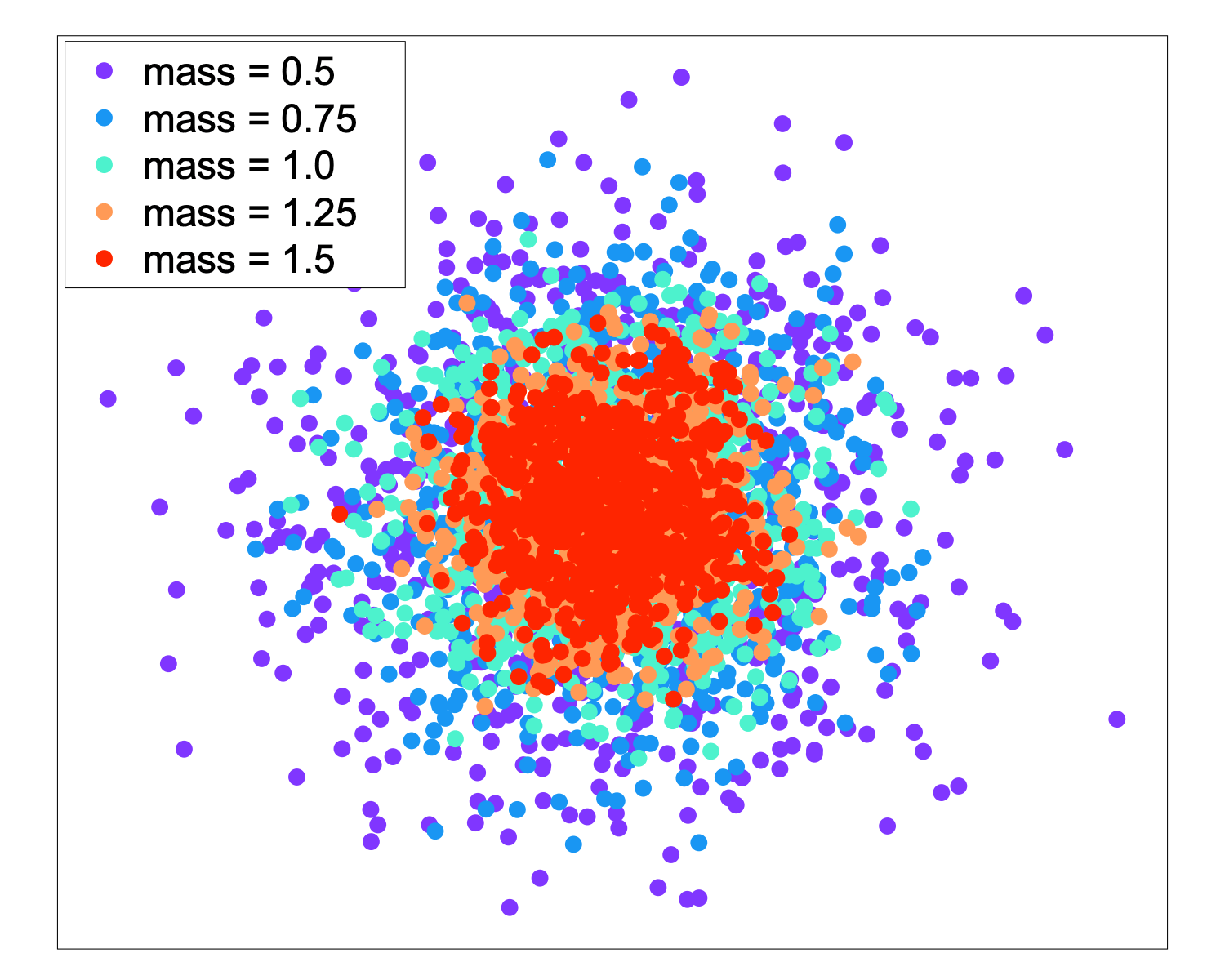} \label{fig:halfcheetah_ablation_raw_pca}}
\vspace{-0.1in}
\\
\subfigure[Ant]
{
\includegraphics[width=0.3\textwidth]{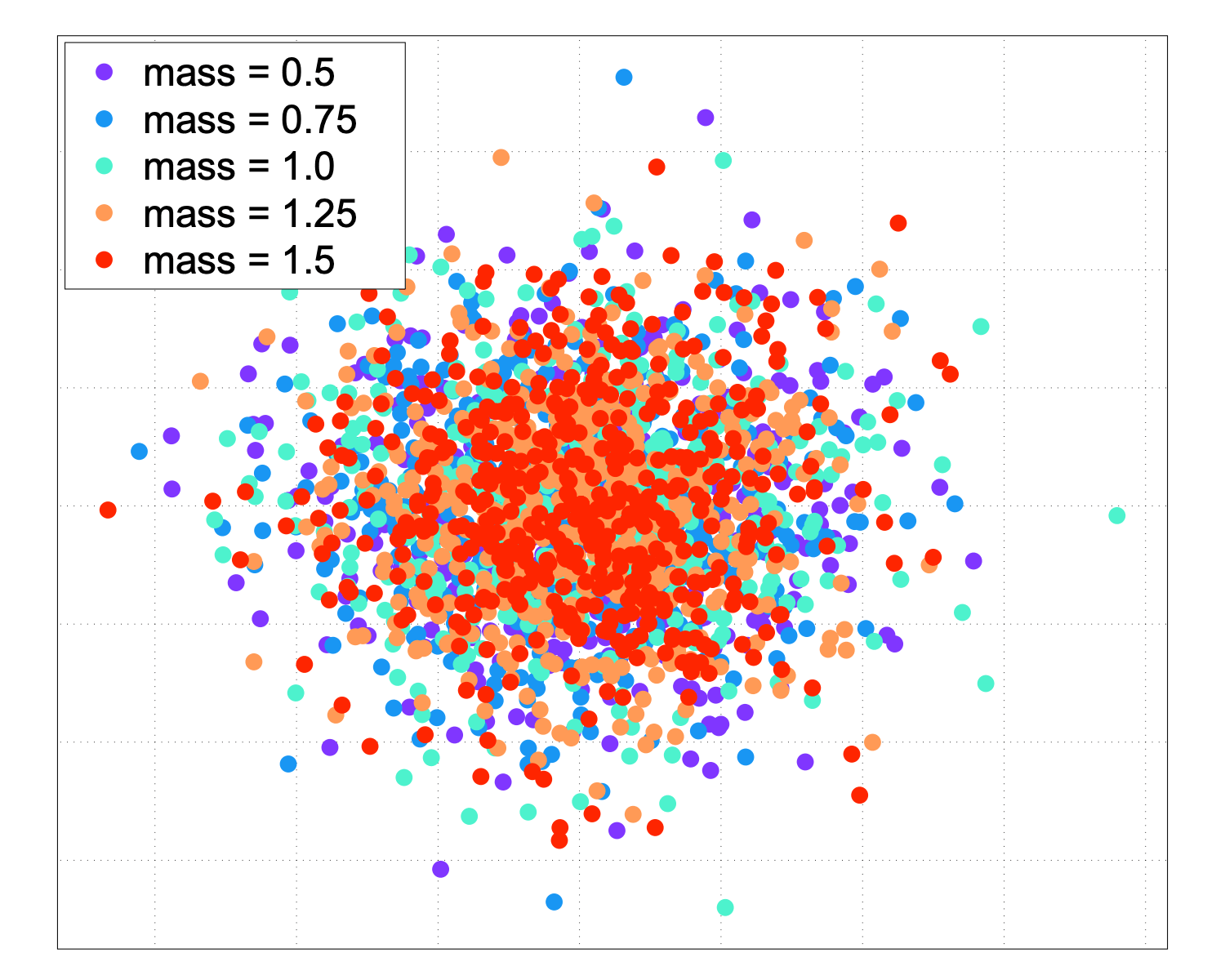} \label{fig:ant_ablation_raw_pca}} 
\,
\subfigure[CrippledHalfcheetah]
{
\includegraphics[width=0.3\textwidth]{figures/ant_raw_pca.png} \label{fig:crippled_halfcheetah_ablation_raw_pca}}
\,
\subfigure[SlimHumanoid]
{
\includegraphics[width=0.3\textwidth]{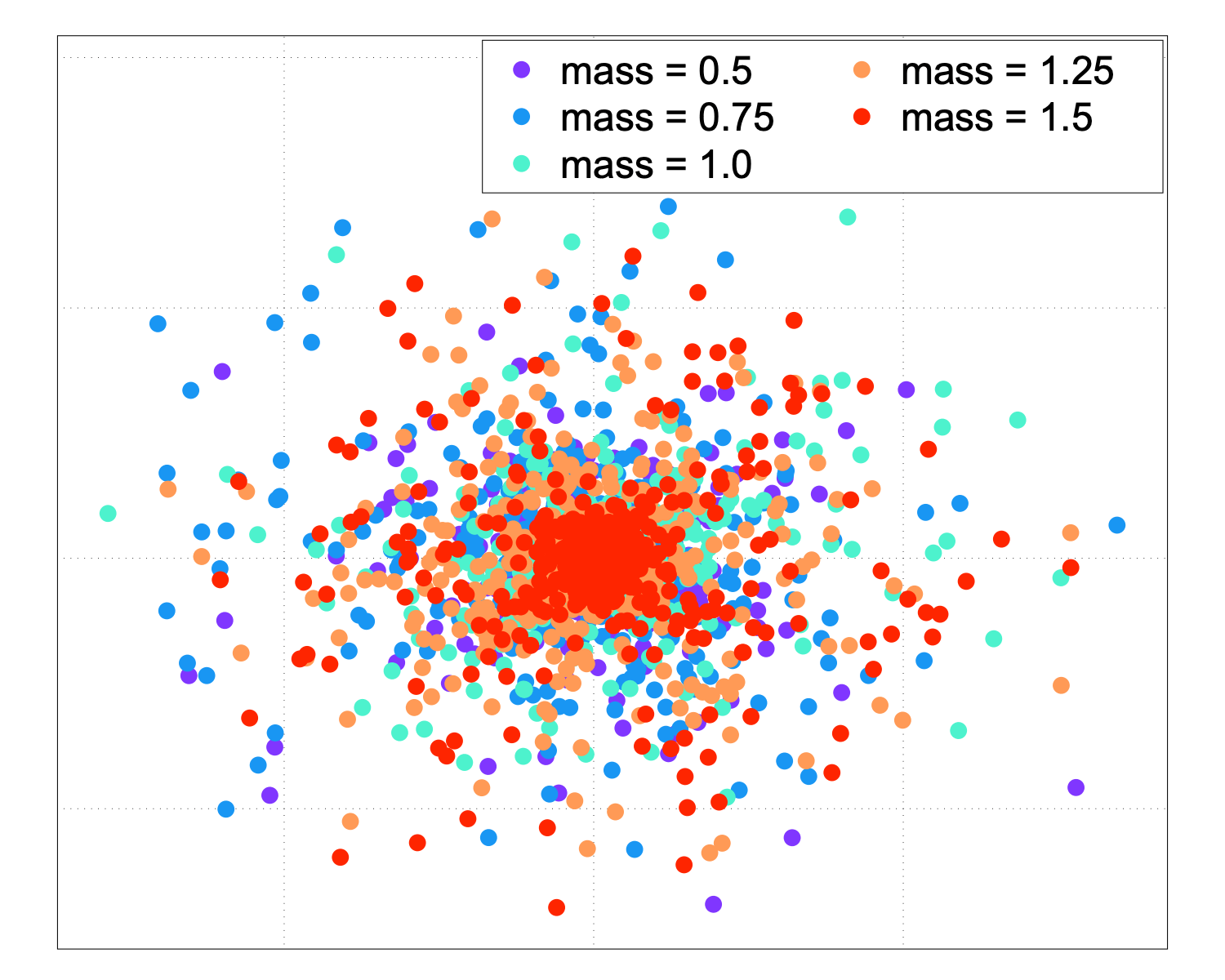} \label{fig:slim_humanoid_ablation_raw_pca}}
\vspace{-0.1in}
\caption{
PCA visualization of raw state-action vectors extracted from trajectories collected in various control tasks. Embedded points from environments with the same parameter have the same color.}
\label{fig:embedding_analysis_raw_pca}
\end{figure*}

\end{document}